\documentclass[10pt,twocolumn,letterpaper]{article}

\usepackage{iccv}              
\usepackage{threeparttable}
\usepackage[normalem]{ulem}
\useunder{\uline}{\ul}{}
\usepackage{multirow}
\usepackage{amssymb}
\usepackage{pifont}
\definecolor{DarkGreen}{RGB}{17,193,56}
\newcommand{\cmark}{\textcolor{DarkGreen}{\ding{51}}}%
\newcommand{\xmark}{\textcolor{red}{\ding{55}}}%
\usepackage[title]{appendix}
\usepackage{subcaption}

\definecolor{iccvblue}{rgb}{0.21,0.49,0.74}
\usepackage[pagebackref,breaklinks,colorlinks,allcolors=iccvblue]{hyperref}

\def\paperID{3994} 
\def\confName{ICCV}
\def\confYear{2025}

\title{Face-LLaVA: Facial Expression and Attribute Understanding through Instruction Tuning}

\author{Ashutosh Chaubey \hspace{1cm} Xulang Guan \hspace{1cm} Mohammad Soleymani\\
Institute for Creative Technologies, University of Southern California\\
Los Angeles, California, USA\\
{\tt\small \{achaubey, xulanggu\}@usc.edu, soleymani@ict.usc.edu}}

\begin{document}
\maketitle
\begin{abstract}
The human face plays a central role in social communication, necessitating the use of performant computer vision tools for human-centered applications. We propose Face-LLaVA, a multimodal large language model for face-centered, in-context learning, including facial expression and attribute recognition. Additionally, Face-LLaVA is able to generate natural language descriptions that can be used for reasoning. 
Leveraging existing visual databases, we first developed FaceInstruct-1M, a face-centered database for instruction tuning MLLMs for face processing. We then developed a novel face-specific visual encoder powered by Face-Region Guided Cross-Attention that integrates face geometry with local visual features. We evaluated the proposed method across nine different datasets and five different face processing tasks, including facial expression recognition, action unit detection, facial attribute detection, age estimation and deepfake detection. Face-LLaVA achieves superior results compared to existing open-source MLLMs and competitive performance compared to commercial solutions. Our model output also receives a higher reasoning rating by GPT under a zero-shot setting across all the tasks. Both our dataset and model wil be released at \href{https://face-llava.github.io/}{https://face-llava.github.io/} to support future advancements in social AI and foundational vision-language research. 

\end{abstract}    
\section{Introduction}
\label{sec:intro}


The human face serves as a fundamental conduit for social communication. Hence, face analysis including facial expression recognition \cite{wang2024surveyfacialexpressionrecognition}, action unit (AU) detection \cite{Zhi2020-survey_action_units}, facial attribute detection \cite{zhu2022celebvhq}, age estimation \cite{Angulu2018-uw_age_survey} has received considerable attention. 
These tasks have a plethora of applications such as human-computer interaction \cite{hci_face}, social behavior assessment \cite{Pixton2011-bk_psychology_face}, e-learning \cite{face_elearning} and surveillance \cite{pei2024deepfakegenerationdetectionbenchmark}, among others.


Existing face analysis approaches suffer from two major limitations: (i) they are developed for specific tasks such as expression recognition \cite{dfew_s2d, transfer_rafdb} or attribute detection \cite{attr_liu_celeba, dmm_cnn}, limiting their generalizability; and (ii) they primarily output labels without natural language description for their predictions. The ability to use natural language to describe predictions is crucial for critical applications such as healthcare and surveillance \cite{Ennab2022-vq_interpretable_healthcare,Wu2022-dn_interpretable_surveillance}.

To address the first limitation, recent advances have focused on developing generalist face models capable of handling multiple tasks \cite{marlin, faceptor}. However, these models still lack a natural language interface, as they do not provide any insight into their decision-making process. The recent advent of Multimodal Large Language Models (MLLMs) such as LLaVA \cite{liu2023llava, zhang2024llava_video, li2024llava_onevision} has led to the application of such models for specific visual processing tasks, including face processing. EmoVIT \cite{Xie2024EmoVIT} uses instruction tuning to enhance visual emotion understanding of large vision and language models. Emotion-LLaMA \cite{cheng2024emotionllama} uses an MLLM to take audio, video and text inputs to generate a description of emotional responses and reasoning in a given video. Foteinopoulou et al. \cite{NEURIPS2024_foteinopoulou} investigate the deepfake detection and reasoning capabilities of existing large language models (LLMs). However, these approaches are task-specific and involve prediction using background information or audio. More recently, MLLMs have been adapted for face processing, e.g., FABA \cite{li2024facialfaba} and VL-FAU \cite{Ge2024-yp-vlfau} attempting to bridge the gap by leveraging powerful visual encoders and LLMs reasoning capabilities. However, the capabilities of these methods are limited to analyzing static images, solely focusing on facial expression analysis.

\begin{figure}
    \centering
    \includegraphics[width=\linewidth]{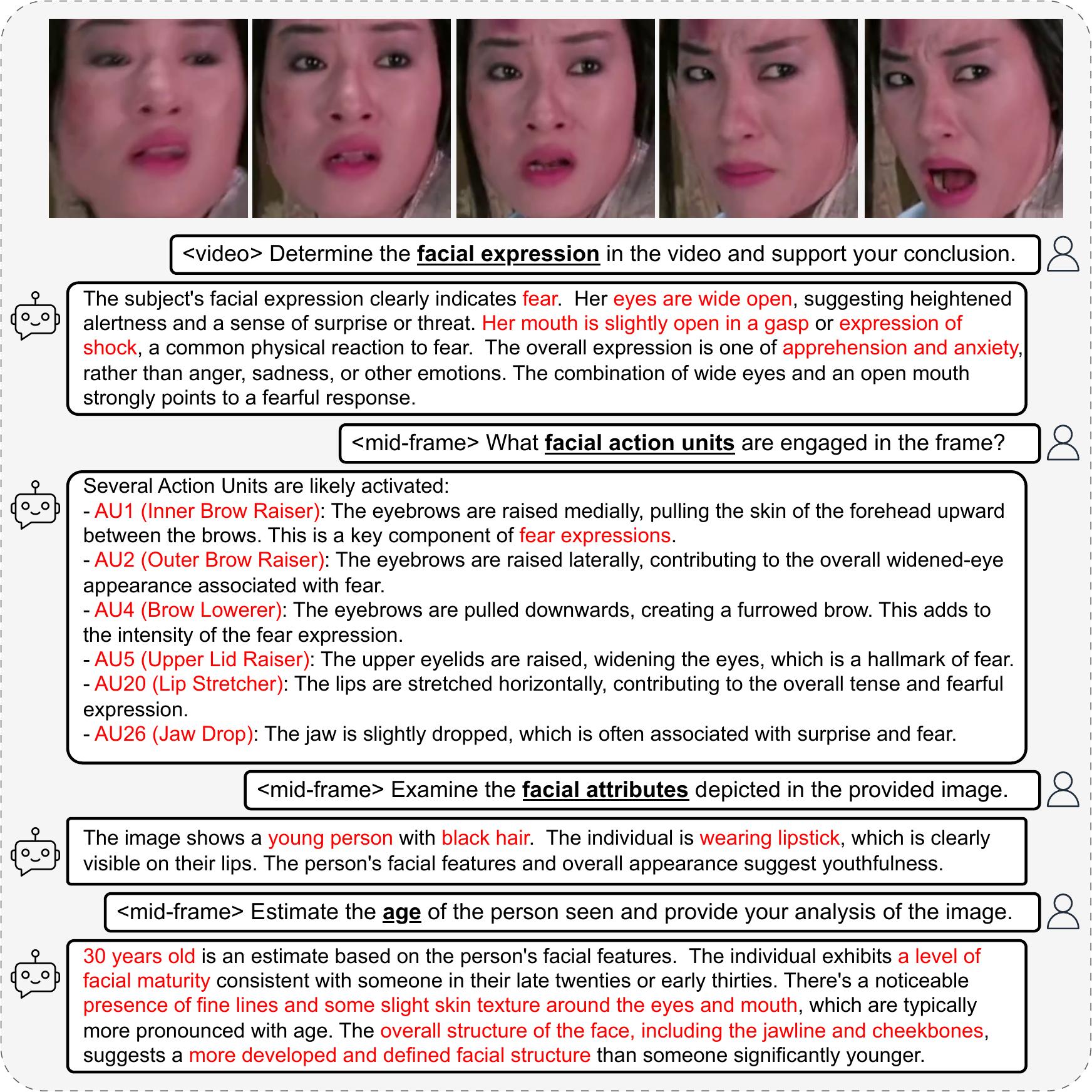}
    \vspace{-1em}
    \caption{A sample conversation with Face-LLaVA highlighting different face tasks that our model is capable of performing.}
    \label{fig:intro_chatbot}
    \vspace{-1em}
\end{figure}

In this paper, we propose Face-LLaVA, a generalist MLLM that can perform perception with reasoning on different face analysis tasks. \cref{fig:intro_chatbot} shows the interaction of a user with our model. Note that given a video or an image, our model can recognize facial expressions, facial action units, age, and facial attributes and identify face forgery (see \cref{fig:qual_samples}). This advancement paves the way for developing general-purpose AI-driven social agents with enhanced facial reasoning capabilities, making them more applicable to real-world scenarios.

Training an MLLM for face analysis requires a vast amount of instruction-tuning data covering different tasks. However, existing instruction-tuning datasets \cite{li2024facialfaba, cheng2024emotionllama, lian2023explainable_emer} for related tasks are relatively small, task-specific, and often limited to either images or videos (see \cref{tab:data_comparison}). To address this limitation, we introduce \emph{FaceInstruct-1M}, a large-scale face-specific dataset comprising one million samples, including around 850k images and 120 hours of video data, annotated with task-specific instructions and descriptions across five distinct face analysis tasks. Furthermore, as the dataset is automatically generated, we employ a GPT-4o \cite{gpt4omini}-assisted evaluation and filtering process to ensure high-quality data for fine-tuning the MLLM.

General-purpose vision encoders used in MLLMs, such as CLIP \cite{radford2021learningtransferablevisualmodels_clip} and LanguageBind \cite{zhu2024languagebind}, are not designed to extract face-specific features such as facial landmarks, which are crucial for fine-grained face analysis. To address this, we propose a novel architecture that is specifically designed for face processing to enhance its performance. Unlike previous approaches \cite{li2024facialfaba}, our method is more efficient in utilizing the LLM's context window and is generalizable to both images and videos. Specifically, we introduce a \emph{face-region landmark projector}, which maps face expert (landmark) features to face-region tokens and refines visual features via cross-attention with the extracted visual tokens.

We conduct extensive experiments on nine datasets spanning five different face analysis tasks, demonstrating that our method outperforms existing MLLMs across all benchmarks in a zero-shot setting and achieves performance competitive with supervised task-specific approaches. To further assess the reasoning capabilities of our model, we conduct automatic GPT-based evaluations comparing our approach against recent MLLMs and showcasing its superior generative and captioning capabilities.

In summary, the key contributions of this work are as follows. (i) We introduce \emph{FaceInstruct-1M}, a large-scale face analysis dataset comprising one million instruction-tuning samples, including both images and videos. The dataset spans five key face-related tasks: expression recognition, action unit (AU) detection, attribute detection, age estimation, and deepfake detection. (ii) We propose \emph{Face-LLaVA}, a multimodal large language model architecture for face analysis that incorporates face landmark features using a face-region projector and face-region-guided cross-attention, enabling effective instruction tuning for face processing tasks. (iii) Through extensive experiments, we demonstrate that our approach outperforms existing MLLMs in a zero-shot setting on nine datasets across five face analysis tasks, achieving superior performance on both traditional benchmarks and GPT-assisted evaluations of generated responses. Additionally, we show that our method achieves competitive performance with supervised task-specific techniques across all tasks. \\
\noindent \emph{FaceInstruct-1M} and \emph{Face-LLaVA} model weights will be released at \href{https://face-llava.github.io/}{https://face-llava.github.io/} upon acceptance.
\section{Related Works}
\subsection{Traditional face analysis}

Extensive research has been conducted on individual face analysis tasks, including expression recognition \cite{transfer_rafdb,rul_rafdb,rafdb_apvit,eac_rafdb_emotion,dfew_s2d,dfew_m3dfel,dfew_mae_dfer,xiang2024mtcaedfer}, action unit detection \cite{Shao2020JANetJF,piap_df,Li_2023_BMVC_recot,luo2022graphau}, age estimation \cite{age_ldl,age_mwr,Angulu2018-uw_age_survey}, attribute detection \cite{attr_liu_celeba,moon_age,dmm_cnn}, and deepfake detection \cite{xception_net,f3_net,mesonet}. However, the research focus is gradually shifting towards developing generalist face analysis models \cite{swinface,faceptor,narayan2024facexformer} and learning robust facial representations \cite{Shi_2020_CVPR_urface,zheng2022general_farl,marlin}.
Faceptor \cite{faceptor} builds upon FaRL \cite{zheng2022general_farl} and SWINFace \cite{swinface} by introducing a single encoder with a dual-decoder transformer to handle multiple face analysis tasks. MARLIN \cite{marlin} and PrefAce \cite{Hu_Wang_Hu_Peng_Wu_Zhu_Ong_2024} leverage self-supervised learning on video data using masked autoencoders (MAE) \cite{he2021maskedautoencodersscalablevision_mae,tong2022videomae} to learn robust facial representations. PCL \cite{Liu_2023_CVPR} employs a pose-disentangled decoder for contrastive learning, generating robust pose and appearance features for face analysis.
Building on these advancements, our work introduces a generalist face analysis model that extends beyond face perception by incorporating reasoning over its predictions. This enhances interpretability and enables more detailed and context-aware facial analyses.
\subsection{Face analysis using instruction tuning}
Multimodal LLMs enhance reasoning and analysis on visual and other inputs by leveraging the extensive knowledge encoded in LLMs \cite{liu2023llava,zhang2024llava_video,damonlpsg2025videollama3,lin-etal-2024-videollava,qwen25vl}. These models require large-scale instruction-tuning datasets consisting of a visual (or other modality) input, an instruction, and corresponding responses \cite{zhang2024llava_video,liu2023llava}.
The use of MLLMs for face analysis remains limited due to the scarcity of large-scale instruction-tuning data. Most traditional face analysis datasets \cite{jiang2020dfew,affectnet,disfa_dataset,attr_liu_celeba,morph_ii_dataset,roessler2019faceforensicspp} provide only categorical or numerical labels (see \cref{tab:data_comparison}). While instruction or description datasets exist for general emotion recognition, they primarily rely on background context and audio rather than facial cues alone \cite{liu_mafw_2022,lian2023explainable_emer,cheng2024emotionllama,Xie2024EmoVIT}. FABAInstruct \cite{li2024facialfaba} is a face-specific dataset but is limited to still images and affective behavior analysis.
In contrast, we introduce \emph{FaceInstruct-1M}, a large-scale instruction-tuning dataset explicitly designed for face analysis across multiple face-related tasks.


While prior research has explored reasoning with large VLMs for face-related tasks \cite{cheng2024emotionllama,NEURIPS2024_foteinopoulou,Xie2024EmoVIT}, dedicated efforts toward comprehensive face-specific reasoning remain scarce. AU-LLAVA \cite{hu2024unifiedfacialactionunit_aullava} utilizes an LLM for AU detection and intensity estimation, yet its outputs are restricted to categorical or numerical predictions without reasoning. VL-FAU \cite{Ge2024-yp-vlfau} introduces a vision-language framework for interpretable AU detection, while EmoLA \cite{li2024facialfaba} integrates landmark prior tokens with visual embeddings to facilitate facial affective behavior analysis with reasoning. Similarly, EMO-LLaMA \cite{xing2024emollama} employs a face-information mining module to enhance facial feature encoding, demonstrating both reasoning and conversational capabilities.

More recently, Face-MLLM \cite{sun2024facemllmlargefaceperception} employed Gemini to automatically annotate face images in the Laion-Face \cite{laion_face} dataset for instruction tuning, although its focus remains solely on perception rather than reasoning over model predictions, and it depends heavily on Gemini’s face analysis capabilities to build its dataset. Similarly, FaVChat \cite{zhao2025favchatunlockingfinegrainedfacial} leverages existing face video datasets for instruction tuning; however, due to significant class imbalances in attributes such as emotion in datasets like CelebV-HQ \cite{zhu2022celebvhq}, the overall dataset becomes imbalanced. Moreover, both approaches do not generalize well to both images and videos and often include background information that might cause model hallucinations. In contrast, Face-LLaVA handles both images and videos, relies on standard, well-established datasets for instruction data construction, and is tailored specifically for videos, thereby reducing the likelihood of hallucinations stemming from background content.

\section{FaceInstruct-1M}
\label{sec:dataset_faceinstruct_1m}
\begin{table}[]
\centering
\resizebox{\linewidth}{!}{%
\begin{tabular}{l|c|cc|cccccc|c}
\hline \hline
\rowcolor[HTML]{C0C0C0} 
\multicolumn{1}{c|}{\cellcolor[HTML]{C0C0C0}} & \multicolumn{1}{c|}{\cellcolor[HTML]{C0C0C0}} & \multicolumn{2}{c|}{\cellcolor[HTML]{C0C0C0}\textbf{Modality}} & \multicolumn{6}{c|}{\cellcolor[HTML]{C0C0C0}\textbf{Number of Samples}} & \cellcolor[HTML]{C0C0C0} \\
\rowcolor[HTML]{C0C0C0} 
\multicolumn{1}{c|}{\multirow{-2}{*}{\cellcolor[HTML]{C0C0C0}\textbf{Dataset}}} & \multicolumn{1}{c|}{\multirow{-2}{*}{\cellcolor[HTML]{C0C0C0}\textbf{\begin{tabular}[c|]{@{}c@{}}Face\\ Spec.\end{tabular}}}} & \textbf{Img.} & \textbf{Vid.} & \textbf{Expr.} & \textbf{AU} & \textbf{Attr.} & \textbf{Age} & \textbf{DF.} & \textbf{Tot.} & \multirow{-2}{*}{\cellcolor[HTML]{C0C0C0}\textbf{\begin{tabular}[c]{@{}c@{}}Anno.\\ Type\end{tabular}}} \\ \hline
RAF-DB \cite{rafdb} & \cmark & \cmark & \xmark & 30k & - & - & - & - & 30k & C \\
AffectNet \cite{affectnet} & \cmark & \cmark & \xmark & 440k & - & - & - & - & 440k & C \\
DFEW \cite{jiang2020dfew} & \cmark & \xmark & \cmark & 16k & - & - & - & - & 16k & C \\
FERV39K \cite{wang2022ferv39k} & \cmark & \xmark & \cmark & 39k & - & - & - & - & 39k & C \\
DISFA \cite{disfa_dataset} & \cmark & \cmark & \xmark & - & 140k & - & - & - & 140k & C \\
BP4D \cite{ZHANG2014692_bp4d_dataset} & \cmark & \cmark & \xmark & - & 150k & - & - & - & 150k & C \\
AffWild2 \cite{kollias2019affwild2extendingaffwilddatabase_affwild2} & \cmark & \cmark & \xmark & 2.6M & 2.6M & - & - & - & 2.6M & C+N \\
CelebA \cite{attr_liu_celeba} & \cmark & \cmark & \xmark & - & - & 200k & - & - & 200k & C \\
MORPH II \cite{morph_ii_dataset} & \cmark & \cmark & \xmark & - & - & - & 50k & - & 50k & N \\
UTK Face \cite{zhifei2017cvpr_utkface} & \cmark & \cmark & \xmark & - & - & - & 20k & - & 20k & N \\
FF++ \cite{roessler2019faceforensicspp} & \cmark & \xmark & \cmark & - & - & - & - & 5k & 5k & C \\
Fake-AVC \cite{khalid2021fakeavceleb} & \cmark & \xmark & \cmark & - & - & - & - & 20k & 20k & C \\ \hline
EMER \cite{lian2023explainable_emer} & \xmark & \xmark & \cmark & 450 & - & - & - & - & 450 & D \\
MAFW \cite{liu_mafw_2022} & \xmark & \xmark & \cmark & 10k & - & - & - & - & 10k & C+SD \\
EmoVIT \cite{Xie2024EmoVIT} & \xmark & \xmark & \cmark & 1.6k & - & - & - & - & 1.6k & I+D \\
MERR \cite{cheng2024emotionllama} & \xmark & \xmark & \cmark & 30k & - & - & - & - & 30k & I+D \\
\begin{tabular}[c]{@{}l@{}}FABA Instr. \cite{li2024facialfaba} \end{tabular} & \cmark & \cmark & \xmark & 20k & 20k & - & - & - & 20k & I+D \\  \hline
\textbf{\begin{tabular}[c]{@{}l@{}}Face-\\ Instruct-1M \end{tabular}} & \cmark & \cmark & \cmark & \textbf{\begin{tabular}[c]{@{}c@{}}350k\end{tabular}} & \textbf{290k} & \textbf{200k} & \textbf{70k} & \textbf{95k} & \textbf{1M} & \textbf{I+D} \\ \hline \hline
\end{tabular}%
}
\caption{Comparison of the proposed \emph{FaceInstruct-1M} dataset with various annotated datasets of face-related tasks. C:category, SD:short description, I:instruction, D:description, N:number (regression), DF.: deepfake detection, Face Spec.: Face-Specific., AU: action unit.}
\label{tab:data_comparison}
\vspace{-1em}
\end{table}


\begin{figure}[hb]
    \centering
    \includegraphics[width=\linewidth]{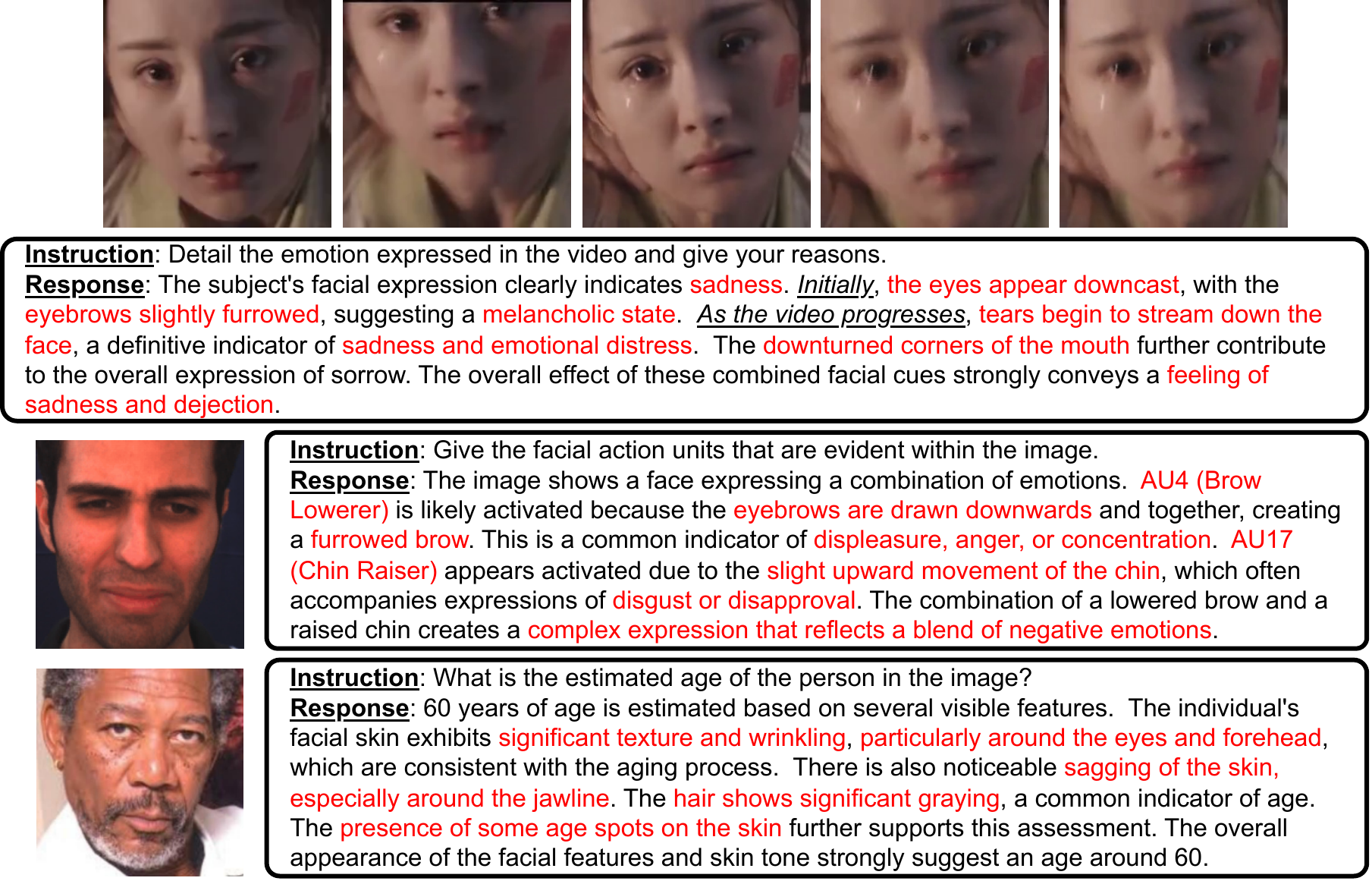}
    \vspace{-4pt}
    \caption{\emph{FaceInstruct-1M} dataset samples for different tasks.}
    \label{fig:qual_samples}
\end{figure}

In this section, we introduce \emph{FaceInstruct-1M}, a dataset designed for instruction-tuning MLLMs on face-centric tasks. As shown in \cref{tab:data_comparison}, it provides instructions and descriptions for expression recognition, AU detection, deepfake detection, attribute detection, and age estimation. Unlike existing datasets \cite{jiang2020dfew, wang2022ferv39k, disfa_dataset, attr_liu_celeba, roessler2019faceforensicspp} that focus on individual tasks, \emph{FaceInstruct-1M} spans multiple domains. While prior instruction-tuning \cite{Xie2024EmoVIT, cheng2024emotionllama} and captioning \cite{liu_mafw_2022, lian2023explainable_emer} datasets annotate entire images/videos, including background context and audio, ours is specifically curated for face analysis. Though FABA-Instruct \cite{li2024facialfaba} is face-specific, it is limited to images, has fewer samples, and covers only two tasks. To the best of our knowledge, \emph{FaceInstruct-1M} is the first large-scale instruction-tuning dataset for face analysis, spanning multiple tasks with both images and videos.

\subsection{Annotation Pipeline}
\label{subsec:annotation_pipeline}
\emph{FaceInstruct-1M} is constructed using images, videos, and annotations from existing task-specific datasets as mentioned in \cref{tab:constituent_datasets} (see \cref{sec:traditional_data_desctiption} for details).
Since, we are only interested in face, we apply some data preprocessing to crop the faces in videos and only use such samples in which there is a single subject present during the entire duration of the video (\cref{subsec:data_preprocessing} for details).

We found that existing manual annotations in various face datasets, despite being categorical or containing minimal information, can significantly aid in guiding the automatic generation of descriptions that align with the data (\cref{subsubsec:data_ablation}). To generate video descriptions for specific tasks, we provide the face video and its categorical label, along with carefully designed annotation instructions, to Gemini 1.5 Flash \cite{geminiteam2024gemini15unlockingmultimodal}. Gemini was selected among open and commercial models due to its superior performance, faster inference, higher rate limits and low cost. As an example, our annotation instruction for the facial expression recognition task follows this structure: \textit{`The given video has \dots emotion. Your task is to reason why the emotion is tagged as \dots. Focus specifically on the facial expression of the subject in the video and describe how it varies. \dots \{output format instructions\}'}. Following FABA-Instruct \cite{li2024facialfaba}, we associate each description and video/image pair with a corresponding set of instructions for instruction-tuning. To achieve this, we carefully curate a collection of 100 hand-crafted instructions for each task and randomly pair them with the generated descriptions. Refer to \cref{sec:face_instruct_1m_additional_details} for more details.


\begin{table}[]
\centering
\resizebox{\linewidth}{!}{%
\begin{tabular}{c|c|cccc}
\hline \hline
\rowcolor[HTML]{C0C0C0} 
\multicolumn{1}{c|}{\cellcolor[HTML]{C0C0C0}} & \cellcolor[HTML]{C0C0C0} & \multicolumn{4}{c}{\cellcolor[HTML]{C0C0C0}\textbf{Mean GPT Rating (1-10)}} \\ \cline{3-6}
\rowcolor[HTML]{C0C0C0} 
\multicolumn{1}{c|}{\multirow{-2}{*}{\cellcolor[HTML]{C0C0C0}\textbf{Task}}} & \multirow{-2}{*}{\cellcolor[HTML]{C0C0C0}\textbf{Datasets}} & \textbf{\begin{tabular}[c]{@{}c@{}}Label\\ Acc.\end{tabular}} & \textbf{\begin{tabular}[c]{@{}c@{}}Desc.-Vid.\\ Consistency\end{tabular}} & \textbf{\begin{tabular}[c]{@{}c@{}}Desc.-Lab.\\ Consistency\end{tabular}} & \textbf{Overall} \\ \hline
\multicolumn{6}{c}{\emph{FaceInstruct-1M}} \\ \hline
\begin{tabular}[c]{@{}l@{}}Expression\end{tabular} & \begin{tabular}[c]{@{}c@{}}DFEW \cite{jiang2020dfew}, MAFW \cite{liu_mafw_2022}, \\ FERV39k \cite{wang2022ferv39k} Crema-D \cite{cremad_dataset}, \\AffectNet \cite{affectnet}, RAF-DB \cite{rafdb} \end{tabular} & 8.81 & 8.74 & 8.79 & 8.71 \\
\begin{tabular}[c]{@{}l@{}}AU\end{tabular} & DISFA \cite{disfa_dataset}, BP4D \cite{ZHANG2014692_bp4d_dataset} & 8.17 & 8.23 & 7.62 & 7.99 \\
\begin{tabular}[c]{@{}l@{}}Attribute \end{tabular} & CelebA \cite{attr_liu_celeba} & 8.79 & 8.75 & 8.52 & 8.69 \\
\begin{tabular}[c]{@{}l@{}}Age \end{tabular} & \begin{tabular}[c]{@{}c@{}}MORPH II \cite{morph_ii_dataset}, UTK Face \cite{zhifei2017cvpr_utkface}\end{tabular} & 8.63 & 7.92 & 7.98 & 7.99 \\
\begin{tabular}[c]{@{}l@{}}Deepfake*\end{tabular} & \begin{tabular}[c]{@{}c@{}}FF++ \cite{roessler2019faceforensicspp}, Fake AV Celeb \cite{khalid2021fakeavceleb}\end{tabular} & 8.84 & 8.13 & 8.60 & 8.20 \\ \hline 
\multicolumn{6}{c}{FABAInstruct \cite{li2024facialfaba}} \\ \hline
\begin{tabular}[c]{@{}l@{}}Expression\end{tabular} & \begin{tabular}[c]{@{}c@{}} AffectNet \cite{affectnet}\end{tabular} & 8.79 & 6.44 & 6.32 & 6.22 \\ \hline \hline
\end{tabular}%
}
\caption{Datasets used to construct the \emph{FaceInstruct-1M} and GPT \cite{gpt4omini} ratings for different aspects of the data. Ratings for FABAInstruct \cite{li2024facialfaba} are calculated as a reference. *For deepfake detection, we annotate a set of real videos from DFEW \cite{jiang2020dfew}, MAFW \cite{liu_mafw_2022}, FERV39k \cite{wang2022ferv39k} as control samples. }
\label{tab:constituent_datasets}
\vspace{-1em}
\end{table}

\subsection{Dataset Rating and Filtering}
\label{subsec:dataset_filtering}
\cref{tab:constituent_datasets} lists the constituent datasets of \emph{FaceInstruct-1M}, where we observe that some datasets \cite{jiang2020dfew, liu_mafw_2022, wang2022ferv39k, attr_liu_celeba} are not exclusively face-specific and therefore include background context and audio. Consequently, the manual annotations in these datasets may be influenced by these additional contextual elements rather than solely by facial expressions or facial features. Furthermore, since our annotations are generated using an LLM, it is crucial to perform a sanity check and apply filtering to ensure high data quality.

We employ a GPT-assisted rating pipeline for data filtering, which takes the video/image, manually annotated label and Gemini-generated description and rate them based on instructions. We use GPT-4o mini \cite{gpt4omini} for rating according to (i) accuracy of the manually annotated label w.r.t. the face video, (ii) consistency of the generated description with the face video, (iii) consistency of the generated description with the manually annotated label, and (iv) overall quality of the sample based on resolution, visibility of the face, etc. \cref{tab:constituent_datasets} summarizes the ratings by GPT-4o-mini \cite{gpt4omini}. To calibrate the ratings, we also perform GPT-evaluation of FABAInstruct \cite{li2024facialfaba}. \emph{FaceInstruct-1M} consistently achieves high ratings from GPT, demonstrating high data quality. Samples with an overall rating less than or equal to six are removed from \emph{FaceInstruct-1M} for training, resulting in the removal of roughly 7\% of the dataset.

\subsection{Qualitative analysis}


\cref{fig:qual_samples} presents samples from the \emph{FaceInstruct-1M} dataset across different tasks. Notably, for video inputs, the descriptions capture subtle variations in various facial attributes and justify the responses to the given instructions based on these observations. It is also important to highlight that the dataset encompasses tasks reliant on muscle movements (e.g., expression recognition, AU detection, and deepfake detection), facial attributes (including age), and overall video quality (deepfake detection). This diverse coverage makes \emph{FaceInstruct-1M} well-suited for learning a broad range of facial features across different categories.

\begin{figure*}
    \centering
    \includegraphics[width=0.82\linewidth]{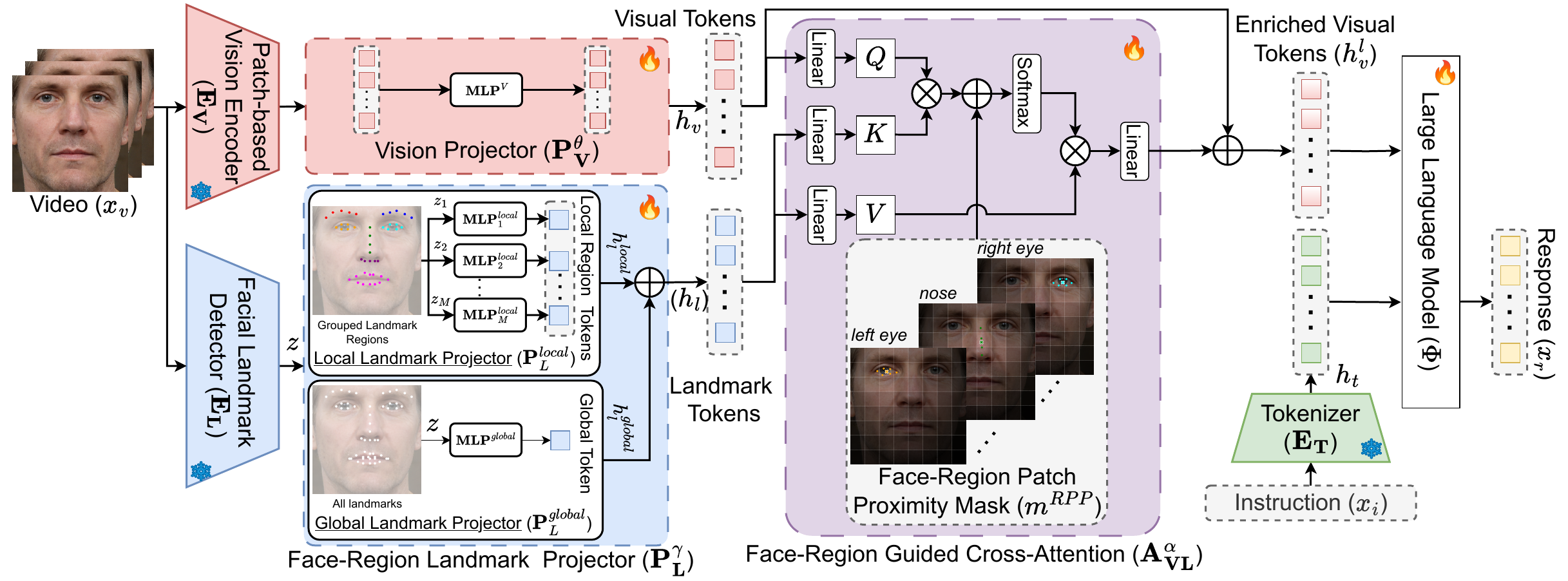}
    \vspace{-1em}
    \caption{Proposed Face-LLaVA architecture. We group landmark points into different face regions and project them through a face-region landmark projector. The visual tokens are enriched by the landmark tokens through cross-attention and then passed as input to the LLM.}
    \label{fig:face_llava_main_figure}
    \vspace{-1em}
\end{figure*}

\subsection{Evaluation}
\label{subsec:dataset_evaluation}

Since an MLLM trained on our dataset generates natural language, its evaluation must consider both (i) metrics for face analysis, such as average recall or accuracy (\cref{subsec:traditional_metrics}), and (ii) the reasoning capabilities w.r.t. instructions. Similar to FABAInstruct \cite{li2024facialfaba}, we assess traditional metrics using synonym matching (refer \cref{subsec:synonyms_used}) for expression recognition, attribute detection, and deepfake detection, while for AU detection and age estimation, we extract numerical values directly from the generated text. To ensure accuracy, negative statements are removed before synonym matching or string parsing. For tasks requiring a single prediction label, majority voting is applied. Once categorical or numerical labels are extracted from the descriptions, task-specific metrics are calculated.

For evaluating the reasoning capabilities of MLLMs on face tasks, traditional text generation metrics such as BLEU \cite{papineni-etal-2002-bleu} or ROUGE \cite{lin-2004-rouge} can be used to match the generated description with the ground truth. However, such metrics are based on crude n-gram matches rather than in-depth semantic analysis of the generated description. Hence, we evaluate the reasoning capabilities using GPT4o-mini \cite{gpt4omini} on the following aspects similar to \cref{subsec:dataset_filtering} - (i) consistency or overlap of the given reasoning with video, (ii) consistency or overlap of the given reasoning with ground truth label, and (iii) overall completeness of the reasoning to support the ground truth label w.r.t the video.

Furthermore, we construct a small test set for \emph{FaceInstruct-1M} consisting of 500 samples for each of the five tasks using DFEW \cite{jiang2020dfew}, DISFA \cite{disfa_dataset}, CelebA \cite{attr_liu_celeba}, UTKFace \cite{zhifei2017cvpr_utkface} and FaceForensics++ \cite{roessler2019faceforensicspp}. We start with the samples that are top-rated by GPT and then refine the set so that the class distribution of test samples for each task match the original dataset. Such a small and carefully crafted set also enables other holistic evaluations of reasoning, such as human evaluation, which is expensive.




\section{Face-LLaVA}
\label{sec:face_llava_method}

\cref{fig:face_llava_main_figure} illustrates the Face-LLaVA architecture for instruction-tuning on face images and videos. Similar to prior MLLMs \cite{liu2023llava,lin-etal-2024-videollava}, it consists of a patch-based vision encoder $\mathbf{E_V}$ to encode input image (or video), a tokenizer $\mathbf{E_T}$ to encode the text instruction and a large language model decoder $\Phi$ to generate responses according to the instruction $x_i$. Inspired by Lin et al. \cite{zhu2024languagebind}, we use LanguageBind \cite{zhu2024languagebind} vision encoders for video and image and learn a joint vision projector $\mathbf{P}_\mathbf{V}^{\theta}$ from the visual token space to the language token space. Specifically, given a visual input $x_v$, we get visual tokens $h_v \in \mathbb{R}^{T\times N\times d}$ as follows,
\begin{equation}
    h_v = \mathbf{P}_\mathbf{V}^{\theta}(\mathbf{E_V}(x_v))
\end{equation}
where $T$ is the number of video frames (1 for images), $N$ is the number of visual tokens from the patch-based vision encoder $\mathbf{E_V}$ (256 for LanguageBind), and $d$ is the dimensionality of hidden representation of the LLM. 

Since we are dealing with faces, the visual information should also incorporate features specifically important for face processing which can be obtained from a \emph{face-expert} model $\mathbf{E_L}$. In Face-LLaVA, we use a landmark detector as our \emph{face-expert} model to extract face-specific features from the visual input $x_v$. We also experimented with a face-parsing model as the expert, but that led to extra computing due to pixel-wise information and no returns (see \cref{subsec:ablation_architecture}). Formally, we obtain the normalized 2-D landmark coordinates, $z = \mathbf{E_L}(x_v)$ where $z \in \mathbb{R}^{T\times L \times 2}$. These expert features are then fed to the proposed novel \emph{Face-Region Landmark Projector} and \emph{Face-Region Guided Cross-Attention} modules.

\subsection{Face-Region Landmark Projector (FRLP)}

To project the facial landmark features to the token space, one can simply flatten the $L$ facial landmarks into the LLM's $d$-dimensional token space by projection \cite{li2024facialfaba}. However, this maps the entire face landmark features into a single token without preserving information about what individual landmark points mean. Moreover, for a lot of complex face-perception tasks, humans tend to look at a group of facial muscles or regions, such as eyes or lips, to make informed decisions. For example, to detect an expression of disgust, visual features around the nose to capture wrinkling are important. Therefore, we propose a \emph{Face-region Landmark Projector} to group landmarks into different face regions and then project them into individual face-region tokens through separate MLPs. Specifically, we use two sub-modules to tokenize the landmark features. 

The \emph{local landmark projector} $\mathbf{P}_L^{local}$ generates $M$ tokens each corresponding to a separate face-region, 
\begin{equation}
    h_l^{local} = \{\mathbf{MLP}^{local}_i(z_i) \hspace{2mm} \forall \hspace{2mm} i \in \{1 \dots M\} \}
\end{equation}

where $z_i \in \mathbb{R}^{T\times L_i \times 2}$ are the $L_i$ 2-D landmark points corresponding to the $i^{th}$ face region. We use a total of 9 landmark groups corresponding to the following face regions - \emph{face boundary, left/right eye, left/right brow, nose, nostril, lips and teeth}. 

Similar to EmoLA \cite{li2024facialfaba}, we also project all the $L$ landmark points through a \emph{global landmark projector} $\mathbf{P}_L^{global}$ to get a single global landmark token for each frame which can capture the overall facial structure and dynamics beyond individual regions. 
\begin{equation}
    h_l^{global} = \mathbf{MLP}^{global}(z)
\end{equation}
and the final landmark tokens $h_l$ are given by,
\begin{equation}
    h_l = h_l^{global} + h_l^{local}
\end{equation}
where the global token $h_l^{global}$ is broadcasted in the token dimension. By combining the local region and the global landmark tokens, our approach ensures that both local and holistic facial features are effectively modeled. To keep the extra compute to a minimum, we use single-layer MLPs for all the landmark projections.

\subsection{Face-Region Guided Cross-Attention(FRGCA)}

We propose to use the landmark tokens $h_l$ generated by the FRLP module via cross-attention with the visual tokens $h_v$. Such an architecture poses two benefits - (i) using cross-attention enables weighing visual tokens that are closer to salient face regions and are highly likely to be used by the downstream face processing tasks, and (ii) as opposed to EmoLA \cite{li2024facialfaba}, the $M$ landmark tokens per frame do not have to be appended to the visual tokens and passed as input to the LLM thus saving the LLM's context window. Formally, we obtain the key ($K\in \mathbb{R}^{M \times d_{attn}}$) and value ($V \in \mathbb{R}^{M \times d_{attn}}$) vectors using \(h_l\), and the query ($Q \in \mathbb{R}^{N \times d_{attn}}$) vector using $h_v$ by passing through separate linear layers. Note that we have dropped the time dimension $T$ for simplicity.

To further enforce the attention weights on face regions, we compute a \emph{face-region patch proximity} (RPP) mask, which is inversely proportional to the pairwise 2-D distances between the centroids of the visual patches used as inputs by $\mathbf{E_V}$ and the centroids of the different face regions as shown in Figure \cref{fig:face_llava_main_figure}. Mathematically, the elements of the mask $m^{RPP}$ are given by
\begin{equation}
    m^{RPP}_{ji} = - || centroid(z_i) - centroid(h_{v,j}) ||_2
\end{equation}
where $centroid(z_i)$ is the centroid of $i^{th}$ face region, $centroid(h_{v,j})$ is the centroid of the $j^{th}$ visual patch associated with the visual token $h_{v,j}$, and $m^{RPP} \in \mathbb{R}^{N \times M}$.

The RPP mask is used to guide the attention weights before the softmax layer. The entire FRGCA module $\mathbf{A}_\mathbf{VL}^{\alpha}$ can be summarized as, 
\begin{equation}
    h_v^{l} = Linear\left(Softmax \left( \frac{Q K^T}{\sqrt{d_{attn}}} + m^{RPP} \right) V\right) + h_v
\end{equation}

\begin{table*}[]

\parbox{0.67\linewidth}{%
\centering
\resizebox{\linewidth}{!}{
\begin{tabular}{lcclcclc}
\hline \hline
\rowcolor[HTML]{C0C0C0} 
\multicolumn{3}{c|}{\cellcolor[HTML]{C0C0C0}\textbf{DFEW \cite{jiang2020dfew}}}  & \multicolumn{3}{c|}{\cellcolor[HTML]{C0C0C0}\textbf{Crema-D \cite{cremad_dataset}}} & \multicolumn{2}{c}{\cellcolor[HTML]{C0C0C0}\textbf{RAF-DB \cite{rafdb}}} \\ \hline
\rowcolor[HTML]{C0C0C0} 
\multicolumn{1}{c}{\cellcolor[HTML]{C0C0C0}\textbf{Method}} & \textbf{UAR $\uparrow$} & \multicolumn{1}{c|}{\cellcolor[HTML]{C0C0C0}\textbf{WAR $\uparrow$}} & \multicolumn{1}{c}{\cellcolor[HTML]{C0C0C0}\textbf{Method}} & \textbf{UAR $\uparrow$} & \multicolumn{1}{c|}{\cellcolor[HTML]{C0C0C0}\textbf{WAR $\uparrow$}} & \multicolumn{1}{c}{\cellcolor[HTML]{C0C0C0}\textbf{Method}} & \textbf{Acc. $\uparrow$} \\ \hline
\multicolumn{8}{c}{\cellcolor[HTML]{FFFFFF}\textit{Closed - source models}} \\ \hline
GPT4o-mini \cite{gpt4omini} & 0.426 & \multicolumn{1}{c|}{0.518} & GPT4o-mini \cite{gpt4omini} & 0.410 & \multicolumn{1}{c|}{0.486} & GPT4o-mini \cite{gpt4omini} & 0.758 \\
Gemini-1.5F \cite{geminiteam2024gemini15unlockingmultimodal} & {0.433} & \multicolumn{1}{c|}{0.481} & Gemini-1.5F \cite{geminiteam2024gemini15unlockingmultimodal} & 0.465 & \multicolumn{1}{c|}{0.635} & Gemini-1.5F \cite{geminiteam2024gemini15unlockingmultimodal} & 0.685 \\ \hline
\rowcolor[HTML]{FFFFFF} 
\multicolumn{8}{c}{\cellcolor[HTML]{FFFFFF}\textit{Zero-shot}} \\ \hline
Vid.LLaMA 3 \cite{damonlpsg2025videollama3}  & 0.286 & \multicolumn{1}{c|}{0.305} & Vid.LLaMA 3 \cite{damonlpsg2025videollama3} & 0.397 & \multicolumn{1}{c|}{0.546} & Vid.LLaMA 3 \cite{damonlpsg2025videollama3} & 0.671 \\
Qwen 2.5 \cite{qwen25vl}  & 0.293 & \multicolumn{1}{c|}{0.399} & Qwen 2.5 \cite{qwen25vl} & 0.395 & \multicolumn{1}{c|}{0.566} & Qwen 2.5 \cite{qwen25vl} & 0.526 \\
Vid.-LLaVA \cite{lin-etal-2024-videollava} & 0.220 & \multicolumn{1}{c|}{0.326} & Vid.-LLaVA \cite{lin-etal-2024-videollava} & 0.367 & \multicolumn{1}{c|}{0.557} & Vid.-LLaVA \cite{lin-etal-2024-videollava} & 0.545 \\
LLaVA-Vid. \cite{zhang2024llava_video} & 0.375 & \multicolumn{1}{c|}{0.498} & LLaVA-Vid. \cite{zhang2024llava_video} & 0.478 & \multicolumn{1}{c|}{0.618} & LLaVA-OV \cite{li2024llava_onevision} & 0.700 \\
EmoLA\textsuperscript{*} \cite{li2024facialfaba} & 0.346 &  \multicolumn{1}{c|}{0.449} & EmoLA\textsuperscript{*} \cite{li2024facialfaba} & 0.431 & \multicolumn{1}{c|}{0.618} & EmoLA \cite{li2024facialfaba} & 0.741 \\
\begin{tabular}[c]{@{}l@{}}Emotion LLaMA\textsuperscript{\textdagger} \cite{cheng2024emotionllama}\end{tabular} & 0.456 & \multicolumn{1}{c|}{\textbf{0.594}} & \begin{tabular}[c]{@{}l@{}}Emotion LLaMA \cite{cheng2024emotionllama}\end{tabular} & 0.225 & \multicolumn{1}{c|}{0.308} & & \\
\begin{tabular}[c]{@{}l@{}}Emotion LLaMA\textsuperscript{\textdaggerdbl} \cite{cheng2024emotionllama}\end{tabular} & 0.302 & \multicolumn{1}{c|}{0.378} & & & \multicolumn{1}{c|}{} & & \\
\textbf{Face-LLaVA (Ours)} & \textbf{0.469} & \multicolumn{1}{c|}{\underline{0.564}} & \textbf{Face-LLaVA (Ours)} & \textbf{0.582} & \multicolumn{1}{c|}{\textbf{0.681}} & \textbf{Face-LLaVA (Ours)} & \textbf{0.780} \\ \hline
\multicolumn{8}{c}{\cellcolor[HTML]{FFFFFF}\textit{Fine-tuned}} \\ \hline
EC-STFL \cite{jiang2020dfew} & 0.454 & \multicolumn{1}{c|}{0.565} & Lei et al. \cite{cremad_lei} & 0.645 & \multicolumn{1}{c|}{0.648} & RUL \cite{rul_rafdb} & 0.890 \\
Former-DFER \cite{zhao2021former} & 0.537 & \multicolumn{1}{c|}{0.657} & PTH-Net \cite{cremad_pthnet} & 0.699 & \multicolumn{1}{c|}{0.700} & EAC \cite{eac_rafdb_emotion} & 0.909 \\
GCA+IAL \cite{Li2023-dfew_ial} & 0.557 & \multicolumn{1}{c|}{0.692} & MAE-DFER \cite{dfew_mae_dfer} & 0.773 & \multicolumn{1}{c|}{0.774} & TransFER \cite{transfer_rafdb} & 0.909 \\
M3DFEL  \cite{dfew_m3dfel} & 0.561 & \multicolumn{1}{c|}{0.693} & MTL-ER* \cite{mtler_cremad} & 0.745 & \multicolumn{1}{c|}{0.756} & Xue et al. \cite{rafdb_apvit} & 0.920 \\
MAE-DFER  \cite{dfew_mae_dfer} & 0.634 & \multicolumn{1}{c|}{0.744} & MT-Former* \cite{mtformer} & 0.793 & \multicolumn{1}{c|}{0.807} & POSTERv2 \cite{mao2023postersimplerstrongerfacial} & \textbf{0.922} \\
S2D \cite{dfew_s2d} & 0.618 & \multicolumn{1}{c|}{0.760} & MTCAE-DFER \cite{xiang2024mtcaedfer} & \textbf{0.847} & \multicolumn{1}{c|}{\textbf{0.850}} & EmoLA \cite{li2024facialfaba} & 0.921 \\
EMO-LLaMA \cite{xing2024emollama} & 0.602 & \multicolumn{1}{c|}{0.659} &  &  & \multicolumn{1}{c|}{} &  &  \\
Emotion-LLaMA  \cite{cheng2024emotionllama} & \textbf{0.642} & \multicolumn{1}{c|}{\textbf{0.771}} &  &  & \multicolumn{1}{c|}{} &  &  \\
\textbf{Face-LLaVA (Ours)} & \underline{0.625} &  \multicolumn{1}{c|}{\underline{0.745}} & \textbf{Face-LLaVA (Ours)} & \underline{0.798} & \multicolumn{1}{c|}{\underline{0.813}} & \textbf{Face-LLaVA (Ours)} & {\ul 0.921} \\ \hline \hline
\end{tabular}%
}
\vspace{-5pt}
\caption{Comparison of the proposed approach with recent MLLMs and supervised techniques for emotion recognition on the DFEW \cite{jiang2020dfew}, Crema-D \cite{cremad_dataset} and RAF-DB \cite{rafdb} dataset. *:Only using middle video frame, \textdagger: Results taken from the paper \cite{cheng2024emotionllama}, \textdaggerdbl: Results computed after running inference code on face-cropped video.}
\label{tab:res_dfew_cremad_rafdb}
}
\hfill
\parbox{0.315\linewidth}{%
\centering
\resizebox{\linewidth}{!}{
\begin{tabular}{l|cc}
\hline \hline
\rowcolor[HTML]{C0C0C0} 
\cellcolor[HTML]{C0C0C0} & \multicolumn{2}{c}{\cellcolor[HTML]{C0C0C0}\textbf{Average F1 $\uparrow$}} \\ \cline{2-3} 
\rowcolor[HTML]{C0C0C0} 
\multirow{-2}{*}{\cellcolor[HTML]{C0C0C0}\textbf{Method}} & \textbf{DISFA \cite{disfa_dataset}} & \textbf{BP4D \cite{ZHANG2014692_bp4d_dataset}} \\ \hline
\multicolumn{3}{c}{\textit{Closed-source models}} \\ \hline
GPT4o-mini \cite{gpt4omini} & 0.429 & 0.496 \\
Gemini-1.5F \cite{geminiteam2024gemini15unlockingmultimodal} & 0.515 & 0.532 \\ \hline
\multicolumn{3}{c}{\textit{Zero-shot}} \\ \hline
VideoLLaMA 3 \cite{damonlpsg2025videollama3} & 0.374 & 0.458 \\
Qwen 2.5 VL \cite{qwen25vl} & 0.431 & 0.467 \\
Video-LLaVA \cite{lin-etal-2024-videollava} & 0.442 & 0.445 \\
LLaVA-OneVision \cite{li2024llava_onevision} & 0.280 & 0.439 \\
EmoLA \cite{li2024facialfaba} & 0.418 & 0.407 \\
\textbf{Face-LLaVA (Ours)} & \textbf{0.553} & \textbf{0.495} \\ \hline
\multicolumn{3}{c}{\textit{Fine-tuned}} \\ \hline
ATCM \cite{jacob2021facial} & 0.615 & 0.642 \\
ReCoT \cite{Li_2023_BMVC_recot} & 0.626 & 0.648 \\
KS \cite{knowledge_spreader_ks_au} & 0.628 &  \\
ME-GraphAU \cite{luo2022graphau} & 0.631 & 0.655 \\
$J\hat{A}A$-Net \cite{Shao2020JANetJF} & 0.635 & 0.624 \\
PIAP-DF \cite{piap_df} & 0.638 & 0.641 \\
VL-FAU \cite{Ge2024-yp-vlfau} & 0.665 & \textbf{0.658} \\
AU-LLaVA \cite{hu2024unifiedfacialactionunit_aullava} & 0.525 & 0.603 \\
EmoLA \cite{li2024facialfaba} & 0.651 & 0.642 \\
\textbf{Face-LLaVA (Ours)} & \textbf{0.729} & \textbf{0.658} \\ \hline \hline
\end{tabular}%
}
\vspace{-5pt}
\caption{Comparison of the proposed approach with recent MLLMs and supervised techniques on average F1 over the 8 AUs of DISFA \cite{disfa_dataset} and 12 AUs of BP4D \cite{ZHANG2014692_bp4d_dataset}.
\label{tab:res_disfa_bp4d}
}
}
\vspace{-1em}
\end{table*}

\subsection{Training}

We train the Face-LLaVA model in multiple stages so that the visual and landmark tokens can be aligned with the language tokens. We initialize the vision encoder($\mathbf{E_V}$), vision projector ($\mathbf{P}_\mathbf{V}^{\theta}$), tokenizer ($\mathbf{E_T}$) and the LLM ($\Phi$) from the pretrained weights of Video-LLaVA \cite{lin-etal-2024-videollava}. For landmark detection, we use pretrained weights from FAN \cite{bulat2017far}. 

\noindent\textbf{Face-Region Pretraining.} In this stage, we only train the FRLP module $\mathbf{P}_\mathbf{L}^{\gamma}$ and the FRGCA module $\mathbf{A}_\mathbf{VL}^{\alpha}$ and keep all the other weights frozen, hence, the trainable parameters are given by $\Theta = \{\gamma, \alpha \}$. This ensures that the newly initialized modules generating landmarks tokens get aligned with the visual and the language tokens generated by $\mathbf{P}_\mathbf{V}^{\theta}$ and $\mathbf{E_T}$ respectively. 

\noindent\textbf{Finetuning.} In this stage, we train the vision projector $\mathbf{P}_\mathbf{V}^{\theta}$ and the LLM model $\Phi$ along with $\mathbf{P}_\mathbf{L}^{\gamma}$ and $\mathbf{A}_\mathbf{VL}^{\alpha}$ with a lower learning rate. This stage jointly finetunes the entire model to further improve the instruction following capabilities of the model. The trainable parameters here are given by $\Theta = \{\gamma, \alpha , \theta, \phi \}$. Note that since we have a good number of training samples in \emph{FaceInstruct-1M}, we train all the parameters $\phi$ of the LLM rather than using LoRA \cite{hu2022lora}.

For both stages, the model is trained in an autoregressive fashion by maximizing the likelihood of the response $x_r$:
\begin{equation}
    P(x_r|h_v^l, h_t) = \prod_{i=1}^{\mathcal{L}} P_\Theta (x_r^i | h_v^l, h_t, x_r^{[1:i-1]})
\end{equation}
where $x_r^i$ is the $i^{th}$ response token, $\mathcal{L}$ is the length of the response $x_r$, and $x_r^{[1:i-1]}$ are the response tokens generated before the $i^{th}$ token. Learning rates for the pretraining and finetuning stages are 1e-4 and 2e-5 respectively, and all the models are trained for one epoch (see \cref{sec:supp_implementation_details} for details).

\section{Experiments}
\label{sec:experiments}

\subsection{Datasets and Evaluation Protocol}

\noindent\textbf{Perception.} \label{subsec:evaluation_protocol} We perform extensive evaluations of the proposed approach on nine benchmarks including DFEW \cite{jiang2020dfew}, Crema-D \cite{cremad_dataset}, and RAF-DB \cite{rafdb} for facial expression recognition; DISFA \cite{disfa_dataset} and BP4D \cite{ZHANG2014692_bp4d_dataset} for action unit detection; FaceForensics++ \cite{roessler2019faceforensicspp} for deepfake detection; MORPH II \cite{morph_ii_dataset} and UTK Face \cite{zhifei2017cvpr_utkface} for age estimation and CelebA \cite{attr_liu_celeba} for facial attribute detection (\cref{sec:traditional_data_desctiption,subsec:traditional_metrics} for details). Since these datasets are also used to construct the \emph{FaceInstruct-1M} dataset, we report performance on these datasets in the following two settings. In the \emph{zero-shot} setting, we remove the entire task-specific dataset from \emph{FaceInstruct-1M} for tuning our model, and in the \emph{fine-tuned} setting we finetune the \emph{zero-shot} model according to the official splits or training protocol of the benchmark (\cref{subsec:traditional_metrics}).

\noindent\textbf{Reasoning.} As mentioned in \cref{subsec:dataset_evaluation}, we perform reasoning evaluation using GPT4o-mini \cite{gpt4omini} on \emph{FaceInstruct-1M} \emph{test} set consisting of 500 samples per task. We do not use the existing EMER \cite{lian2023explainable_emer} benchmark for reasoning evaluation because we want to assess the reasoning based on facial expressions and not the background context or audio.

\begin{table}[]
\centering
\resizebox{\linewidth}{!}{%
\begin{tabular}{lcclclc}
\hline \hline
\rowcolor[HTML]{C0C0C0} 
\multicolumn{3}{c|}{\cellcolor[HTML]{C0C0C0}\textbf{Age Estimation (MAE $\downarrow$)}} & \multicolumn{2}{c|}{\cellcolor[HTML]{C0C0C0}\textbf{Face Attribute (mAcc. $\uparrow$)}} & \multicolumn{2}{c}{\cellcolor[HTML]{C0C0C0}\textbf{DeepFake Det. (Acc. $\uparrow$)}} \\ \hline
\rowcolor[HTML]{C0C0C0} 
\multicolumn{1}{c}{\cellcolor[HTML]{C0C0C0}\textbf{Method}} & \textbf{M \cite{morph_ii_dataset}} & \multicolumn{1}{c|}{\cellcolor[HTML]{C0C0C0}\textbf{U\cite{zhifei2017cvpr_utkface}}} & \multicolumn{1}{c}{\cellcolor[HTML]{C0C0C0}\textbf{Method}} & \multicolumn{1}{c|}{\cellcolor[HTML]{C0C0C0}\textbf{CA \cite{attr_liu_celeba}}} & \multicolumn{1}{c}{\cellcolor[HTML]{C0C0C0}\textbf{Method}} & \textbf{FF \cite{roessler2019faceforensicspp}} \\ \hline
\multicolumn{7}{c}{\textit{Closed-source models}} \\ \hline
GPT4o-m \cite{gpt4omini} & 4.09 & \multicolumn{1}{c|}{5.04} & GPT4o-m \cite{gpt4omini} & \multicolumn{1}{c|}{0.780} & GPT4o-m \cite{gpt4omini} & 0.807 \\
Gem.-1.5F \cite{geminiteam2024gemini15unlockingmultimodal} & 4.78 & \multicolumn{1}{c|}{6.13} & Gem.-1.5F \cite{geminiteam2024gemini15unlockingmultimodal} & \multicolumn{1}{c|}{0.814} & Gem.-1.5F \cite{geminiteam2024gemini15unlockingmultimodal} & 0.770 \\ \hline
\multicolumn{7}{c}{\textit{Zero-shot}} \\ \hline
V-LLaMA3 \cite{damonlpsg2025videollama3} & 6.98 & \multicolumn{1}{c|}{6.91} & V-LLaMA3 \cite{damonlpsg2025videollama3} & \multicolumn{1}{c|}{0.813} & V-LLaMA3 \cite{damonlpsg2025videollama3} & 0.793 \\
Qwen 2.5 \cite{qwen25vl} & 6.09 & \multicolumn{1}{c|}{5.25} & Qwen 2.5 \cite{qwen25vl} & \multicolumn{1}{c|}{0.786} & Qwen 2.5 \cite{qwen25vl} & 0.653 \\
V-LLaVA \cite{lin-etal-2024-videollava} & 6.75 & \multicolumn{1}{c|}{5.89} & V-LLaVA \cite{lin-etal-2024-videollava} & \multicolumn{1}{c|}{0.795} & V-LLaVA \cite{lin-etal-2024-videollava} & 0.697 \\
LLaVA-OV \cite{li2024llava_onevision} & 6.33 & \multicolumn{1}{c|}{6.87} & LLaVA-OV \cite{li2024llava_onevision} & \multicolumn{1}{c|}{0.805} & LLaVA-V \cite{zhang2024llava_video} & 0.751 \\
\textbf{Face-LLaVA} &  \textbf{3.34} & \multicolumn{1}{c|}{\textbf{4.89}} & \textbf{Face-LLaVA} & \multicolumn{1}{c|}{\textbf{0.868}} & \textbf{Face-LLaVA} & \textbf{0.845} \\ \hline
\multicolumn{7}{c}{\textit{Fine-tuned}} \\ \hline
PML \cite{pml_age} & 2.15 & \multicolumn{1}{c|}{-} & Liu et al. \cite{attr_liu_celeba} & \multicolumn{1}{c|}{0.873} & MesoNet  \cite{mesonet} & \multicolumn{1}{c}{0.705} \\
Berg et al. \cite{berg2021deep} & - & \multicolumn{1}{c|}{4.55} & MOON \cite{moon_age} & \multicolumn{1}{c|}{0.909} & Xception  \cite{xception_net} & \multicolumn{1}{c}{0.869} \\
DLDL-v2 \cite{gao2021learningexpectationlabeldistribution} & 1.97 & \multicolumn{1}{c|}{4.42} & SwinFace \cite{swinface} & \multicolumn{1}{c|}{0.913} & MARLIN \cite{marlin} & \multicolumn{1}{c}{0.894} \\
MWR \cite{age_mwr} & 2.00 & \multicolumn{1}{c|}{4.37} & Faceptor \cite{faceptor} & \multicolumn{1}{c|}{0.914} & M2TR \cite{m2tr_deepfake} & \multicolumn{1}{c}{0.929} \\
Faceptor \cite{faceptor} & \textbf{1.96} & \multicolumn{1}{c|}{4.10} & DMM-CNN \cite{dmm_cnn} & \multicolumn{1}{c|}{\textbf{0.917}} & F3-Net \cite{f3_net} & \textbf{0.930} \\
\textbf{Face-LLaVA} & {\ul 2.02} & \multicolumn{1}{c|}{\textbf{4.06}} & \textbf{Face-LLaVA} & \multicolumn{1}{c|}{{\ul 0.901}} & \textbf{Face-LLaVA} & \multicolumn{1}{c}{{\ul 0.888}} \\ \hline \hline
\end{tabular}%
}
\vspace{-1em}
\caption{Comparison of the proposed approach with recent MLLMs and supervised techniques on the age estimation, face attribute detection and deepfake detection tasks on different datasets \cite{morph_ii_dataset, zhifei2017cvpr_utkface,attr_liu_celeba, roessler2019faceforensicspp}. M: MORPH II \cite{morph_ii_dataset}, U: UTKFace \cite{zhifei2017cvpr_utkface}, CA: CelebA \cite{attr_liu_celeba}, FF: FaceForensics++ \cite{roessler2019faceforensicspp}.}
\label{tab:age_celeba_dfd}
\vspace{-2em}
\end{table}

\subsection{Evaluation on traditional tasks}
\label{subsec:traditional_evaluation_results}

We assess Face-LLaVA against multiple baselines under both \emph{zero-shot} and \emph{fine-tuned} settings, following the evaluation protocol detailed in \cref{subsec:evaluation_protocol}. Unless otherwise, we apply the official inference code to face-cropped inputs for \emph{zero-shot} baselines (\cref{subsec:implementation_baselines}) and report the officially published results for \emph{fine-tuned} baselines. Additionally, for GPT4o-mini \cite{gpt4omini} and Gemini 1.5 Flash \cite{geminiteam2024gemini15unlockingmultimodal}, official APIs are used to perform batch inference.

\noindent\textbf{Facial Expression Recognition.} \cref{tab:res_dfew_cremad_rafdb} summarizes the results of the proposed approach on FER compared to different baselines. Face-LLaVA outperforms almost all the baselines on all the benchmarks under a zero-shot setting. Emotion-LLaMA \cite{cheng2024emotionllama} achieves superior weighted average recall (WAR) compared to our approach, but it is important to note that it is a multimodal model using audio as well as background context to make its predictions. As reported in the table, if we remove the background context, the performance of Emotion-LLaMA deteriorates significantly. Moreover, Face-LLaVA achieves competitive performance compared to the baselines in a finetuned setting as well despite lacking any background and audio context for DFEW \cite{jiang2020dfew} and Crema-D \cite{cremad_dataset} datasets. It is also important to note that, in contrast to other MLLM baselines \cite{cheng2024emotionllama, li2024facialfaba}, our model for finetuned setting is still trained to provide descriptions rather than outputting a single categorical label.

\noindent\textbf{AU Detection.} \cref{tab:res_disfa_bp4d} summarize the results for action unit detection on DISFA \cite{disfa_dataset} and BP4D \cite{ZHANG2014692_bp4d_dataset} datasets. Note that since BP4D and DISFA have a disjoint set of AU annotations, for training the model on a zero-shot setting, we augment \emph{FaceInstruct-1M} with Gemini-1.5 Flash \cite{geminiteam2024gemini15unlockingmultimodal} annotated AffectNet \cite{affectnet} images. Hence, AU information is injected in the zero-shot training by the AffectNet samples and the dataset other than the benchmark dataset (BP4D/DISFA) on which we are reporting the numbers. Face-LLaVA outperfroms all the zero-shot and fine-tuned baselines on average F1 scores on both datasets. On DISFA, we achieve a relative improvement of around 10\% on the previous finetuned SOTA. Moreover, we achieve around 7\% performance boost over Gemini-1.5 Flash \cite{geminiteam2024gemini15unlockingmultimodal}, further showing the effectiveness of our model and training set.

\noindent\textbf{Age Estimation, Attribute Detection, and Deepfake Detection.} \cref{tab:age_celeba_dfd} contains the results of the proposed approach on Morph II \cite{morph_ii_dataset}, UTKFace \cite{zhifei2017cvpr_utkface}, CelebA \cite{attr_liu_celeba} and FaceForensics++ \cite{roessler2019faceforensicspp} benchmarks. Similar to AU detection, Face-LLaVA outperforms all the zero-shot baselines and achieves competitive performance with finetuning for age estimation and facial attribute detection benchmarks. It is notable to achieve competitive results in comparison to regression-based baselines for age estimation while using an MLLM. For attribute detection, due to the lack of large-scale supervised datasets, we annotate FFHQ \cite{stylegan_ffhq} through Gemini-1.5 Flash without labels and use that to train Face-LLaVA for zero-shot performance comparison on CelebA \cite{attr_liu_celeba} dataset. For deepfake detection, we report all the results on low-quality videos of FaceForensics++ \cite{roessler2019faceforensicspp}. All the zero-shot MLLM baselines, including GPT4o-mini and Gemini-1.5 Flash, achieve close to or worse than the baseline random accuracy of 80\%. However, Face-LLaVA achieves significantly better accuracy in a zero-shot setting. The significant gap between the SOTA of deepfake detection \cite{f3_net} and Face-LLaVA in a fine-tuned setup can be attributed to the fact that the fine-tuned baselines use a higher number of frames per video while our model can only take eight frames per video. 

\begin{figure}
    \centering
    \includegraphics[width=\linewidth]{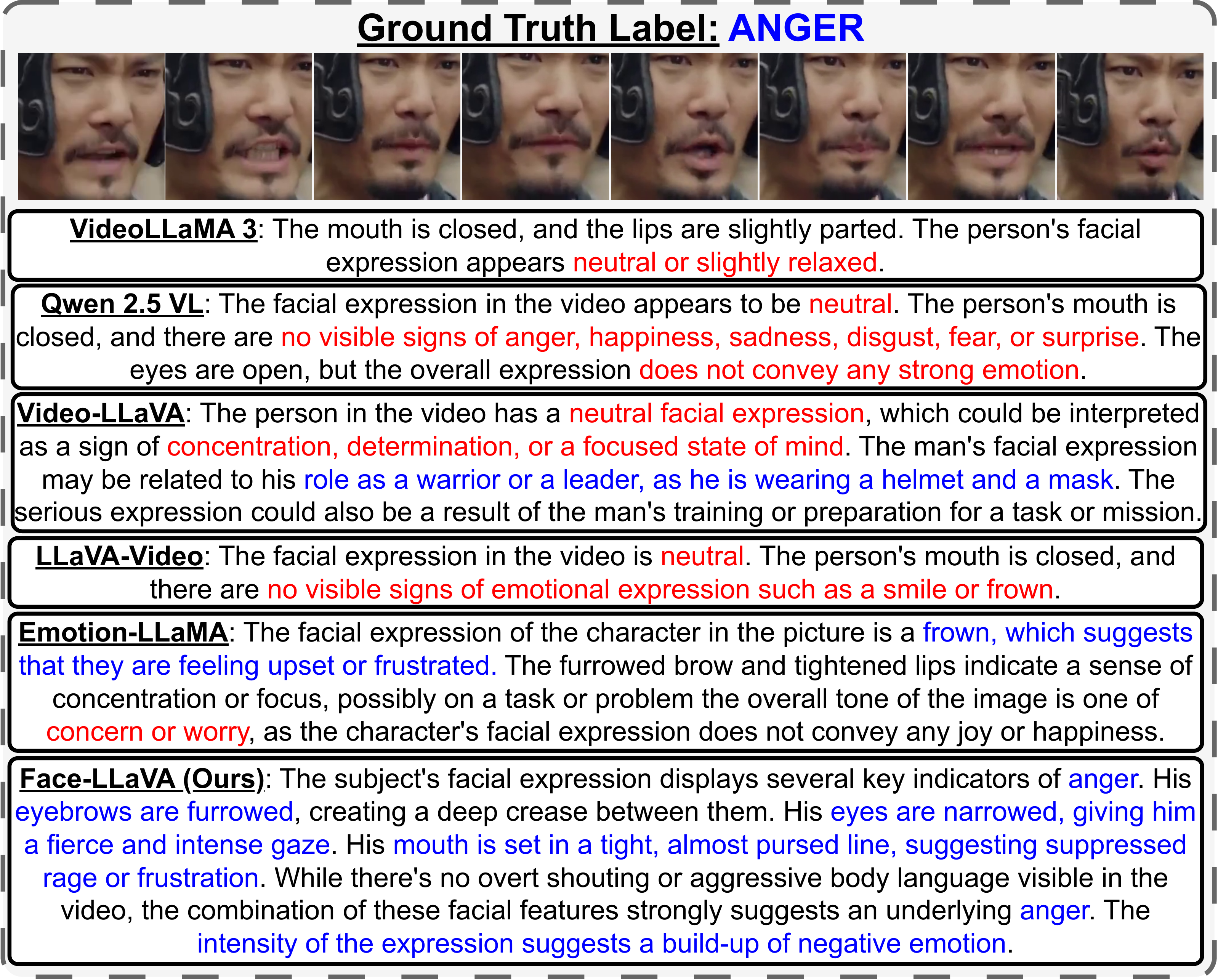}
    \vspace{-2em}
    \caption{Comparison of Face-LLaVA with other MLLM baselines on the task of facial expression recognition. \textcolor{red}{Red text} indicates misaligned text and \textcolor{blue}{blue text} indicates aligned/correct text according to the video and ground truth label.}
    \label{fig:qualitative_comparison}
    \vspace{-1.5em}
\end{figure}

\subsection{Evaluation of reasoning}

\label{subsec:reasoning_comparison_main}

As mentioned in \cref{subsec:dataset_evaluation}, we compare the reasoning capabilities of baselines with Face-LLaVA using GPT evaluation on the \emph{FaceInstruct-1M} \emph{test} set. Face-LLaVA achieves superior ratings compared to other MLLM baselines \cite{damonlpsg2025videollama3,lin-etal-2024-videollava,li2024llava_onevision,qwen25vl}. Across all tasks, Face-LLaVA's mean rating for completeness of reasoning is around 33\% higher than the best baseline, with a mean rating of 7.60/10. Moreover, Face-LLaVA responses achieve higher mean ratings than the baselines on response-video consistency and response-GT consistency as well highlighting excellent vision language alignment and high accuracy. Detailed summary of results can be found in \cref{tab:gpt4_evaluation} in \cref{sec:detailed_result_appendix}. \cref{fig:qualitative_comparison} presents a comparison of reasoning generated by different MLLMs for facial expression recognition in a video.

\subsection{Ablation Study}
\label{subsec:ablation_architecture}

We evaluate the contribution of different modules in an ablation study and report zero-shot model performance on DFEW \cite{jiang2020dfew} using MAFW \cite{liu_mafw_2022}, FERV39k \cite{liu_mafw_2022} and Crema-D \cite{cremad_dataset} for training. Our baseline model for this analysis is Video-LLaVA \cite{lin-etal-2024-videollava}, which only uses the visual tokens $h_v$ as input. 
We can clearly observe that using landmark tokens $h_l$ (or $h_l^{local}$/$h_l^{global}$) similar to EmoLA \cite{li2024facialfaba} as inputs in addition to the visual tokens $h_v$, improves baseline performance but not as much as incorporating the landmark tokens using cross-attention with the visual tokens $h_v$. Moreover, using FRGCA, which incorporates masked attention through region-patch proximity mask $m^{RPP}$, leads to a larger performance gain than a simple cross-attention. Finally, to compare using landmarks as opposed to face-parse heatmaps, we encode the face-parse heatmaps through the visual encoder $\mathbf{E_V}$ and replace the landmark tokens for cross-attention with visual tokens. Face parse maps show competitive performance compared to FRLP+FRGCA, at an extra compute and memory cost for the face-parse tokens (same size as visual tokens $h_v$).

\begin{table}[ht]
\centering
\resizebox{\linewidth}{!}{%
\begin{tabular}{l|ccc|cc}
\hline \hline
\rowcolor[HTML]{C0C0C0} 
\multicolumn{1}{c|}{\cellcolor[HTML]{C0C0C0}} & \cellcolor[HTML]{C0C0C0} & \cellcolor[HTML]{C0C0C0} & \cellcolor[HTML]{C0C0C0} & \multicolumn{2}{c}{\cellcolor[HTML]{C0C0C0}\textbf{DFEW \cite{jiang2020dfew}}} \\ \cline{5-6}
\rowcolor[HTML]{C0C0C0} 
\multicolumn{1}{c|}{\multirow{-2}{*}{\cellcolor[HTML]{C0C0C0}\textbf{Model}}} & \multirow{-2}{*}{\cellcolor[HTML]{C0C0C0}\textbf{\begin{tabular}[c]{@{}c@{}}Landmark\\ Projector\end{tabular}}} & \multirow{-2}{*}{\cellcolor[HTML]{C0C0C0}\textbf{\begin{tabular}[c]{@{}c@{}}Cross-\\ Attention\end{tabular}}} & \multirow{-2}{*}{\cellcolor[HTML]{C0C0C0}\textbf{\begin{tabular}[c]{@{}c@{}}Input\\ tokens\end{tabular}}} & \textbf{UAR} & \textbf{WAR} \\ \hline
Baseline & - & - & $h_v$ & 0.391 & 0.479  \\ \hline
 & only global & - & $h_v + h_l^{global}$ & 0.402 & 0.483 \\
 & only local & - & $h_v + h_l^{local}$ & 0.409 & 0.491 \\
 & FRLP & - & $h_v + h_l$ & 0.410 & 0.491 \\ \cline{2-6}
 & only global & simple & $h_v^{l_{global}}$ & 0.401 & 0.483 \\
 & only local & simple & $h_v^{l_{local}}$ & 0.409 & 0.494 \\
 & FRLP & simple & $h_v^l$ & 0.416 & 0.512 \\ \cline{2-6}
 & only local & FRGCA & $h_v^{l_{local}}$ & 0.412 & 0.511 \\
\multirow{-6}{*}{\begin{tabular}[c]{@{}l@{}}Baseline +\\ Landmarks\end{tabular}} & \textbf{FRLP} & \textbf{FRGCA} & $h_v^l$ & \textbf{0.424} & \textbf{0.520} \\ \hline
\begin{tabular}[c]{@{}l@{}}Baseline +\\ Face Parsing\end{tabular} & - & simple & $h_v^p$ & 0.413 & 0.498 \\ \hline \hline
\end{tabular}%
}
\vspace{-1em}
\caption{Ablations showing the effectiveness of the proposed FRLP and FRGCA modules compared to other design choices for zero-shot performance on DFEW \cite{jiang2020dfew}.}
\vspace{-1.8em}
\label{tab:archi_ablation}
\end{table}

\section{Ethics and Limitations}

\noindent\textbf{Ethics Statement.} As our approach involves face analysis, we acknowledge ethical concerns related to privacy, bias, and potential misuse. \emph{FaceInstruct-1M} is constructed from publicly available datasets with appropriate licensing, and we release only annotations and instructions, ensuring compliance with data protection regulations like GDPR. However, we recognize that facial recognition technologies may impact privacy and encourage responsible use. Since our dataset is derived from existing sources, it may inherit biases, and addressing these remains for future work. Our research is intended for ethical academic use, and we strongly discourage any application that infringes on individual rights or promotes discrimination.


\noindent\textbf{Limitations.} Face-LLaVA is limited to single-turn interactions and lacks advanced chain-of-thought reasoning. Moreover, we did not explore face identification or dense prediction tasks. Future work can address these limitations by integrating multi-turn dialogue capabilities and expanding to other facial tasks. 

\section{Conclusions}  
This work advances MLLMs for facial analysis across diverse tasks, including expression recognition, AU detection, attribute detection, age estimation, and deepfake detection. We introduce \emph{FaceInstruct-1M}, a large-scale dataset with over one million samples, automatically annotated using Gemini and GPT-4o, and propose Face-LLaVA, an MLLM leveraging novel FRLP and FRGCA modules to enrich visual representations for advanced face analysis. Extensive evaluations demonstrate its superiority over open-source MLLMs and competitive performance against task-specific methods, with GPT-4 confirming its advanced reasoning capabilities.

\section{Acknowledgements}
Research was sponsored by the Army Research Office and was accomplished under Cooperative Agreement Number W911NF-25-2-0040. The views and conclusions contained in this document are those of the authors and should not be interpreted as representing the official policies, either expressed or implied, of the Army Research Office or the U.S. Government. The U.S. Government is authorized to reproduce and distribute reprints for Government purposes notwithstanding any copyright notation herein.
{
    \small
    \bibliographystyle{ieeenat_fullname}
    \bibliography{main}
}
\clearpage
\setcounter{page}{1}
\onecolumn
\appendix

{
        \centering
        \Large
        \textbf{\thetitle}\\
        \vspace{0.5em}Supplementary Material \\
        \vspace{1.0em}
}


{\noindent \large \textbf{Table of Contents}
}

\begin{itemize}
    \item Ethics Statement \dotfill \cref{sec:ethics_appendix}
    \item Limitations and Future Work \dotfill \cref{sec:limitations_appendix}
    \item Detailed Results \dotfill \cref{sec:detailed_result_appendix}
    \item Task-specific Datasets Used \dotfill \cref{sec:traditional_data_desctiption}
    \item Additional Details about FaceInstruct-1M \dotfill \cref{sec:face_instruct_1m_additional_details}
    \item Evaluation \dotfill \cref{sec:supp_evaluation_protocol}
    \item Implementation Details \dotfill \cref{sec:supp_implementation_details}
    \item Reasoning Comparison with Baselines \dotfill \cref{subsec:reasoning_comparison_samples_supplementary}
    \item Failure Cases \dotfill \cref{sec:failure_cases}
\end{itemize}

\section{Ethics Statement}
\label{sec:ethics_appendix}
As the proposed approach involves face analysis, we acknowledge the ethical considerations associated with privacy, bias and potential misuse.

\noindent\textbf{Privacy and data protection.} \emph{FaceInstruct-1M} is constructed using existing task-specific datasets with appropriate licensing for research use. We will only release the annotations and instructions for \emph{FaceInstruct-1M} and not the video or images from the individual data sources, and the users of the dataset are recommended to obtain the videos from their original sources with appropriate consent. We do not collect or use private user data, ensuring compliance with data protection regulations such as GDPR. However, we recognize that facial recognition technologies may pose risks to individual privacy, and we encourage responsible use of our dataset and model.

\noindent \textbf{Bias and fairness.}
Facial analysis models often exhibit biases due to imbalanced training data, leading to disparities in performance across different demographic groups. Since our dataset is based on existing datasets for face analysis, it inherits some biases that are already present in those datasets. Evaluation of model bias is however left as a future work.

\noindent\textbf{Responsible use.}
Face analysis technologies have applications in various domains, including healthcare and accessibility. However, they also present risks if misused for mass surveillance, or profiling. Our research is intended for academic and ethical use, and we strongly discourage any application that infringes upon individuals' rights, promotes discrimination, or compromises security. 

We believe in transparency and open research. Upon acceptance, we will release our dataset and model to support further advancements in social AI development.

\section{Limitations and Future Work}
\label{sec:limitations_appendix}

While this work demonstrates pioneering efforts in using MLLMs for general face analysis, there are some limitations that can be addressed by future works. First, Face-LLaVA is trained on single-turn conversations and hence lacks advanced abilities such as conversationing and chain-of-thought reasoning. Such training will require augmenting \emph{FaceInstruct-1M} with conversation and reasoning data. Second, we only explored face-perception tasks and not face recognition or dense prediction tasks. While reasoning makes less sense in some of those tasks, there exists a potential to explore the performance of MLLMs on other facial tasks. Finally, since our dataset is automatically annotated using closed-source MLLMs, Gemini and GPT4, it contains some noise introduced by model hallucinations. 

\section{Detailed Results}
\label{sec:detailed_result_appendix}

This section contains detailed results from \cref{sec:experiments}. 

\noindent\textbf{Facial expression recognition.} \cref{tab:res_dfew_cremad_rafdb_long} is an expanded version of \cref{tab:res_dfew_cremad_rafdb} and contains the individual class recall values for the DFEW \cite{jiang2020dfew} dataset. We can observe that Face-LLaVA achieves better recall than the baselines on classes that are underrepresented in the data (disgust and fear). We noticed that some of the MLLM baselines are biased towards some expression categories, which is also evident by their recall values. For example, VideoLLaMA 3 \cite{damonlpsg2025videollama3} is biased towards the disgust and fear classes, while performing poorly on classes that other models are good at.

\noindent\textbf{Action unit detection.} \cref{tab:res_disfa,tab:res_bp4d} expand \cref{tab:res_disfa_bp4d} to show the F1 scores for individual action units. Notice that for the finetuning setting on DISFA \cite{disfa_dataset}, Face-LLaVA not only outperforms the baselines on average F1 score, but achieves the best F1 score on majority of the possible AUs in the dataset. A similar observation can be made for the analysis on the BP4D dataset \cite{ZHANG2014692_bp4d_dataset} in \cref{tab:res_bp4d}. 

\noindent\textbf{GPT Evaluation.} We report the mean GPT-4o-mini scores for all the tasks and scoring criterias in \cref{tab:gpt4_evaluation}. For all five tasks, the reasoning capabilities of Face-LLaVA generated results are rated higher than the baselines. Moreover, the high consistency of reasoning with ground truth suggests that Face-LLaVA provides a description or reason that aligns with the correct ground truth label. Notice that for AU detection, all the general MLLMs perform poorly compared to Face-LLaVA and manual verification of the results revealed that these models have less knowledge about the Facial Action Unit Coding System (FACS) and hence their outputs include hallucinations. To compare the quality of responses generated by our model in comparison to the baselines, please refer to \cref{subsec:reasoning_comparison_samples_supplementary}.

\begin{table*}[]
\centering
\resizebox{\linewidth}{!}{%
\begin{tabular}{lccccccccclcclc}
\hline \hline
\rowcolor[HTML]{C0C0C0} 
\multicolumn{10}{c|}{\cellcolor[HTML]{C0C0C0}\textbf{DFEW \cite{jiang2020dfew}}}  & \multicolumn{3}{c|}{\cellcolor[HTML]{C0C0C0}\textbf{Crema-D \cite{cremad_dataset}}} & \multicolumn{2}{c}{\cellcolor[HTML]{C0C0C0}\textbf{RAF-DB \cite{rafdb}}} \\ \hline
\rowcolor[HTML]{C0C0C0} 
\multicolumn{1}{c}{\cellcolor[HTML]{C0C0C0}\textbf{Method}} & \textbf{Hap $\uparrow$} & \textbf{Sad $\uparrow$} & \textbf{Neu $\uparrow$} & \textbf{Ang $\uparrow$} & \textbf{Sur $\uparrow$} & \textbf{Dis $\uparrow$} & \textbf{Fea $\uparrow$} & \textbf{UAR $\uparrow$} & \multicolumn{1}{c|}{\cellcolor[HTML]{C0C0C0}\textbf{WAR $\uparrow$}} & \multicolumn{1}{c}{\cellcolor[HTML]{C0C0C0}\textbf{Method}} & \textbf{UAR $\uparrow$} & \multicolumn{1}{c|}{\cellcolor[HTML]{C0C0C0}\textbf{WAR $\uparrow$}} & \multicolumn{1}{c}{\cellcolor[HTML]{C0C0C0}\textbf{Method}} & \textbf{Acc. $\uparrow$} \\ \hline
\multicolumn{15}{c}{\cellcolor[HTML]{FFFFFF}\textit{Closed - source models}} \\ \hline
GPT4o-mini \cite{gpt4omini} & 0.847 & 0.859 & 0.221 & 0.392 & 0.331 & 0.048 & 0.284 & 0.426 & \multicolumn{1}{c|}{0.518} & GPT4o-mini \cite{gpt4omini} & 0.410 & \multicolumn{1}{c|}{0.486} & GPT4o-mini \cite{gpt4omini} & 0.758 \\
Gemini-1.5F \cite{geminiteam2024gemini15unlockingmultimodal} & 0.646 & 0.762 & {0.357} & 0.287 & 0.320 & {0.227} & {0.435} & {0.433} & \multicolumn{1}{c|}{0.481} & Gemini-1.5F \cite{geminiteam2024gemini15unlockingmultimodal} & 0.465 & \multicolumn{1}{c|}{0.635} & Gemini-1.5F \cite{geminiteam2024gemini15unlockingmultimodal} & 0.685 \\ \hline
\rowcolor[HTML]{FFFFFF} 
\multicolumn{15}{c}{\cellcolor[HTML]{FFFFFF}\textit{Zero-shot}} \\ \hline
Vid.LLaMA 3 \cite{damonlpsg2025videollama3} & 0.619 & 0.131 & 0.502 & 0.031 & 0.000 & \textbf{0.364} & \textbf{0.357} & 0.286 & \multicolumn{1}{c|}{0.305} & Vid.LLaMA 3 \cite{damonlpsg2025videollama3} & 0.397 & \multicolumn{1}{c|}{0.546} & Vid.LLaMA 3 \cite{damonlpsg2025videollama3} & 0.671 \\
Qwen 2.5 \cite{qwen25vl} & 0.199 & 0.424 & \textbf{0.945} & 0.072 & 0.310 & 0.000 & 0.100 & 0.293 & \multicolumn{1}{c|}{0.399} & Qwen 2.5 \cite{qwen25vl} & 0.395 & \multicolumn{1}{c|}{0.566} & Qwen 2.5 \cite{qwen25vl} & 0.526 \\
Vid.-LLaVA \cite{lin-etal-2024-videollava} & 0.663 & 0.000 & 0.772 & 0.009 & 0.006 & 0.091 & 0.0 & 0.220 & \multicolumn{1}{c|}{0.326} & Vid.-LLaVA \cite{lin-etal-2024-videollava} & 0.367 & \multicolumn{1}{c|}{0.557} & Vid.-LLaVA \cite{lin-etal-2024-videollava} & 0.545 \\
LLaVA-Vid. \cite{zhang2024llava_video} & 0.639 & 0.671 & 0.807 & 0.097 & 0.012 & 0.000 & 0.286 & 0.375 & \multicolumn{1}{c|}{0.498} & LLaVA-Vid. \cite{zhang2024llava_video} & 0.478 & \multicolumn{1}{c|}{0.618} & LLaVA-OV \cite{li2024llava_onevision} & 0.700 \\
EmoLA\textsuperscript{*} \cite{li2024facialfaba} & 0.548 & 0.589 & 0.383 & 0.587 & 0.214 & 0.000 & 0.106 & 0.346 &  \multicolumn{1}{c|}{0.449} & EmoLA\textsuperscript{*} \cite{li2024facialfaba} & 0.431 & \multicolumn{1}{c|}{0.618} & EmoLA \cite{li2024facialfaba} & 0.741 \\
\begin{tabular}[c]{@{}l@{}}Emotion LLaMA\textsuperscript{\textdagger} \cite{cheng2024emotionllama}\end{tabular} & \textbf{0.720} & 0.763 & 0.620 & \textbf{0.719} & 0.337 & 0.000 & 0.033 & 0.456 & \multicolumn{1}{c|}{\textbf{0.594}} & \begin{tabular}[c]{@{}l@{}}Emotion LLaMA \cite{cheng2024emotionllama}\end{tabular} & 0.225 & \multicolumn{1}{c|}{0.308} & & \\
\begin{tabular}[c]{@{}l@{}}Emotion LLaMA\textsuperscript{\textdaggerdbl} \cite{cheng2024emotionllama}\end{tabular} & 0.412 & 0.516 & 0.289 & 0.444 & \textbf{0.351} & 0.000 & 0.100 & 0.302 & \multicolumn{1}{c|}{0.378} & & & \multicolumn{1}{c|}{} & & \\
\textbf{Face-LLaVA (Ours)} & 0.639 & \textbf{0.809} & 0.582 & 0.471 & 0.316 & 0.136 & 0.329 & \textbf{0.469} & \multicolumn{1}{c|}{\underline{0.564}} & \textbf{Face-LLaVA (Ours)} & \textbf{0.582} & \multicolumn{1}{c|}{\textbf{0.681}} & \textbf{Face-LLaVA (Ours)} & \textbf{0.780} \\ \hline
\multicolumn{15}{c}{\cellcolor[HTML]{FFFFFF}\textit{Fine-tuned}} \\ \hline
EC-STFL \cite{jiang2020dfew} & 0.792 & 0.491 & 0.579 & 0.610 & 0.461 & 0.028 & 0.215 & 0.454 & \multicolumn{1}{c|}{0.565} & Lei et al. \cite{cremad_lei} & 0.645 & \multicolumn{1}{c|}{0.648} & RUL \cite{rul_rafdb} & 0.890 \\
Former-DFER \cite{zhao2021former} & 0.841 & 0.626 & 0.675 & 0.700 & 0.564 & 0.035 & 0.318 & 0.537 & \multicolumn{1}{c|}{0.657} & PTH-Net \cite{cremad_pthnet} & 0.699 & \multicolumn{1}{c|}{0.700} & EAC \cite{eac_rafdb_emotion} & 0.909 \\
GCA+IAL \cite{Li2023-dfew_ial} & 0.880 & 0.672 & 0.701 & 0.761 & 0.622 & 0.000 & 0.264 & 0.557 & \multicolumn{1}{c|}{0.692} & MAE-DFER \cite{dfew_mae_dfer} & 0.773 & \multicolumn{1}{c|}{0.774} & TransFER \cite{transfer_rafdb} & 0.909 \\
M3DFEL  \cite{dfew_m3dfel} & 0.896 & 0.684 & 0.679 & 0.742 & 0.597 & 0.000 & 0.316 & 0.561 & \multicolumn{1}{c|}{0.693} & MTL-ER* \cite{mtler_cremad} & 0.745 & \multicolumn{1}{c|}{0.756} & Xue et al. \cite{rafdb_apvit} & 0.920 \\
MAE-DFER  \cite{dfew_mae_dfer} & 0.929 & 0.775 & 0.746 & 0.769 & 0.610 & \textbf{0.186} & 0.424 & 0.634 & \multicolumn{1}{c|}{0.744} & MT-Former* \cite{mtformer} & 0.793 & \multicolumn{1}{c|}{0.807} & POSTERv2 \cite{mao2023postersimplerstrongerfacial} & \textbf{0.922} \\
S2D \cite{dfew_s2d} & \textbf{0.936} & 0.803 & \textbf{0.771} & 0.811 & 0.645 & 0.014 & 0.347 & 0.618 & \multicolumn{1}{c|}{0.760} & MTCAE-DFER \cite{xiang2024mtcaedfer} & \textbf{0.847} & \multicolumn{1}{c|}{\textbf{0.850}} & EmoLA \cite{li2024facialfaba} & 0.921 \\
EMO-LLaMA \cite{xing2024emollama} & - & - & - & - & - & - & - & 0.602 & \multicolumn{1}{c|}{0.659} &  &  & \multicolumn{1}{c|}{} &  &  \\
Emotion-LLaMA  \cite{cheng2024emotionllama} & 0.931 & 0.794 & 0.725 & \textbf{0.841} & \textbf{0.728} & 0.035 & \textbf{0.442} & \textbf{0.642} & \multicolumn{1}{c|}{\textbf{0.771}} &  &  & \multicolumn{1}{c|}{} &  &  \\
\textbf{Face-LLaVA (Ours)} & 0.873 & \textbf{0.924} & 0.653 & 0.798 & 0.586 & 0.136 & 0.401 & \underline{0.625} &  \multicolumn{1}{c|}{\underline{0.745}} & \textbf{Face-LLaVA (Ours)} & \underline{0.798} & \multicolumn{1}{c|}{\underline{0.813}} & \textbf{Face-LLaVA (Ours)} & {\ul 0.921} \\ \hline \hline
\end{tabular}%

}
\caption{Comparison of the proposed approach with recent MLLMs and supervised techniques for emotion recognition on the DFEW \cite{jiang2020dfew}, Crema-D \cite{cremad_dataset} and RAF-DB \cite{rafdb} dataset. *:Only using middle video frame, \textdagger: Results taken from the paper \cite{cheng2024emotionllama}, \textdaggerdbl: Results computed after running inference code on face-cropped video.}
\label{tab:res_dfew_cremad_rafdb_long}
\end{table*}

\begin{table*}[]
\centering
\resizebox{0.85\textwidth}{!}{%
\begin{tabular}{lccccccccc}
\hline \hline
\rowcolor[HTML]{C0C0C0}
\multicolumn{1}{c}{\cellcolor[HTML]{C0C0C0}\textbf{Method}} & \textbf{AU1 $\uparrow$} & \textbf{AU2 $\uparrow$} & \textbf{AU4 $\uparrow$} & \textbf{AU6 $\uparrow$} & \textbf{AU9 $\uparrow$} & \textbf{AU12 $\uparrow$} & \textbf{AU25 $\uparrow$} & \textbf{AU26 $\uparrow$} & \textbf{Avg. F1 $\uparrow$} \\ \hline
\multicolumn{10}{c}{\emph{Closed-source models}} \\ \hline
GPT-4o-mini \cite{geminiteam2024gemini15unlockingmultimodal} & 0.292 & 0.302 & 0.565 & 0.416 & 0.493 & 0.244 & 0.582 & 0.536 & 0.429 \\
Gemini-1.5F \cite{geminiteam2024gemini15unlockingmultimodal} & 0.504 & 0.558 & 0.421 & 0.376 & 0.512 & 0.405 & 0.724 & 0.624 & 0.515 \\ \hline
\multicolumn{10}{c}{\emph{Zero-shot}} \\ \hline
VideoLLaMA 3 \cite{damonlpsg2025videollama3} & 0.101 & 0.253 & 0.394 & 0.369 & 0.450 & 0.395 & 0.493 & 0.534 & 0.374 \\
Qwen 2.5 VL \cite{qwen25vl} & 0.204 & 0.291 & \textbf{0.664} & 0.571 & 0.468 & 0.301 & 0.478 & 0.475 & 0.431 \\
Video-LLaVA \cite{lin-etal-2024-videollava} & 0.167 & 0.368 & 0.459 & 0.518 & 0.508 & \textbf{0.532} & 0.501 & 0.483 & 0.442 \\
LLaVA-OV \cite{li2024llava_onevision} & 0.044 & 0.143 & 0.201 & 0.146 & 0.469 & 0.182 & 0.520 & \textbf{0.535} & 0.280 \\
EmoLA \cite{li2024facialfaba} & 0.193 & 0.141 & 0.406 & 0.513 & 0.434 & 0.515 & \textbf{0.608} & \textbf{0.535} & 0.418 \\
\textbf{Face-LLaVA (Ours)} & \textbf{0.517} & \textbf{0.650} & 0.551 & \textbf{0.578} & \textbf{0.530} & 0.511 & 0.571 & 0.521 & \textbf{0.553} \\ \hline
\multicolumn{10}{c}{\emph{Fine-tuned}} \\ \hline
ATCM \cite{jacob2021facial} & 0.461 & 0.486 & 0.728 & 0.567 & 0.500 & 0.721 & 0.908 & 0.554 & 0.615 \\
ReCoT \cite{Li_2023_BMVC_recot} & 0.513 & 0.362 & 0.668 & 0.501 & 0.524 & 0.788 & \textbf{0.953} & \textbf{0.697} & 0.626 \\
KS \cite{knowledge_spreader_ks_au} & 0.538 & 0.599 & 0.692 & 0.542 & 0.508 & 0.758 & 0.922 & 0.468 & 0.628 \\
ME-GraphAU \cite{luo2022graphau} & 0.546 & 0.471 & 0.729 & 0.540 & 0.557 & 0.767 & 0.911 & 0.530 & 0.631 \\
$J\hat{A}A$-Net \cite{Shao2020JANetJF} & 0.624 & 0.607 & 0.671 & 0.411 & 0.451 & 0.735 & 0.909 & 0.674 & 0.635 \\
PIAP-DF  \cite{piap_df} & 0.502 & 0.518 & 0.719 & 0.506 & 0.545 & 0.797 & 0.941 & 0.572 & 0.638 \\
VL-FAU \cite{Ge2024-yp-vlfau} & 0.609 & 0.564 & 0.740 & 0.463 & 0.608 & 0.724 & 0.943 & 0.665 & 0.665 \\
AU-LLaVA \cite{hu2024unifiedfacialactionunit_aullava} & 0.520 & 0.592 & 0.444 & 0.308 & 0.223 & 0.661 & 0.908 & 0.546 & 0.525 \\
EmoLA \cite{li2024facialfaba} & 0.505 & 0.569 & \textbf{0.835} & 0.552 & 0.431 & 0.801 & 0.916 & 0.600 & 0.651 \\
\textbf{Face-LLaVA (Ours)} & \textbf{0.636} & \textbf{0.623} & 0.790 & \textbf{0.733} & \textbf{0.710} & \textbf{0.832} & 0.902 & 0.606 & \textbf{0.729} \\ \hline \hline
\end{tabular}%
}
\caption{Comparison of the proposed approach with recent MLLMs and supervised techniques on the 8 AUs of the DISFA \cite{disfa_dataset} dataset.}
\label{tab:res_disfa}
\end{table*}

\begin{table*}[]
\centering
\resizebox{\textwidth}{!}{%
\begin{tabular}{lccccccccccccc}
\hline \hline
\rowcolor[HTML]{C0C0C0} 
\multicolumn{1}{c}{\cellcolor[HTML]{C0C0C0}\textbf{Method}} & \textbf{AU1 $\uparrow$} & \textbf{AU2 $\uparrow$} & \textbf{AU4 $\uparrow$} & \textbf{AU6 $\uparrow$} & \textbf{AU7 $\uparrow$} & \textbf{AU10 $\uparrow$} & \textbf{AU12 $\uparrow$} & \textbf{AU14 $\uparrow$} & \textbf{AU15 $\uparrow$} & \textbf{AU17 $\uparrow$} & \textbf{AU23 $\uparrow$} & \textbf{AU24 $\uparrow$} & \textbf{Avg. F1 $\uparrow$} \\ \hline
\multicolumn{14}{c}{\emph{Closed-source models}} \\ \hline
GPT4o-mini \cite{gpt4omini} & {0.458} & {0.449} & 0.630 & {0.618} & {0.331} & 0.393 & {0.660} & 0.525 & {0.516} & 0.440 & 0.465 & 0.462 & {0.496} \\ 
Gemini-1.5F \cite{geminiteam2024gemini15unlockingmultimodal} & {0.533} & {0.556} & 0.606 & {0.687} & {0.487} & 0.432 & {0.756} & 0.395 & {0.543} & 0.427 & 0.466 & 0.493 & {0.532} \\ \hline
\multicolumn{14}{c}{\emph{Zero-shot}} \\ \hline
VideoLLaMA 3 \cite{damonlpsg2025videollama3} & 0.426 & 0.450 & 0.444 & 0.488 & \textbf{0.448} & 0.362 & 0.508 & 0.417 & 0.502 & 0.454 & 0.499 & 0.498 & 0.458 \\
Qwen 2.5 VL \cite{qwen25vl} & 0.260 & 0.422 & 0.549 & 0.702 & 0.328 & \textbf{0.503} & 0.579 & 0.441 & 0.455 & 0.446 & 0.459 & 0.461 & 0.467 \\
VideoLLaVA \cite{lin-etal-2024-videollava} & 0.477 & 0.495 & 0.461 & 0.396 & 0.296 & 0.470 & 0.357 & \textbf{0.491} & 0.489 & 0.459 & 0.470 & 0.483 & 0.445 \\
LLaVA-OV \cite{li2024llava_onevision} & 0.342 & 0.427 & 0.373 & 0.473 & 0.392 & 0.335 & 0.454 & 0.482 & 0.462 & \textbf{0.505} & \textbf{0.516} & \textbf{0.507} & 0.439 \\
EmoLA \cite{li2024facialfaba} & 0.185 & 0.143 & 0.584 & 0.541 & 0.304 & 0.266 & 0.647 & 0.375 & 0.419 & 0.500 & 0.450 & 0.468 & 0.407 \\
\textbf{Face-LLaVA (Ours)} & \textbf{0.494} & \textbf{0.498} & \textbf{0.648} & \textbf{0.687} & 0.312 & 0.295 & \textbf{0.746} & 0.375 & \textbf{0.510} & 0.445 & 0.462 & 0.470 & \textbf{0.495} \\ \hline
\multicolumn{14}{c}{\emph{Fine-tuned}} \\ \hline
$J\hat{A}A$-Net \cite{Shao2020JANetJF} & 0.538 & 0.478 & 0.582 & 0.785 & 0.758 & 0.827 & 0.882 & 0.637 & 0.433 & 0.618 & 0.456 & 0.499 & 0.624 \\
PIAP-DF  \cite{piap_df} & 0.542 & 0.471 & 0.540 & 0.790 & 0.782 & 0.863 & 0.895 & 0.661 & 0.497 & 0.632 & 0.499 & 0.520 & 0.641 \\
ATCM \cite{jacob2021facial} & 0.517 & 0.493 & 0.610 & 0.778 & 0.795 & 0.829 & 0.863 & \textbf{0.676} & 0.519 & 0.630 & 0.437 & 0.563 & 0.642 \\
ReCoT \cite{Li_2023_BMVC_recot} & 0.515 & 0.478 & 0.589 & 0.792 & 0.802 & 0.849 & 0.884 & 0.616 & 0.533 & 0.646 & 0.518 & 0.554 & 0.648 \\
\begin{tabular}[c]{@{}l@{}}ME-GraphAU \\ (SWIN) \cite{luo2022graphau}\end{tabular} & 0.527 & 0.443 & 0.609 & 0.799 & 0.801 & 0.853 & 0.892 & 0.694 & 0.554 & 0.644 & 0.498 & 0.551 & 0.655 \\
AU-LLaVA \cite{hu2024unifiedfacialactionunit_aullava} & 0.582 & 0.459 & 0.619 & 0.786 & 0.756 & \textbf{0.878} & \textbf{0.905} & 0.590 & 0.324 & 0.625 & 0.305 & 0.403 & 0.603 \\
EmoLA \cite{li2024facialfaba} & \textbf{0.574} & 0.524 & 0.610 & 0.781 & 0.778 & 0.819 & 0.895 & 0.605 & 0.493 & \textbf{0.649} & 0.460 & 0.524 & 0.642 \\
VL-FAU \cite{Ge2024-yp-vlfau} & 0.563 & 0.499 & 0.626 & 0.795 & \textbf{0.801} & 0.826 & 0.886 & 0.668 & 0.513 & 0.635 & 0.513 & 0.571 & \textbf{0.658} \\
\textbf{Face-LLaVA (Ours)} & 0.541 & \textbf{0.610} & \textbf{0.642} & \textbf{0.801} & 0.663 & 0.718 & 0.861 & 0.644 & \textbf{0.566} & 0.627 & \textbf{0.579} & \textbf{0.649} & \textbf{0.658} \\ \hline \hline
\end{tabular}%
}
\caption{Comparison of the proposed approach with recent MLLMs and supervised techniques on the 12 AUs of the BP4D \cite{ZHANG2014692_bp4d_dataset} dataset.}
\label{tab:res_bp4d}
\end{table*}

\begin{table*}[]
\centering
\resizebox{1\textwidth}{!}{%
\begin{tabular}{l|cccccc|cccccc|cccccc}
\hline \hline
\rowcolor[HTML]{C0C0C0} 
\multicolumn{1}{c|}{\cellcolor[HTML]{C0C0C0}} & \multicolumn{6}{c|}{\cellcolor[HTML]{C0C0C0}\textbf{Reason-Video Consistency}} & \multicolumn{6}{c|}{\cellcolor[HTML]{C0C0C0}\textbf{Reason-GT Consistency}} & \multicolumn{6}{c}{\cellcolor[HTML]{C0C0C0}\textbf{Reasoning Completeness}} \\ \cline{2-16} 
\rowcolor[HTML]{C0C0C0} 
\multicolumn{1}{c|}{\multirow{-2}{*}{\cellcolor[HTML]{C0C0C0}\textbf{Method}}} & \textbf{Emo.} & \textbf{AU} & \textbf{Attr.} & \textbf{Age} & \textbf{DF.} & \textbf{All} & \textbf{Emo.} & \textbf{AU} & \textbf{Attr.} & \textbf{Age} & \textbf{DF.} & \textbf{All} & \textbf{Emo.} & \textbf{AU} & \textbf{Attr.} & \textbf{Age} & \textbf{DF.} & \textbf{All} \\ \hline
GT from \emph{FaceInstruct-1M} & 9.47 & 8.52 & 9.80 & 9.27 & 8.85 & 9.18 & 9.70 & 8.84 & 9.88 & 9.55 & 9.56 & 9.51 & 9.21 & 8.26 & 9.75 & 9.02 & 8.41 & 8.93 \\ \hline
VideoLLaMA 3 \cite{damonlpsg2025videollama3} & 5.14 & 2.58 & 6.90 & 5.82 & 7.02 & 5.49 & 5.27 & 2.06 & 6.27 & 5.13 & 7.64 & 5.27 & 4.90 & 2.73 & 6.50 & 5.37 & 6.51 & 5.20  \\
Qwen 2.5 VL \cite{qwen25vl} & 5.82 & 3.02 & 5.48 & 7.36 & 5.48  & 5.43 & 5.96 & 2.54 & 4.86 & 7.02 & 5.76  & 5.23 & 5.57 & 3.34 & 5.21 & 6.89 & 5.30 & 5.26  \\
Video LLaVA \cite{lin-etal-2024-videollava} & 4.31 & 2.58 & 5.82 & 7.79 & 6.19 & 5.34 & 4.47 & 2.06 & 5.28 & 7.10 & 6.58 & 5.10 & 4.20 & 2.73 & 5.30 & 7.00 & 5.84 & 5.01  \\
LLaVA-OV \cite{li2024llava_onevision} & 7.11 & 2.18 & 6.08 & \textbf{7.97} & 6.69 & 6.01 & 7.30 & 1.95 & 5.48 & 7.44 & 7.33 & 5.9 & 6.67 & 2.34 & 5.72 & 7.52 & 6.19 & 5.69 \\
EmoLA \cite{li2024facialfaba} & 7.33 & 5.17 & - & - & - & - & 7.58 & 5.04 & - & - & - & - & 6.81 & 5.32 & - & - & - & - \\
Emotion-LLaMA \cite{cheng2024emotionllama} & 6.77 & - & - & - & - & - & 6.90 & - & - & - & - & - & 6.50 & - & - & - & - & - \\
\textbf{Face-LLaVA (Ours)} & \textbf{7.95} & \textbf{6.90} & \textbf{8.34} & {\ul 7.68} & \textbf{8.56} & \textbf{7.89} & \textbf{8.14} & \textbf{6.68} & \textbf{8.13} & \textbf{7.53} & \textbf{9.20} & \textbf{7.94} & \textbf{7.79} & \textbf{6.62} & \textbf{7.89} & \textbf{7.59} & \textbf{8.11} & \textbf{7.60} \\ \hline \hline
\end{tabular}%
}
\caption{Mean GPT4o ratings (on a scale of 1-10) for different methods in a zero-shot setting on the \emph{FaceInstruct-1M} Test Set.}
\label{tab:gpt4_evaluation}
\end{table*}



\section{Task-specific datasets used}
\label{sec:traditional_data_desctiption}

As mentioned in \cref{sec:dataset_faceinstruct_1m}, \emph{FaceInstruct-1M} is constructed using task-specific face analysis datasets for facial expression recognition, facial action unit detection, age estimation, facial attributes detection and deepfake detection. We present a summary of these datasets in this section. We refer the readers to \cref{sec:supp_evaluation_protocol} for a detailed description of the evaluation protocol on these datasets.

\subsection{Facial Expression Recognition}

\noindent \textbf{Dynamic Facial Expression in-the-Wild (DFEW) \cite{jiang2020dfew}} is a large-scale facial expression dataset comprising 16,372 video clips sourced from movies. Each clip is manually annotated by 12 expert annotators, with 10 independent labels per clip. The dataset includes seven basic emotion categories: happiness, sadness, neutral, anger, surprise, fear, and disgust. Although some clips have multiple emotion labels, we observed that the perceived emotion is often ambiguous in these cases. Therefore, we conduct all experiments on the single-labeled subset of 11.7k clips. As a multimodal dataset, DFEW contains audio and background context (e.g., multiple actors, body gestures), meaning the emotion labels may not be solely based on facial expressions.

\noindent \textbf{MAFW \cite{liu_mafw_2022}} is another large-scale dynamic facial expression dataset containing approximately 10k movie clips. Each clip is annotated by 11 professional annotators for 11 emotion categories, including the seven basic emotions plus contempt, anxiety, helplessness, and disappointment. However, since these additional emotion categories have relatively few samples, we exclude them from our experiments. Like DFEW, MAFW is a multimodal dataset containing audio.

\noindent \textbf{FERV39k \cite{wang2022ferv39k}} is a large-scale multi-scene dataset featuring 39k video samples, each categorized into one of seven basic emotions across 22 different scene types. The dataset is annotated through crowd-sourcing and professional annotators, with 30 independent annotations per clip. Unlike DFEW and MAFW, FERV39k does not contain audio, but it still includes background information such as multiple actors, body movements, and hand gestures, which can influence perceived emotions.

\noindent \textbf{Crema-D \cite{cremad_dataset}} is a emotional multimodal dataset, containing 7,442 clips from 91 actors, including 48 male and 43 females. They are between 20 and 74, from a variety of races and ethnicities. The actors spoke 12 sentences, which were presented using one of 6 emotion categories, and four different emotion levels. The dataset is annotated by 2,443 participants, promising 95\% of the clips have more than 7 ratings.

\noindent \textbf{AffectNet \cite{affectnet}} is a large-scale facial expression dataset containing more than 1M facial images. The dataset is collected by querying 1250 emotion related keywords in 6 different languages on the Internet with three major search engines. About half of the retrieved images were manually annotated with seven facial expression categories. 

\noindent \textbf{RAF-DB \cite{rafdb}} is a large-scale facial expression dataset, containing 29,672 facial images with a variety of age, gender and ethnicity. It is annotated by 40 independent annotators. EM algorithm was applied to filter out unreliable labels.
\subsection{Action Unit Detection}

\noindent \textbf{DISFA \cite{disfa_dataset}} is a non-posed facial expression dataset containing videos of 27 adults with different ethnicities, with high resolution. All video frames are annotated by two human FACS experts for the intensity of AUs (0-5 scale). 

\noindent \textbf{BP4D \cite{ZHANG2014692_bp4d_dataset}} is a 3D video database of facial expressions of 41 young adults between ages 18 to 29, The dataset is manually annotated with 12 action units, and contains automatically tracked head pose and 2D/3D facial landmarks.

\subsection{Facial Attribute Detection}

\noindent\textbf{CelebA \cite{attr_liu_celeba}} is a large-scale face attributes dataset, containing more than 200K celebrity images, each with 40 attribute annotations. Celeba includes images with large diversity in pose and background. 

\subsection{Age Estimation}

\noindent\textbf{MORPH II \cite{morph_ii_dataset}} is a dataset containing 55134 mugshots, annotated with age estimate, gender, and race classification.

\noindent\textbf{UTKFace \cite{zhifei2017cvpr_utkface}} is a large-scale face dataset with people aged between 0 to 116 years old, containing over 20K images annotated with age, gender, and ethnicity.

\subsection{Deepfake Detection}

\noindent\textbf{FaceForensics++ \cite{roessler2019faceforensicspp}} is a dataset consisting of 1000 original video sequence, manipulated with 4 face manipulation methods: Deepfakes, Face2Face, FaceSwap, and NeuralTextures. The videos were generated from 977 YouTube source videos.

\noindent\textbf{Fake AV-Celeb \cite{khalid2021fakeavceleb}} is a dataset that includes about 20K manipulated videos generated using various deepfake synthesis methods. The base set consists of 500 real videos of celebrities from YouTube.

\begin{figure*}
    \centering
    \includegraphics[width=0.9\linewidth]{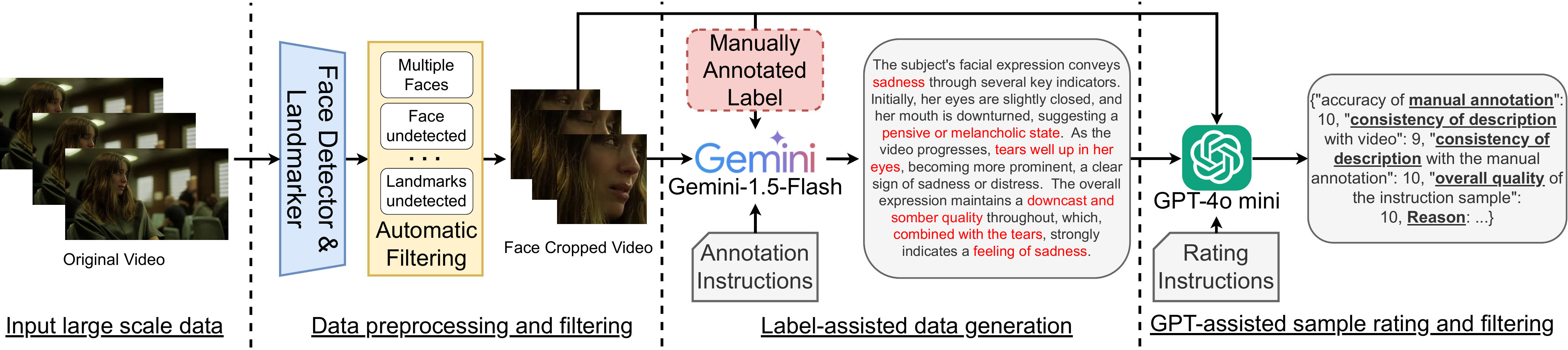} 
    \caption{Data annotation pipeline used for creating \emph{FaceInstruct-1M} dataset.}
    \label{fig:data_annotation_pipeline}
\end{figure*}

\section{Additional details about FaceInstruct-1M}
\label{sec:face_instruct_1m_additional_details}

\cref{fig:data_annotation_pipeline} illustrates the annotation pipeline used to create our dataset from a large-scale, manually annotated dataset. 

\begin{table}[]
\centering
\resizebox{0.5\linewidth}{!}{%
\begin{tabular}{l|c|ccc}
\hline \hline
\rowcolor[HTML]{C0C0C0} 
\multicolumn{1}{c|}{\cellcolor[HTML]{C0C0C0}} & \cellcolor[HTML]{C0C0C0} & \multicolumn{3}{c}{\cellcolor[HTML]{C0C0C0}\textbf{Number of Samples}} \\ \cline{3-5} 
\rowcolor[HTML]{C0C0C0} 
\multicolumn{1}{c|}{\multirow{-2}{*}{\cellcolor[HTML]{C0C0C0}\textbf{Dataset}}} & \multirow{-2}{*}{\cellcolor[HTML]{C0C0C0}\textbf{Task}} & \textbf{Initial} & \textbf{\begin{tabular}[c]{@{}c@{}}After\\ Preproc.\end{tabular}} & \textbf{\begin{tabular}[c]{@{}c@{}}After\\ GF.\end{tabular}} \\ \hline
DFEW \cite{jiang2020dfew} & Expression & 11.7k & 6.7k & 6.2k \\
MAFW \cite{liu_mafw_2022} & Expression & 10k & 6.9k & 6.6k \\
FERV39k \cite{wang2022ferv39k} & Expression & 39k & 30.7k & 28.8k \\
Crema-D \cite{cremad_dataset} & Expression & 7.4k & 7.4k & 6.8k \\
AffectNet \cite{affectnet}& Expression & 287k & 280k & 260k \\
RAF-DB \cite{rafdb} & Expression & 15k & 15k & 14.8k \\
DISFA \cite{disfa_dataset} & AU & 131k & 130k & 123k \\
BP4D \cite{ZHANG2014692_bp4d_dataset}& AU & 150k & 146k & 128k \\
CelebA \cite{attr_liu_celeba} & Attributes & 203k & 201k & 196k \\
UTK Face \cite{zhifei2017cvpr_utkface} & Age & 24.1k & 23.5k & 22.8k \\
MORPH II \cite{morph_ii_dataset} & Age & 50k & 49.9k & 49k \\
FaceForensics++ \cite{roessler2019faceforensicspp} & Deepfake & 30k & 25.9k & 24.7k \\
Fake AV-Celeb \cite{khalid2021fakeavceleb} & Deepfake & 20k & 19.5k & 19.3k \\ 
Real Faces* \cite{jiang2020dfew, liu_mafw_2022, wang2022ferv39k} & Deepfake & 60.7k & 44.3k & 43.8k \\ \hline
\textbf{Total} & - & \textbf{1.04M} & \textbf{987k} & \textbf{930k} \\
\hline 
\hline
\end{tabular}%
}
\caption{Statistics about data preprocessing and GPT-filtering (GF.) of for each of the constituent datasets of \emph{FaceInstruct-1M}. *: Real faces augmentation for deepfake detection task is created from DFEW, MAFW and FERV39k.}
\label{tab:dataset_filtering_preprocessing}
\end{table}
 
\subsection{Data preprocessing}
\label{subsec:data_preprocessing}
As mentioned in \cref{subsec:annotation_pipeline} and shown in \cref{fig:data_annotation_pipeline}, we pre-process the visual inputs, i.e. images and videos in each of the constituent datasets before getting the annotations from Gemini. We use Mediapipe \cite{BlazeFace} with default parameters to detect face bounding boxes within the video or image. For video datasets, we filter away all the videos in which the face of the subject goes out of the frame and cases where multiple faces are detected. We do not apply any smoothing to the detection bounding boxes for videos as we found that such smoothing fails in cases where the subject's face has a sudden movement and hence, in such cases, the face moves out of the smoothed bounding box. We do not align the faces after cropping because for some tasks such as facial expression recognition, certain head movements such as nodding and head shakes might contain signals to predict and reason the correct output. After cropping the face bounding box from a video or an image, we resize the image into 256x256 for all downstream processing and training.

After face cropping, we used FAN \cite{bulat2017far} for detecting 68 2-D facial landmarks for all the samples in constituent datasets. We filter away all the samples in which landmark is not detected even in one of the frames of the video. Such, strict filtering ensures high data quality. 

We report the statistics of our data after preprocessing in \cref{tab:dataset_filtering_preprocessing}. Note that for FaceForensics++ \cite{roessler2019faceforensicspp} dataset, since the number of samples in the training set are quite small (only 4k), we split the dataset into chunks of 3 seconds and use those for all the processing. 

\subsection{Getting descriptions/reasoning from Gemini}
\begin{figure*}[!b]
    \centering
    \includegraphics[width=0.8\linewidth]{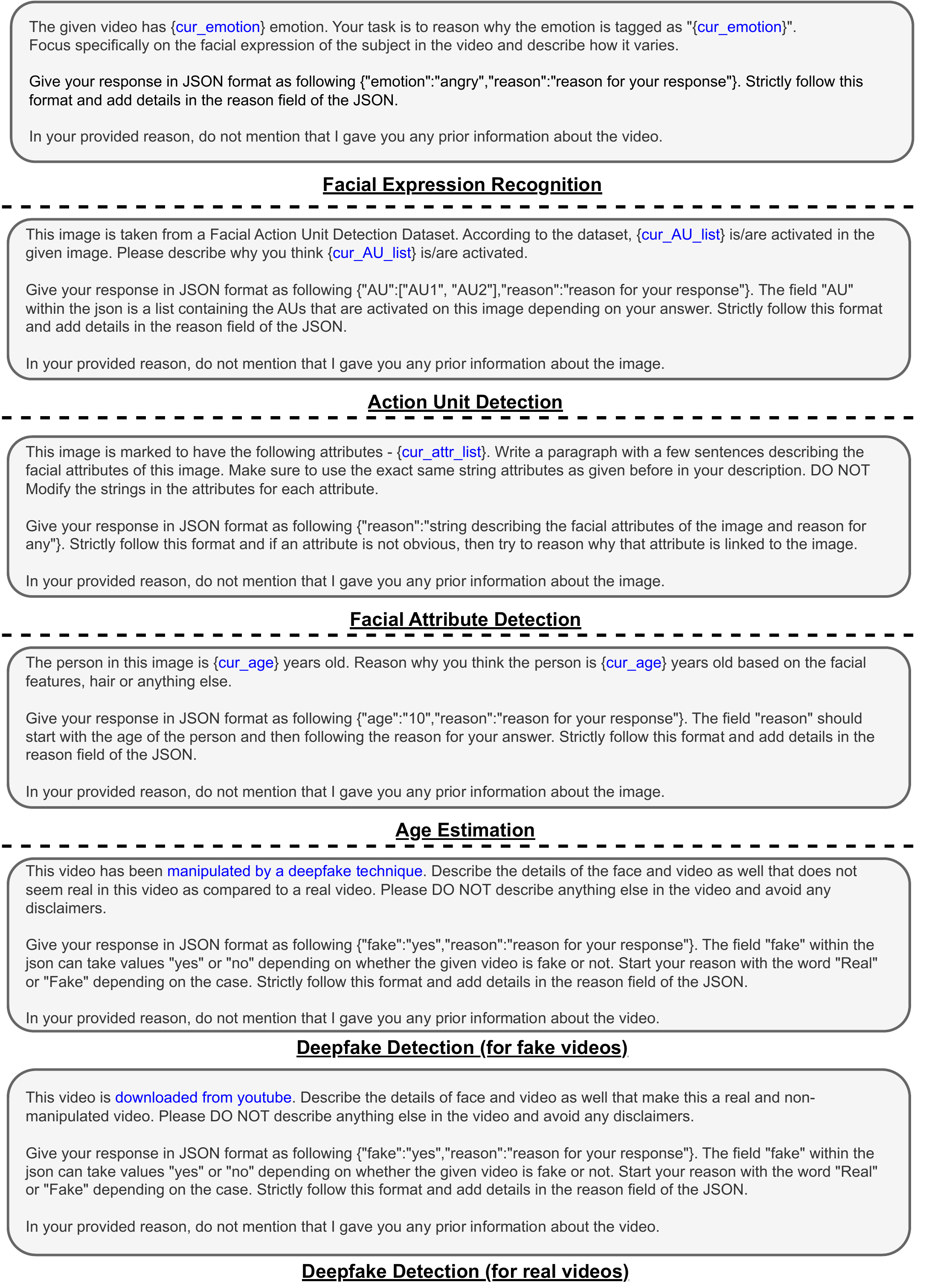}
    \caption{Prompts used for generating descriptions or reasoning from Gemini-1.5 Flash. Notice that we pass the label information about the data through \textcolor{blue}{blue text}.}
    \label{fig:gemini_annotation_prompts}
\end{figure*}

We use Gemini-1.5 Flash \cite{geminiteam2024gemini15unlockingmultimodal} for getting all the data annotations for constituent datasets. Note that Gemini-2.0 was still under an experimental phase during the creation of \emph{FaceInstruct-1M} so we used Gemini-1.5 Flash. \cref{fig:gemini_annotation_prompts} illustrates the prompts that we have used to get annotations from Gemini. Note that we do some prompt tuning for each task and provide negative prompts to control the output format, and to restrict the model from generating disclaimers. Moreover, we explicitly prompt the model to not mention that it has been provided with the ground truth information so that the generated descriptions can directly be used for prompt tuning.

It is also important to note that for the deepfake detection and age estimation tasks, we explicitly ask the model to start its response with the ground truth label so that it is easier to parse the responses, when computing traditional metrics for these datasets.

\subsection{Task-specific instructions}
\begin{figure*}
    \centering
    \includegraphics[width=0.8\linewidth]{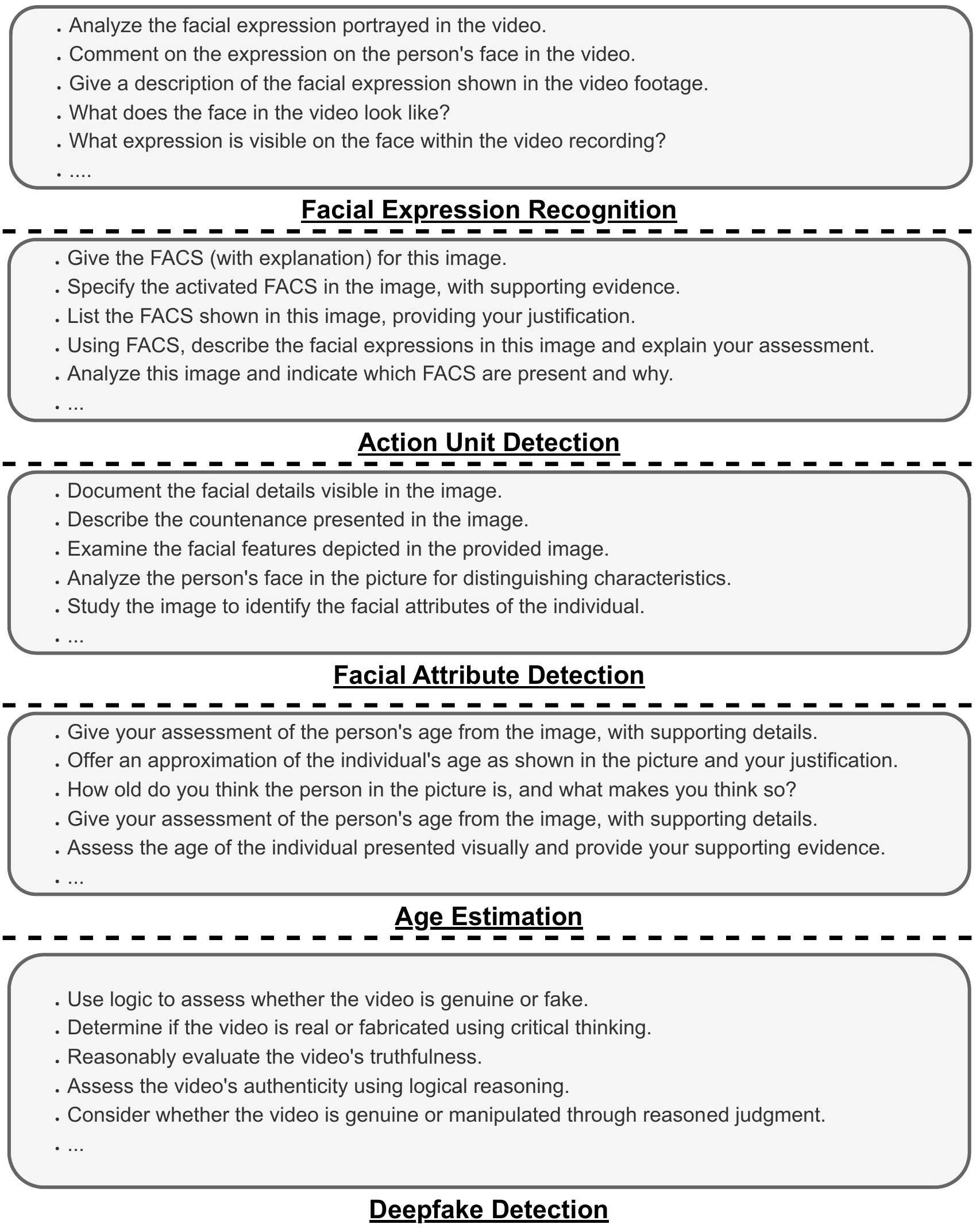}
    \caption{Samples of various instructions used for different face analysis tasks. Note that for each task, the instructions are analogous which allows us to use them randomly with any sample belonging to that task during training, thereby augmenting the data size.}
    \label{fig:task_specific_instructions}
\end{figure*}

To train our model on \emph{FaceInstruct-1M} dataset, we need task-specific instructions for different tasks. To that extent, we carefully collected 100 handcrafted instructions for each of the five tasks. Some example instructions for different tasks are illustrated in \cref{fig:task_specific_instructions}. Notice that since the instructions for a particular task are semantically similar, we can use the instructions randomly during training similar to \cite{cheng2024emotionllama} during training. Note that we replace the \emph{`video'}/\emph{`image'} string in the instruction with appropriate type of data depending on the current sample.

\subsection{GPT-4 Filtering}
\begin{figure*}
    \centering
    \includegraphics[width=0.92\linewidth]{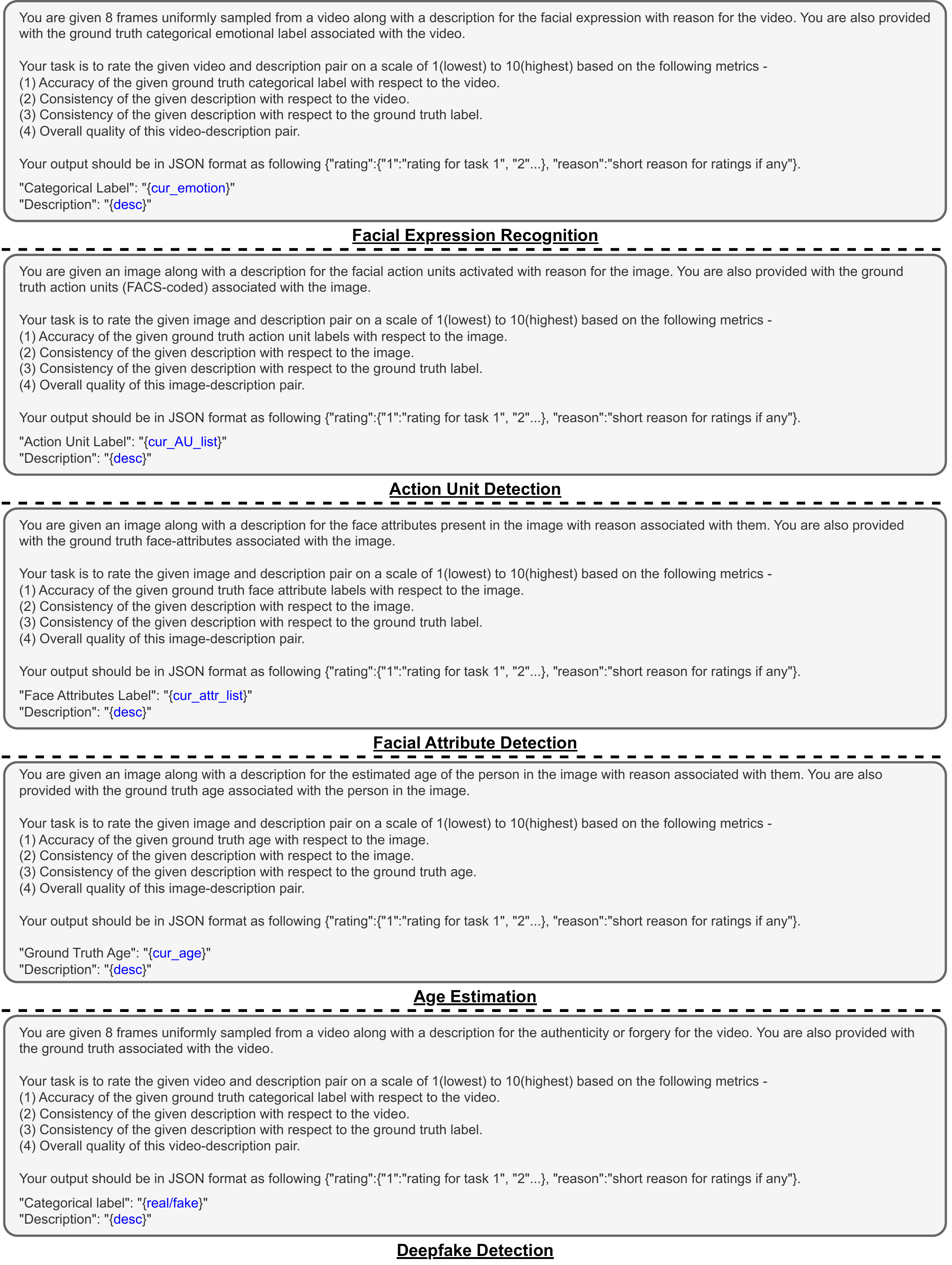}
    \caption{Rating instructions (Prompts) used for data filtering using GPT4o-mini. Notice that we pass the label information about the data and the description generated from Gemini through \textcolor{blue}{blue text}.}
    \label{fig:gpt_filtering_prompts}
\end{figure*}

As mentioned in \cref{subsec:dataset_filtering}, we employ GPT4o-mini to rate the annotations obtained from Gemini to perform additional filtering. Moreover, this automated rating works as a sanity check to understand the quality of annotations generated from Gemini. \cref{fig:gpt_filtering_prompts} shows the rating instructions or prompts that we use for GPT4 assisted rating. We rate the annotations for all the prompts on the same four criteria - (i) accuracy of the manually annotated label w.r.t. the face-cropped video, (ii) consistency of the generated description with the face-cropped video, (iii) consistency of the generated description with the manually annotated label, and (iv) overall quality of the sample based on resolution, visibility of the face, etc. Moreover, we also ask the model to provide a short reason for its ratings to assist us in determining the reason for an unexpectedly high or low rating. \cref{tab:dataset_filtering_preprocessing} summarizes the number of videos that are filtered after GPT filtering. As mentioned in \cref{subsec:dataset_filtering}, we filter away all the samples with an overall rating less than or equal to 6, thereby resulting in about 7\% of the initial dataset getting filtered away.

\clearpage

\begin{table}[]
\centering
\resizebox{0.5\linewidth}{!}{%
\begin{tabular}{l|cc|c|cc}
\hline \hline
\rowcolor[HTML]{C0C0C0} 
\multicolumn{1}{c|}{\cellcolor[HTML]{C0C0C0}} & \cellcolor[HTML]{C0C0C0} & \cellcolor[HTML]{C0C0C0} & \cellcolor[HTML]{C0C0C0} & \multicolumn{2}{c}{\cellcolor[HTML]{C0C0C0}\textbf{DFEW \cite{jiang2020dfew}}} \\ \cline{5-6}
\rowcolor[HTML]{C0C0C0} 
\multicolumn{1}{c|}{\multirow{-2}{*}{\cellcolor[HTML]{C0C0C0}\textbf{Constituents}}} & \multirow{-2}{*}{\cellcolor[HTML]{C0C0C0}\textbf{\begin{tabular}[c]{@{}c@{}}Labels \\used\end{tabular}}} & \multirow{-2}{*}{\cellcolor[HTML]{C0C0C0}\textbf{\begin{tabular}[c]{@{}c@{}}GPT- \\filtering\end{tabular}}} & \multirow{-2}{*}{\cellcolor[HTML]{C0C0C0}\textbf{\begin{tabular}[c]{@{}c@{}}No. of\\ samples\end{tabular}}} & \textbf{UAR} & \textbf{WAR} \\ \hline
\begin{tabular}[c]{@{}l@{}}M, F, C\end{tabular} & \xmark & \xmark & 44.7k & 0.298 & 0.350 \\
\begin{tabular}[c]{@{}l@{}}M, F, C, MER\end{tabular} & \xmark & \xmark & 89.7k & 0.293 & 0.355 \\
\begin{tabular}[c]{@{}l@{}}M, F, C\end{tabular} & \xmark & \cmark & 15.4k & 0.318 & 0.375 \\ 
\begin{tabular}[c]{@{}l@{}}M, F, C, MER\end{tabular} & \xmark & \cmark & 25.9k & 0.327 & 0.392 \\ \hline
\begin{tabular}[c]{@{}l@{}}M, F, C\end{tabular} & \cmark & \xmark & 44.7k & 0.415 & 0.501 \\
\begin{tabular}[c]{@{}l@{}}M, F, C\end{tabular} & \cmark & \cmark & 40.2k & 0.424 & 0.520 \\ \hline \hline
\end{tabular}%
}
\caption{Ablations showing the effectiveness of using annotation labels for data generation and GPT filtering on zero-shot model performance on DFEW \cite{jiang2020dfew} dataset. M:MAFW \cite{liu_mafw_2022}, D: DFEW \cite{jiang2020dfew}, F:FERV39k \cite{wang2022ferv39k}, C: Crema-D \cite{cremad_dataset}, MER: MER2023 \cite{mer2023}.}
\label{tab:data_ablation}
\end{table}

\subsection{Ablation for data annotation pipeline}
\label{subsubsec:data_ablation}

To demonstrate the effectiveness of incorporating labels as additional signals and applying GPT-4o-mini filtering during dataset construction, we conduct ablation experiments on a subset of \emph{FaceInstruct-1M} for zero-shot expression recognition on DFEW \cite{jiang2020dfew}. As baseline datasets for this study, we use MAFW \cite{liu_mafw_2022}, FERV39k \cite{wang2022ferv39k}, and Crema-D \cite{cremad_dataset}. Additionally, to assess whether increasing the number of unlabeled samples improves performance, we include an unlabeled dataset from MER2023 \cite{mer2023}, containing approximately 45k samples.

\cref{tab:data_ablation} summarizes our findings. The results indicate a significant performance improvement when leveraging ground truth labels from constituent datasets to generate annotations using Gemini. Unlike traditional self-supervised learning approaches, where increasing data volume typically enhances performance, we observe that adding more unlabeled data does not lead to better task performance -- an important consideration for face analysis. Finally, filtering the dataset with GPT-4o-mini results in additional performance gain, as the data is labeled by one expert model (Gemini) and subsequently rated and filtered by another (GPT), improving overall data quality.

\clearpage

\subsection{Data statistics}

\begin{figure*}[h!]
    \centering
    \begin{subfigure}[t]{0.5\textwidth}
        \centering
        \includegraphics[trim={0 0 0 5mm},clip,width=\textwidth]{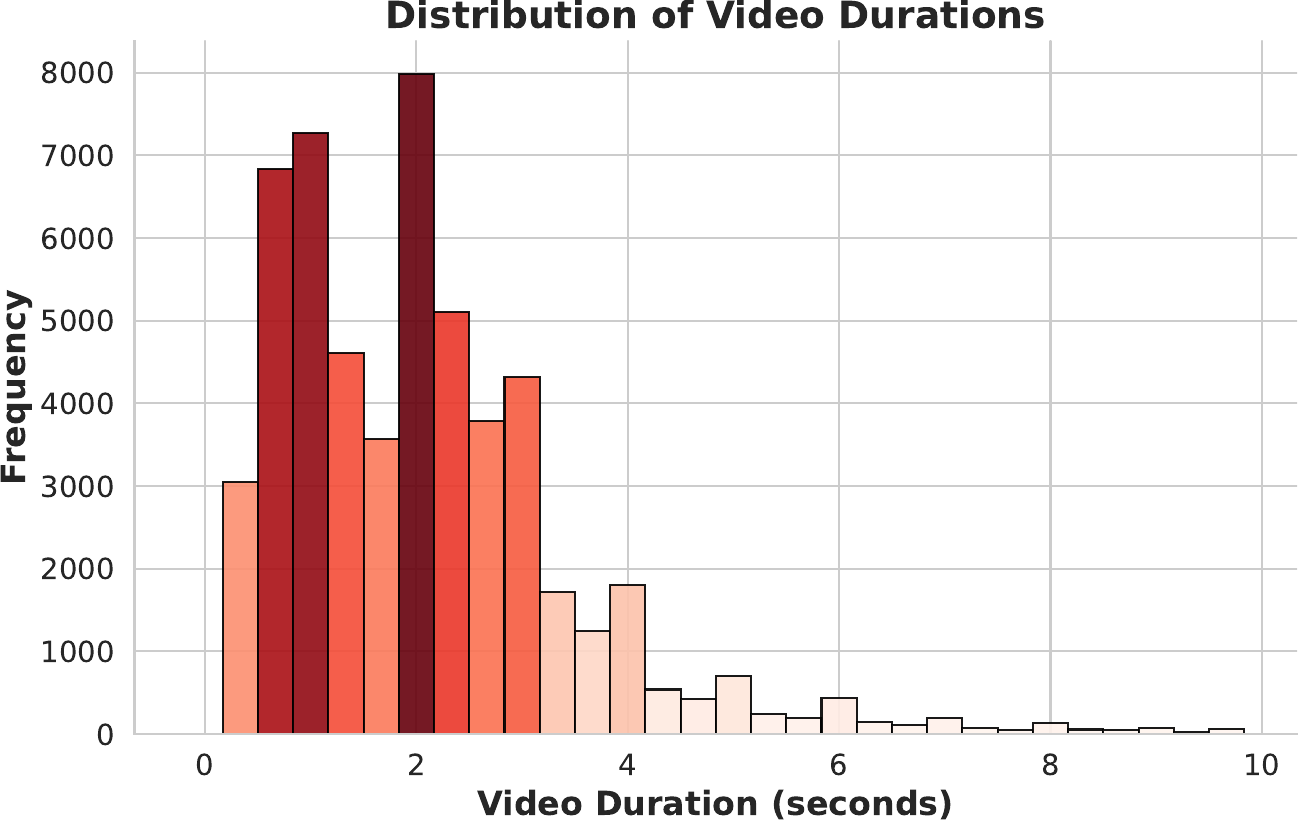}
        \caption{Facial Expression Recognition}
    \end{subfigure}%
    ~ 
    \begin{subfigure}[t]{0.5\textwidth}
        \centering
        \includegraphics[trim={0 0 0 5mm},clip,width=\textwidth]{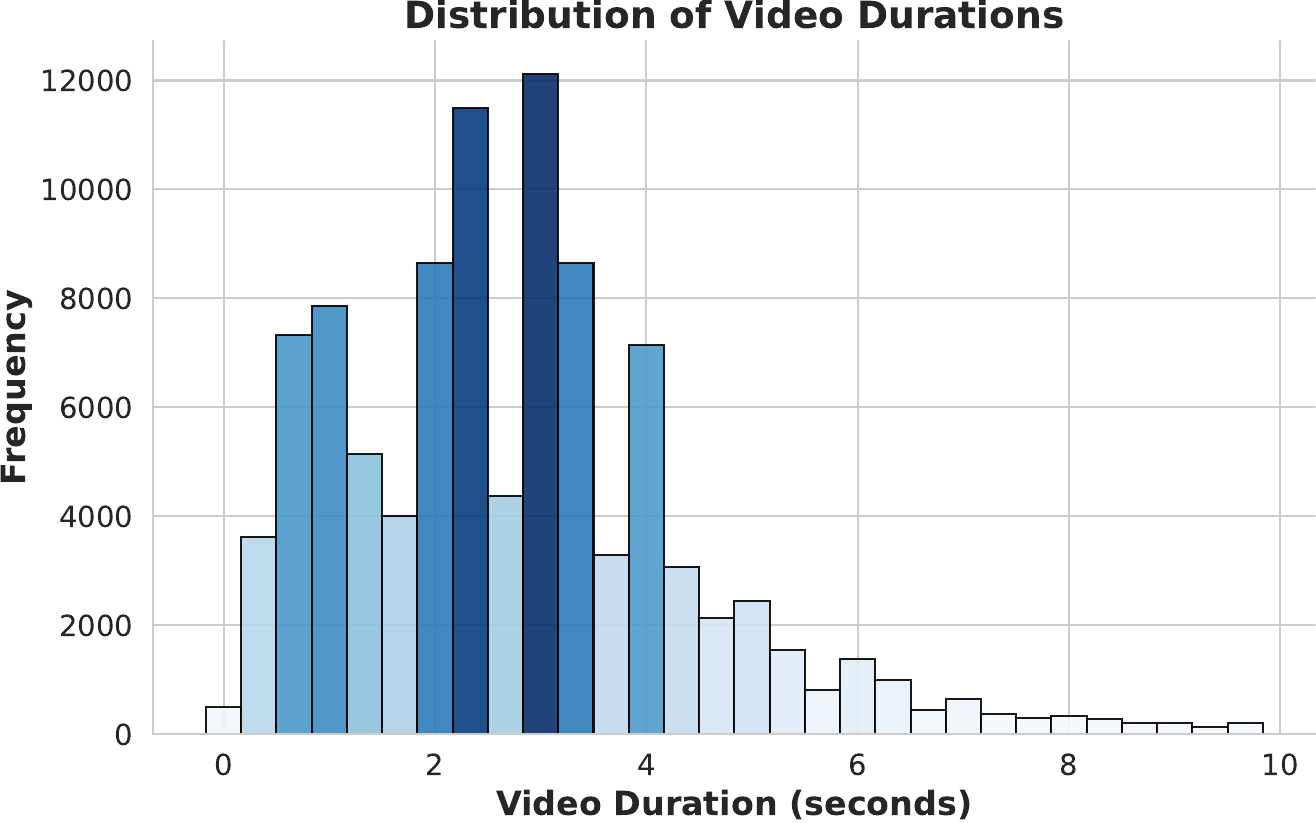}
        \caption{Deepfake Detection}
    \end{subfigure}
    \caption{Distribution of video durations for the facial expression recognition and deepfake detection tasks. Note that except these tasks, other tasks are image/frame-based. For deepfake detection, since we cropped the longer videos of FaceForensics++ \cite{roessler2019faceforensicspp} into chunks of 3 seconds, so we can see a peak around 3 seconds.}
    \label{fig:video_duration_distribution}
\end{figure*}

\begin{figure*}[h!]
    \centering
    \begin{subfigure}[t]{0.19\textwidth}
        \centering
        \includegraphics[width=\textwidth]{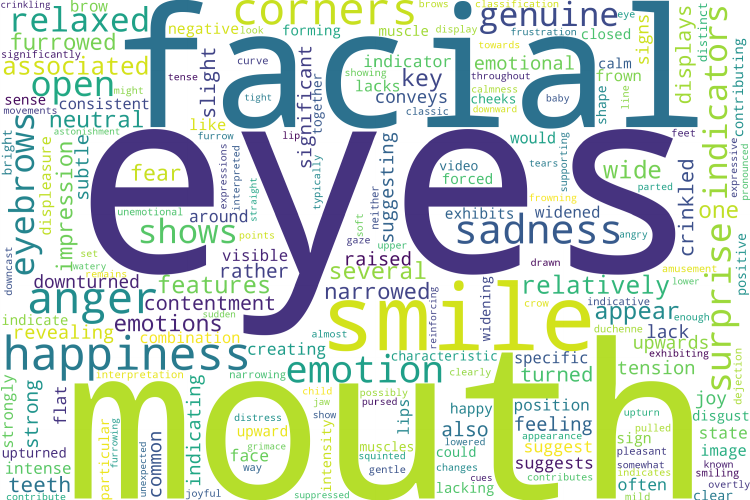}
        \caption{Facial Expression Recog.}
    \end{subfigure}%
    ~ 
    \begin{subfigure}[t]{0.19\textwidth}
        \centering
        \includegraphics[width=\textwidth]{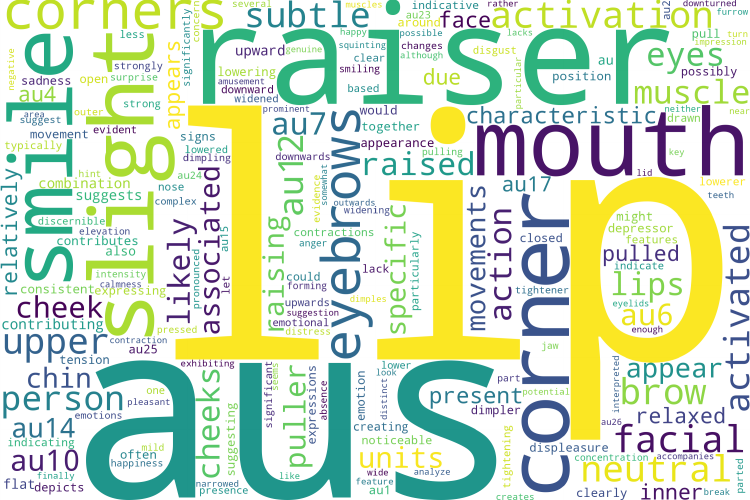}
        \caption{Action Unit Detection}
    \end{subfigure}
    \begin{subfigure}[t]{0.19\textwidth}
        \centering
        \includegraphics[width=\textwidth]{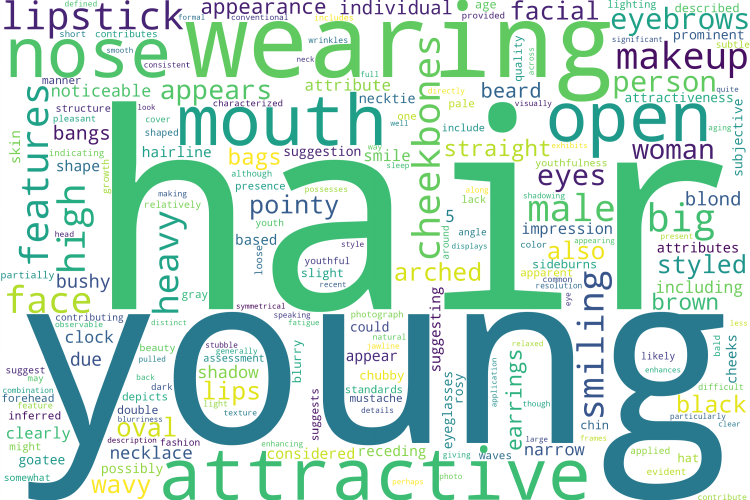}
        \caption{Facial Attribute Detection}
    \end{subfigure}
    \begin{subfigure}[t]{0.19\textwidth}
        \centering
        \includegraphics[width=\textwidth]{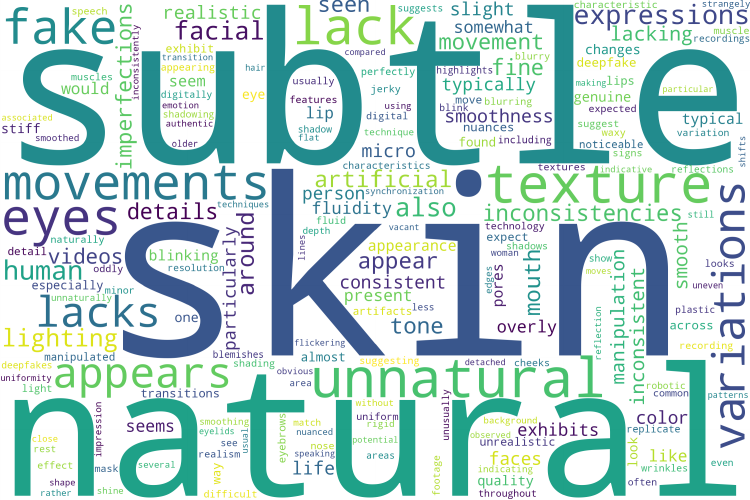}
        \caption{Deepfake Detection}
    \end{subfigure}
    \begin{subfigure}[t]{0.19\textwidth}
        \centering
        \includegraphics[width=\textwidth]{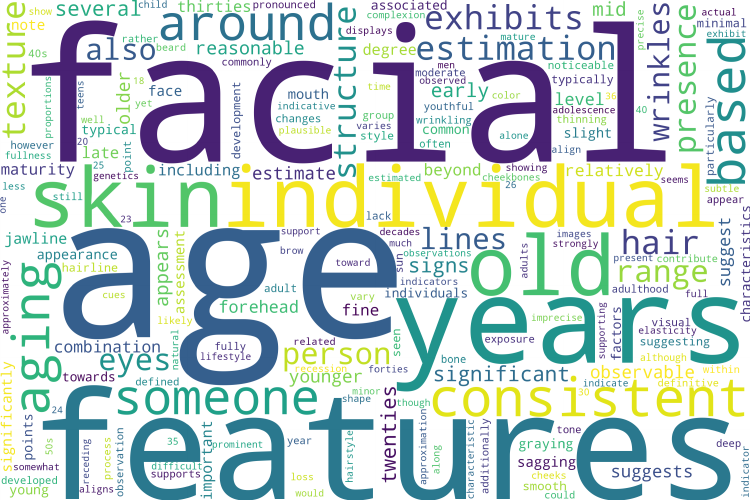}
        \caption{Age Estimation}
    \end{subfigure}
    \caption{Wordcloud for the descriptions or reasons present in \emph{FaceInstruct-1M} belonging to different tasks.}
    \label{fig:word_cloud}
\end{figure*}

\noindent\textbf{Video durations.} \cref{fig:video_duration_distribution} shows the distribution of video durations for video-related tasks in \emph{FaceInstruct-1M}. The total duration of videos for facial expression recognition is around 35 hours, with a mean of 2.28 seconds, and that of deepfake detection is around 84 hours, with a mean of 3.87 seconds. We can observe that the median duration is around 2 seconds for facial expression recognition while it is around 3-4 seconds for deepfake detection. It is also important to mention that since the FaceForensics++ \cite{roessler2019faceforensicspp} dataset has longer videos, we chunk the videos into smaller ones of around 3 seconds.

\begin{figure*}[t!]
    \centering
    \begin{subfigure}[t]{0.49\textwidth}
        \centering
        \includegraphics[width=\textwidth]{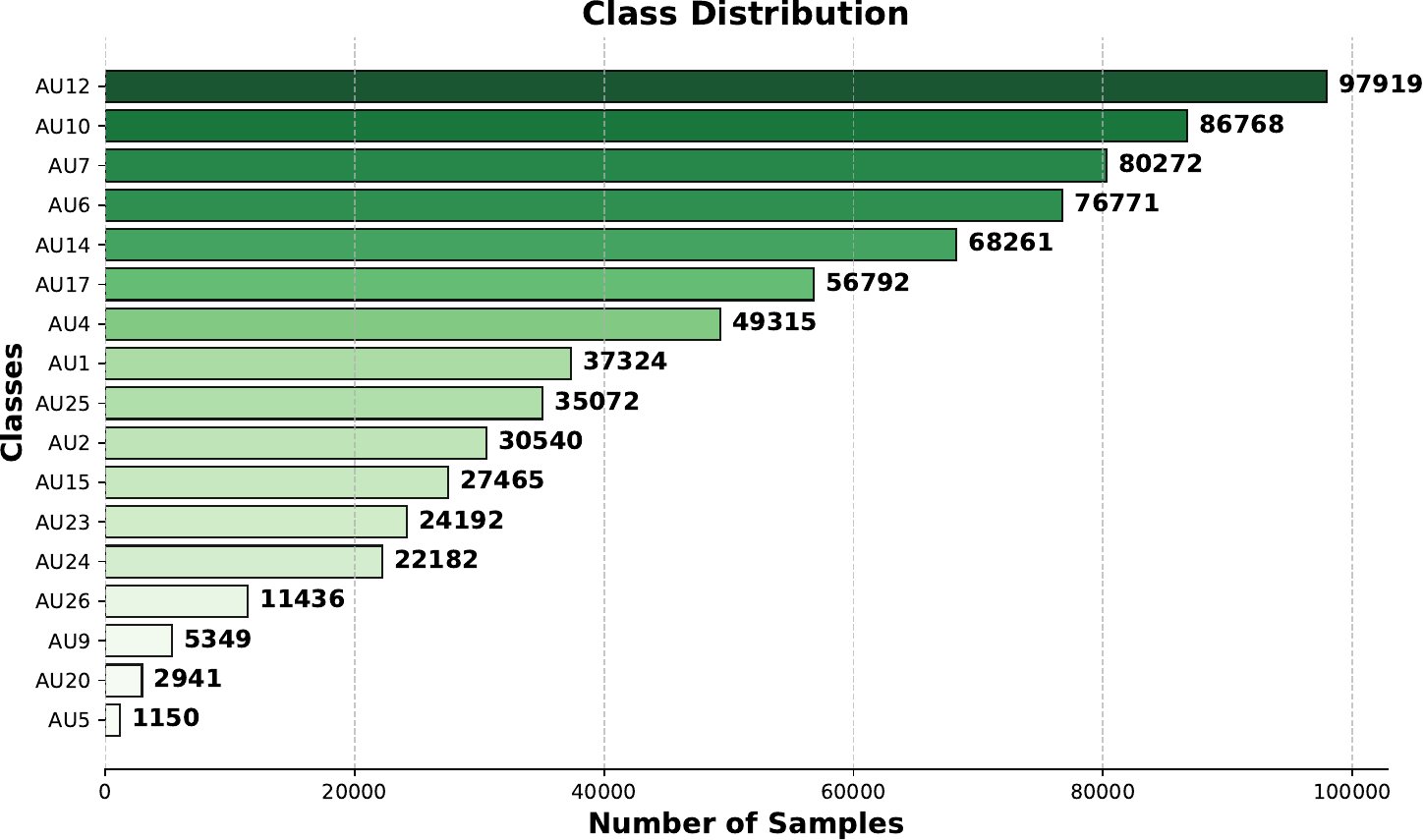}
        \caption{Action Unit Detection}
    \end{subfigure}%
    \begin{subfigure}[t]{0.49\textwidth}
        \centering
        \includegraphics[width=\textwidth]{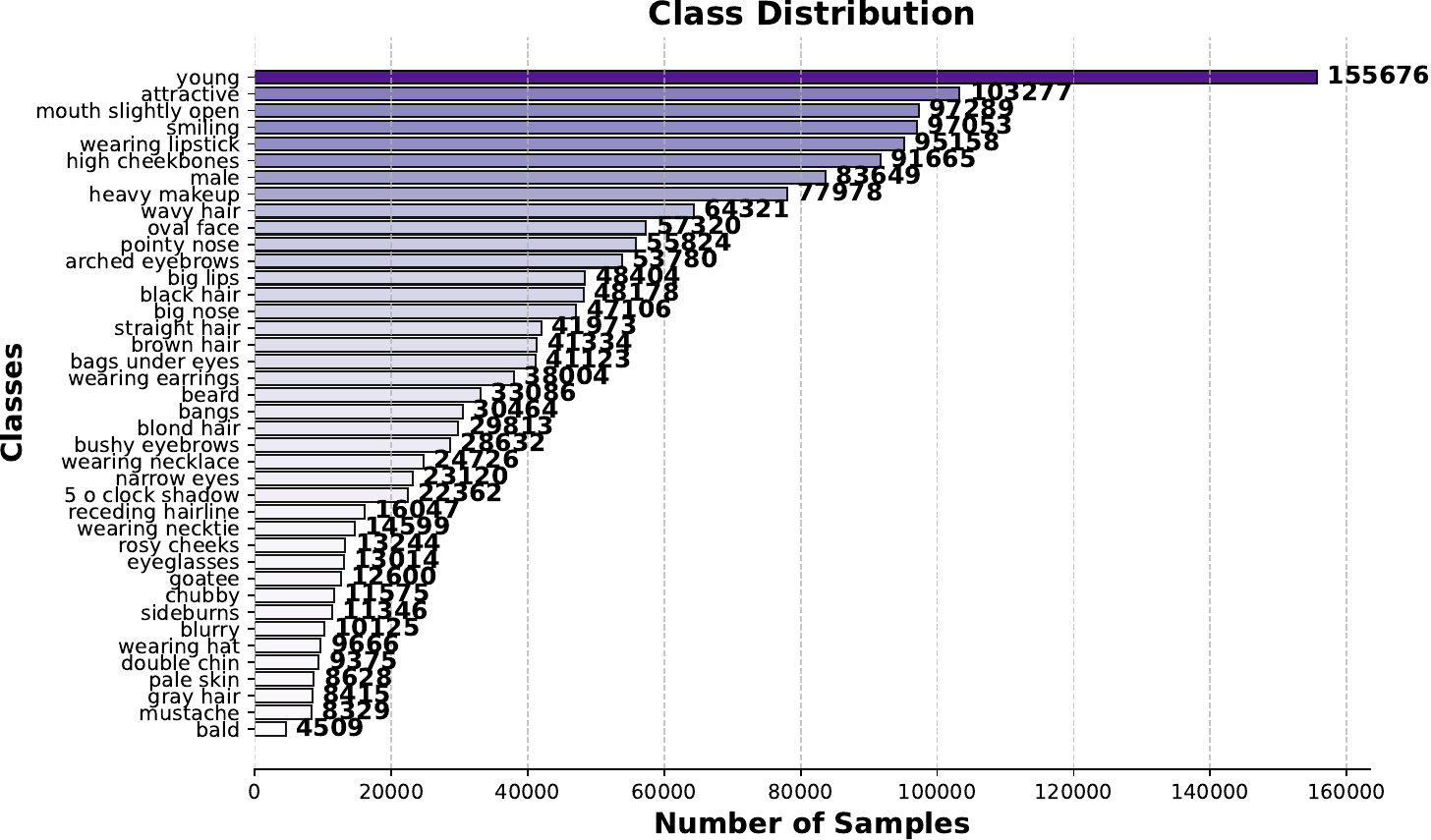}
        \caption{Facial Attribute Detection}
    \end{subfigure}
    \begin{subfigure}[t]{0.32\textwidth}
        \centering
        \includegraphics[width=\textwidth]{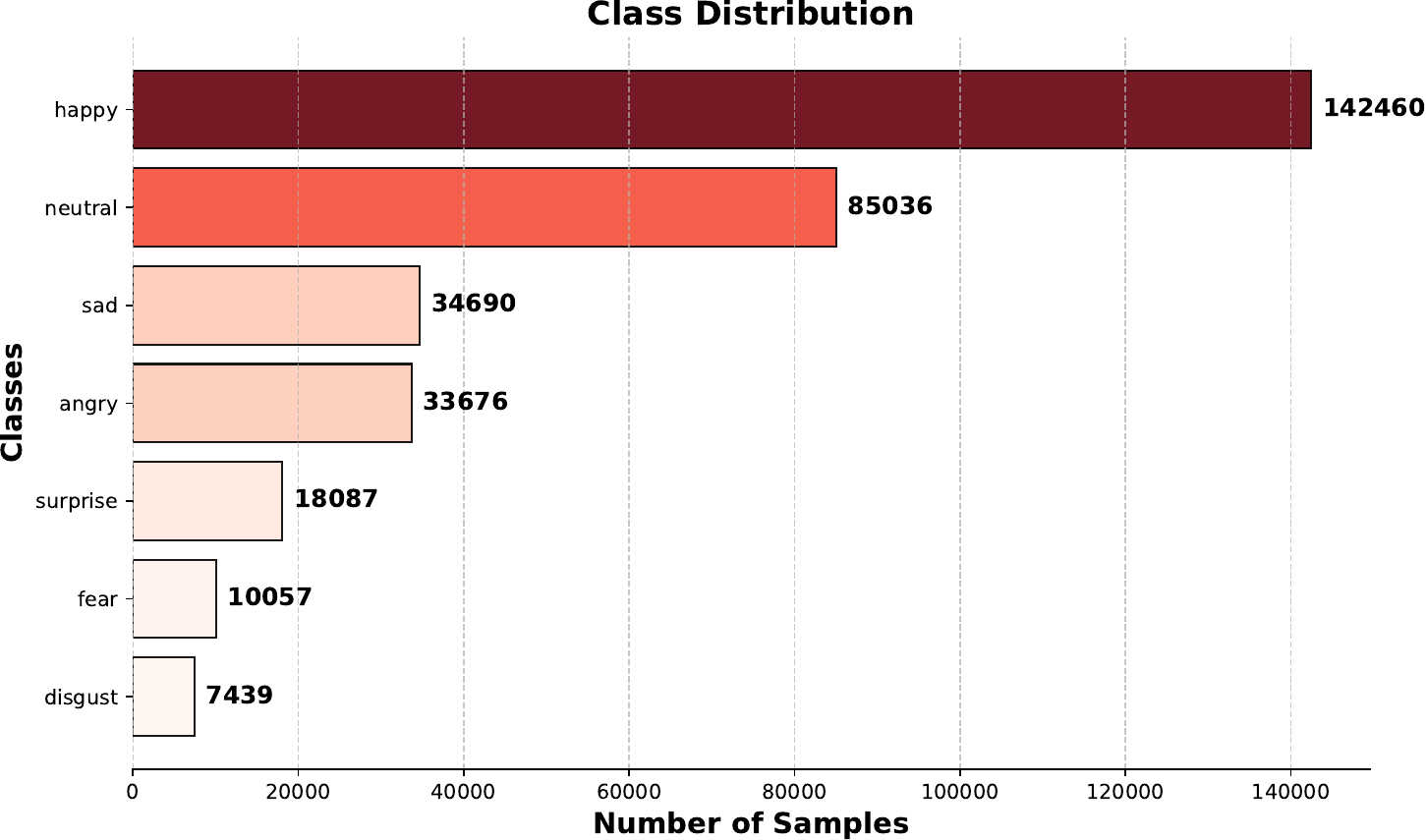}
        \caption{Facial Expression Recognition}
    \end{subfigure}
    \begin{subfigure}[t]{0.32\textwidth}
        \centering
        \includegraphics[width=\textwidth]{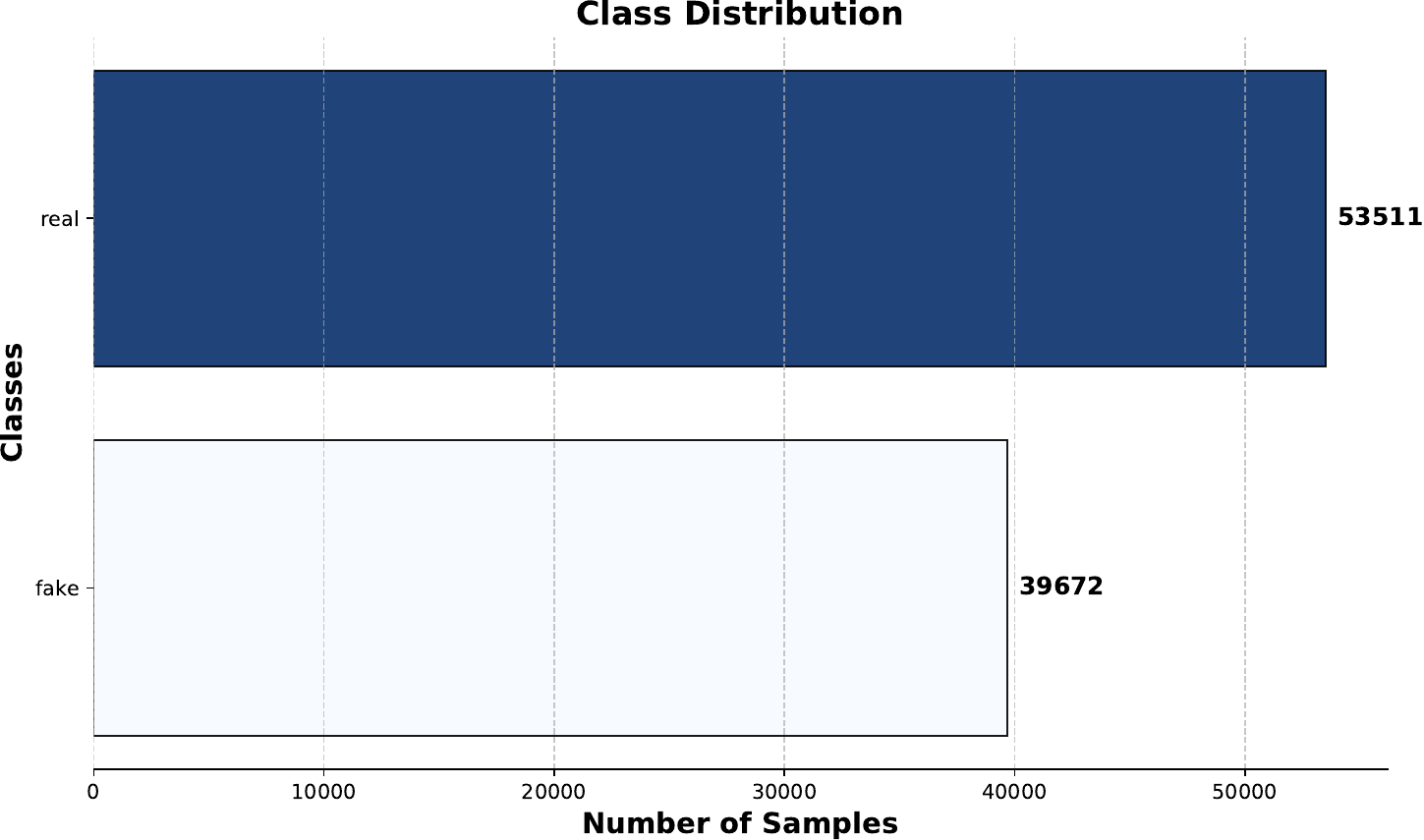}
        \caption{Deepfake Detection}
    \end{subfigure}
    \begin{subfigure}[t]{0.32\textwidth}
        \centering
        \includegraphics[width=\textwidth]{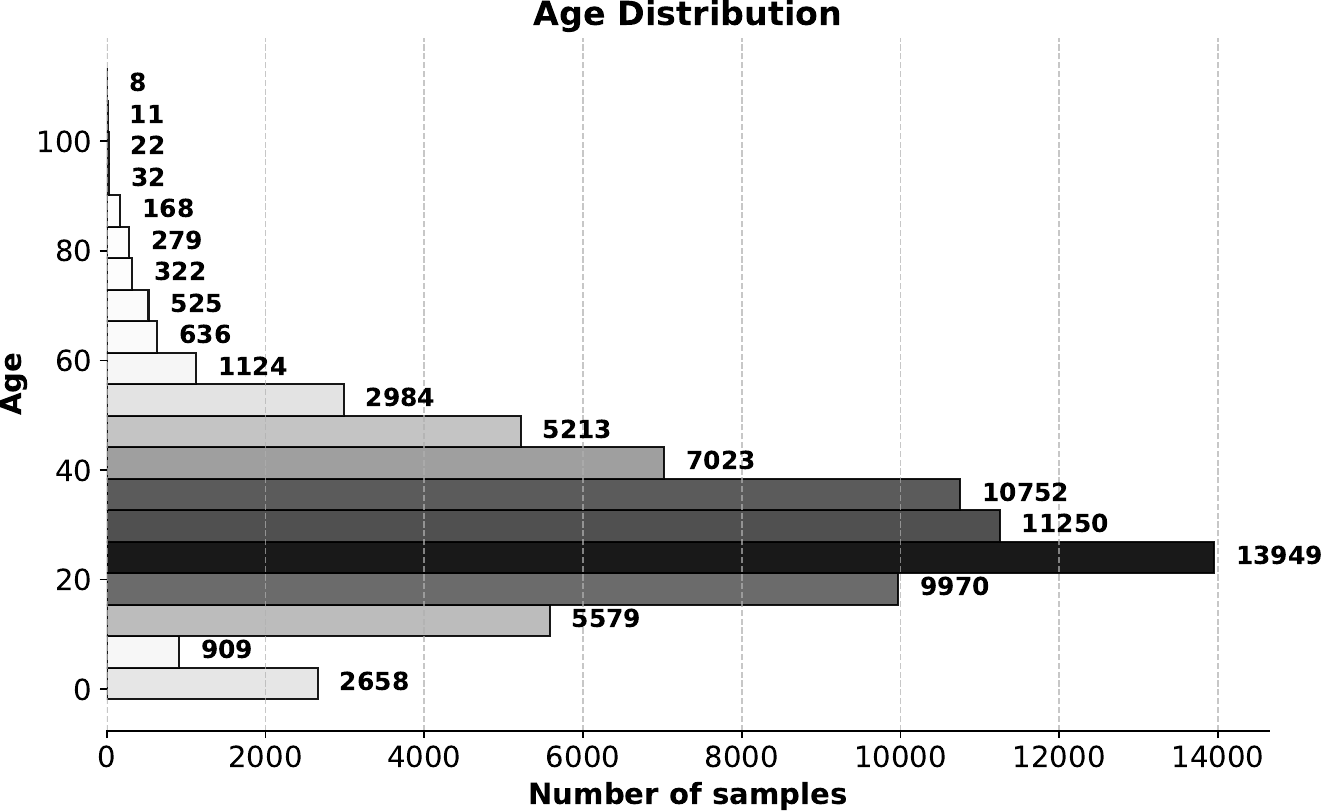}
        \caption{Age Estimation}
    \end{subfigure}
    \caption{Class distribution for different face tasks present in \emph{FaceInstruct-1M}.}
    \label{fig:label distribution}
\end{figure*}

\noindent\textbf{Class Distributions.} \cref{fig:label distribution} summarizes the class distribution for different tasks present in \emph{FaceInstruct-1M}. For AU detection, the majority classes are those classes that are present in both BP4D \cite{ZHANG2014692_bp4d_dataset} and DISFA \cite{disfa_dataset} datasets. AUs such as AU5 and AU20, being only present in one of the datasets, are highly underrepresented in the dataset. Notice that for facial expression recognition, the classes \emph{surprise}, \emph{fear} and \emph{disgust} are highly under-represented, thus explaining their poor recall on zero-shot performance on DFEW \cite{jiang2020dfew} dataset. For deepfake detection, the distribution of \emph{real} and \emph{fake} classes is slightly imbalanced with a bias towards \emph{real} class. For age estimation, since MORPH II \cite{morph_ii_dataset} only contains face images of people in the age of 16-76, hence the number of samples are more in the 20-40 age group in our dataset. It is important to note that the class imbalances present in \emph{FaceInstruct-1M} is also present in the traditional datasets since \emph{FaceInstruct-1M} is constructed using the images/videos and label information from those datasets. 

\noindent\textbf{Word Clouds.} \cref{fig:word_cloud} shows the word clouds generated for the descriptions or reasons for the samples belonging to different tasks present in \emph{FaceInstruct-1M}. Word cloud for facial expression recognition further highlights the bias in the dataset for \emph{happiness} class with the word cloud for \emph{happiness} and \emph{smile} being big. Similarly, the word cloud for facial attribute detection correlates with the class distribution for facial attribute detection. It is important to note that all of the word clouds have bigger clouds for face or face regions (such as mouth, eyes, lip, nose, hair) important for reasoning on that task. For example, for deepfake detection, since the texture of skin plays an important role, so \emph{skin} has a larger cloud.

\subsection{Data samples}

\begin{figure*}
    \centering
    \includegraphics[width=\linewidth]{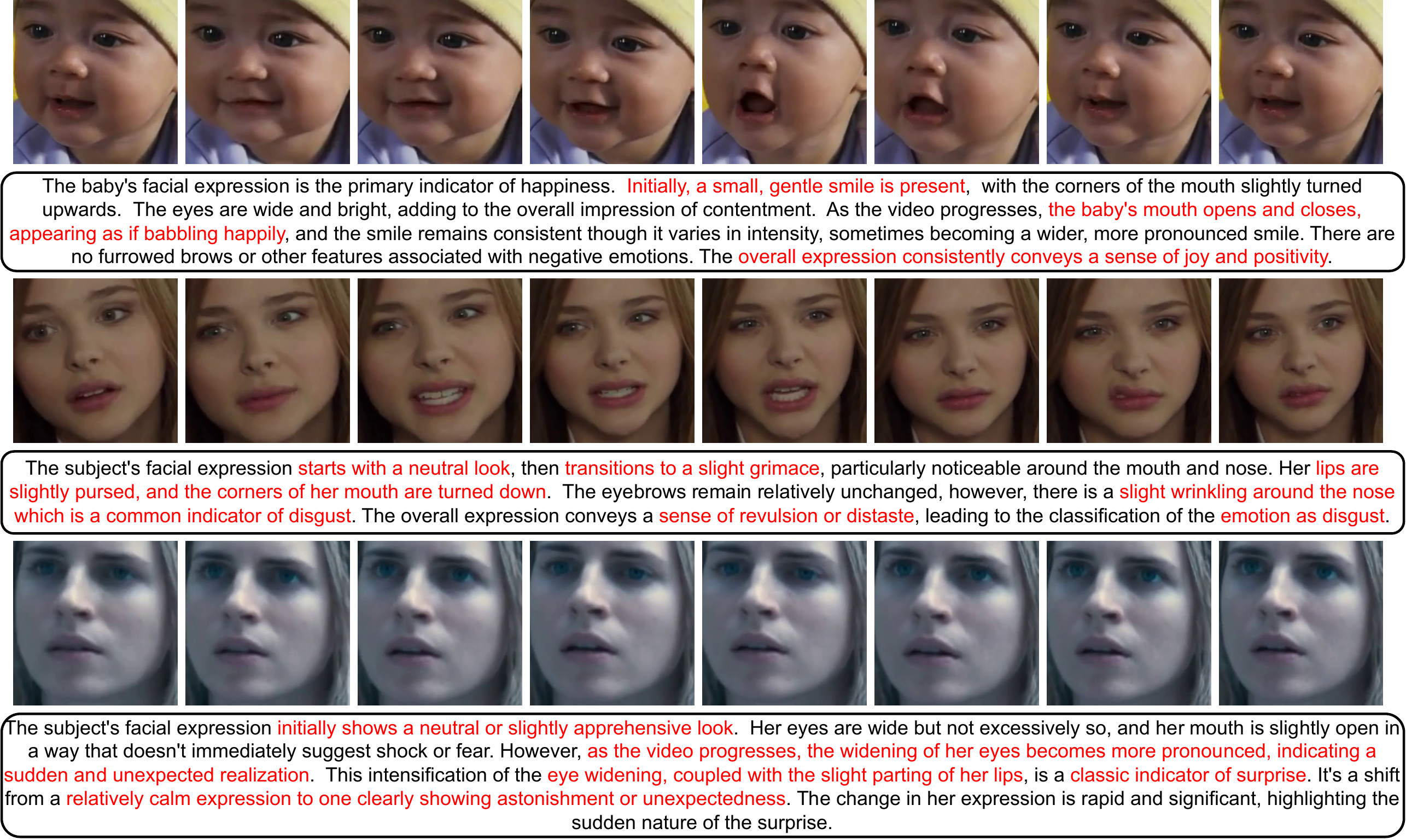}
    \caption{Examples of facial expression recognition task from \emph{FaceInstruct-1M}. }
    \label{fig:data_example_fer}
\end{figure*}

\begin{figure*}
    \centering
    \includegraphics[width=\linewidth]{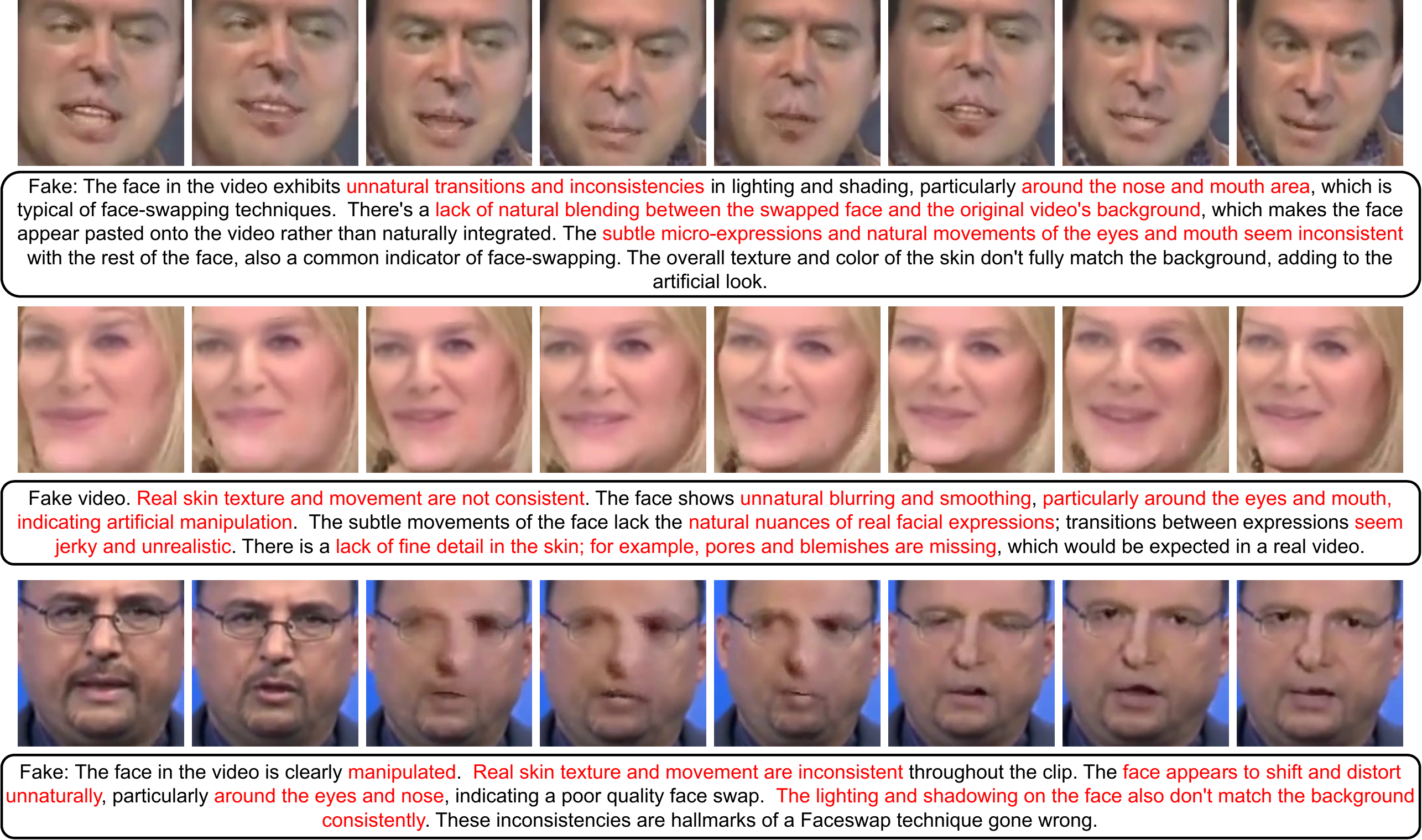}
    \caption{Examples of deepfake detection task from \emph{FaceInstruct-1M}.}
    \label{fig:data_example_ff}
\end{figure*}

\begin{figure*}
    \centering
    \includegraphics[width=\linewidth]{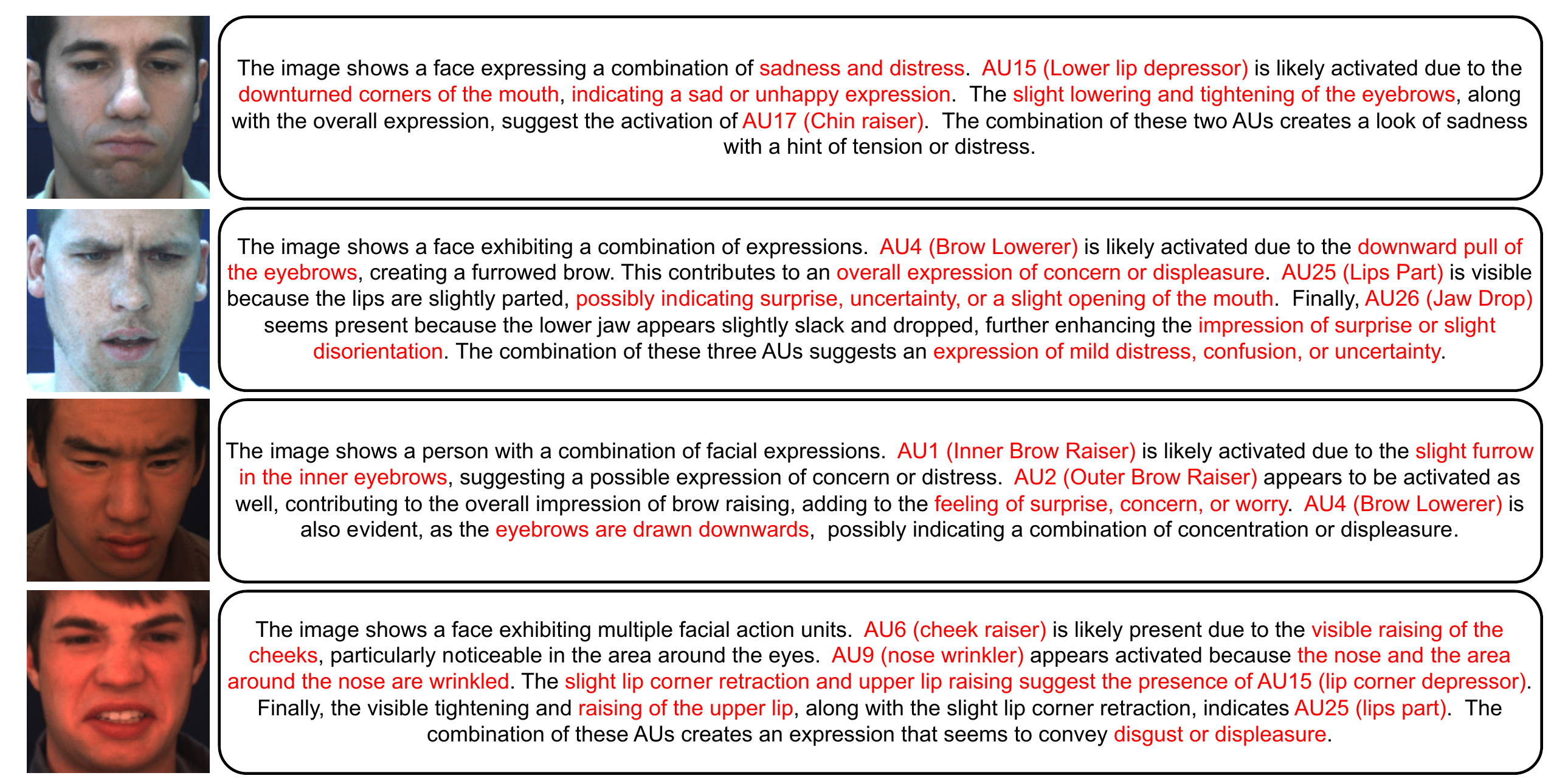}
    \caption{Examples of action unit detection task from \emph{FaceInstruct-1M}.}
    \label{fig:data_example_au}
\end{figure*}

\begin{figure*}
    \centering
    \includegraphics[width=0.6\linewidth]{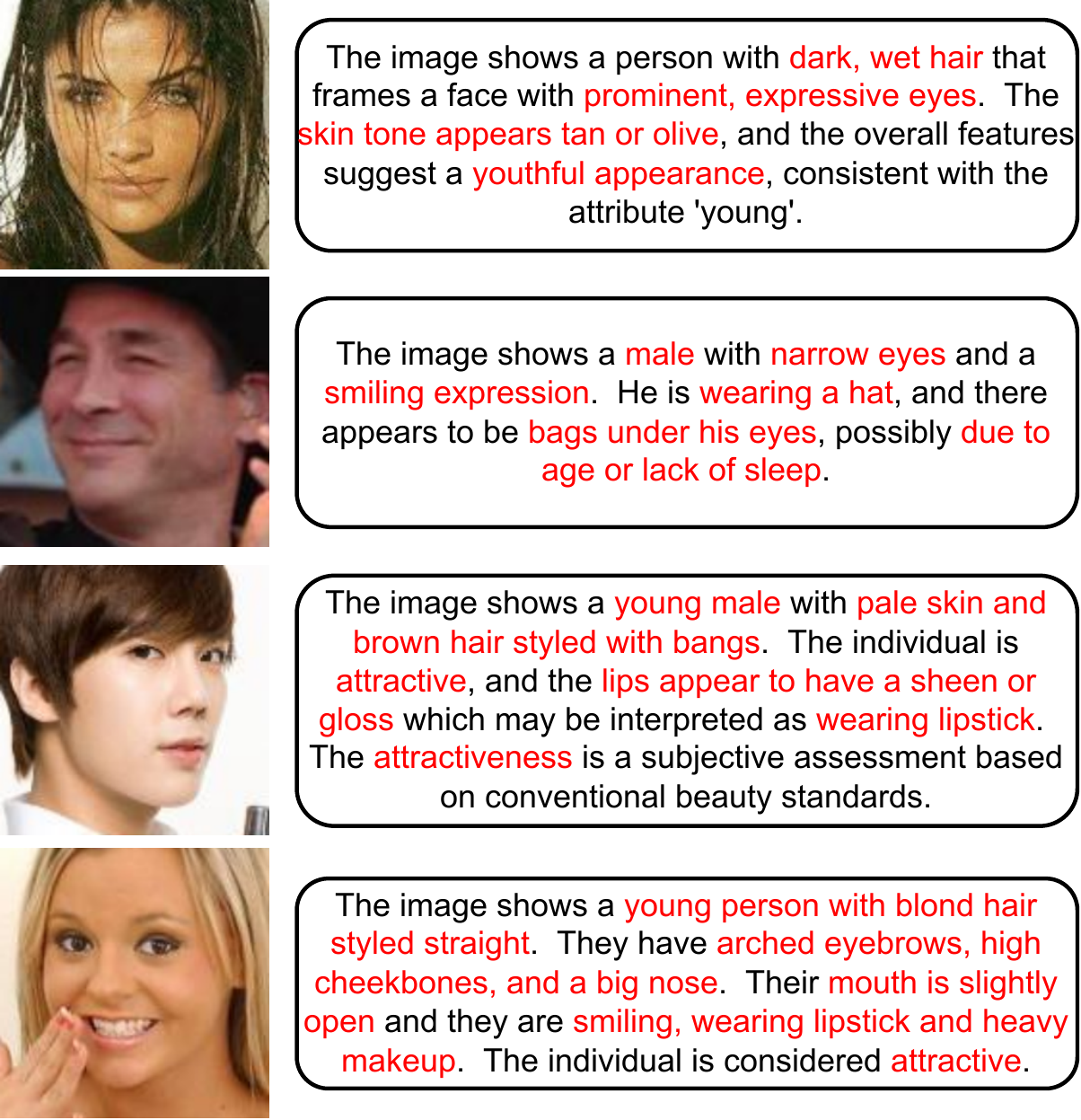}
    \caption{Examples of facial attribute detection task from \emph{FaceInstruct-1M}.}
    \label{fig:data_example_celeba}
\end{figure*}

\begin{figure*}
    \centering
    \includegraphics[width=\linewidth]{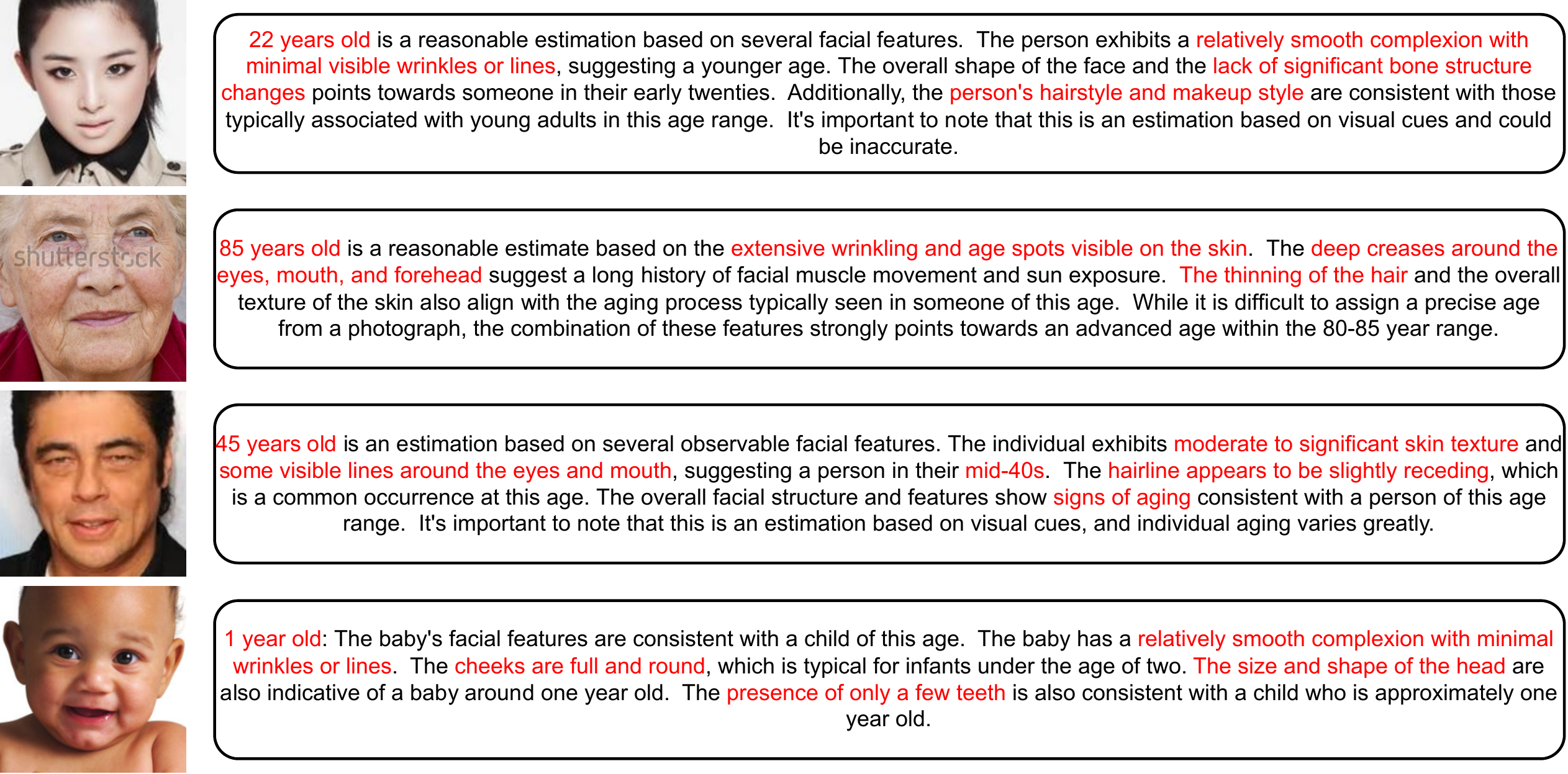}
    \caption{Examples of age estimation task from \emph{FaceInstruct-1M}.}
    \label{fig:data_example_age}
\end{figure*}

\cref{fig:data_example_age,fig:data_example_au,fig:data_example_celeba,fig:data_example_fer,fig:data_example_ff} show samples of reasoning or descriptions for different tasks present in \emph{FaceInstruct-1M}. Notice in \cref{fig:data_example_fer} that the annotations not only capture the overall emotion of the video, but they also contain information about how the facial expression and facial movements varied throughout the video to reason the overall emotion. For facial expression recognition, it is also important to note that in addition to the 7 emotion categories, the annotations also capture sub-emotions or feelings such as revulsion, distaste, etc. 

For deepfake detection (\cref{fig:data_example_ff}), the annotations successfully capture the necessary imperfections or alterations that are visible in the video frames. Moreover, we can also observe that such descriptions incorporate references to different face regions which highlights the importance of different face regions for different face tasks.

For action unit detection (\cref{fig:data_example_au}), the annotations capture the overall facial expression in addition to the action units or facial muscle movements responsible for the facial expressions. Thus, such annotations not only contain information about action units activated in the image, but also provide pseudo data for facial expression recognition. This shows how annotations for one task may help improving performance in a related task.

In \cref{fig:data_example_celeba}, we can see that the annotations for facial attribute detection are pretty short and straightforward as they capture features or attributes that are clearly visible in the image without the need to "reason" a particular aspect of the image. \cref{fig:data_example_age} shows the annotations for age estimation task. For this task, notice that since we have prompted Gemini to start descriptions with the ground truth age (see \cref{fig:gemini_annotation_prompts}), so all the descriptions start with the ground truth age. This is done to ease string parsing as mentioned later in \cref{subsec:string_parsing}. 

\clearpage

\section{Evaluation}
\label{sec:supp_evaluation_protocol}

\begin{table*}[!t]
\centering
\resizebox{\textwidth}{!}{%
\begin{tabular}{ccccc|ccccc|cc}
\hline \hline
\rowcolor[HTML]{C0C0C0} 
\multicolumn{5}{c|}{\cellcolor[HTML]{C0C0C0}\textbf{Expression Recognition}} & \multicolumn{5}{c|}{\cellcolor[HTML]{C0C0C0}\textbf{Attribute Detection}} & \multicolumn{2}{c}{\cellcolor[HTML]{C0C0C0}\textbf{Deep-Fake Detection}} \\ \hline
\rowcolor[HTML]{C0C0C0} 
\textbf{Happiness} & \textbf{Sadness} & \textbf{Neutral} & \textbf{Anger} & \textbf{...} & \textbf{Attractive} & \textbf{Chubby} & \textbf{Rosy Cheeks} & \textbf{Young} & \textbf{...} & \textbf{Real} & \textbf{Fake} \\ \hline
cheerful & crying & calm & annoyed &  & appealing & plump & blushed cheeks & childish &  & authentic & fabricated \\
content & distress & expressionless & enraged &  & beautiful & puffy face & flushed cheeks & juvenile &  & genuine & forged \\
joy & melancholy & unemotional & incensed &  & good looking & soft cheeks & pinkish cheeks & teenager &  & legitimate & fraudulent \\
smiling & sob & unmoving & mad &  & handsome & round face & red cheeks & youthful &  & original & manipulated \\
... & ... & ... & ... &  & ... & ... & ... & ... &  & ... & ... \\ \hline \hline
\end{tabular}%
}
\caption{Synonyms used for categorizing descriptions into labels for different tasks. We have not shown the complete list of synonyms for all the classes to keep the table succinct.}
\label{tab:supp_synonyms}
\end{table*}

\subsection{Synonyms matching}
\label{subsec:synonyms_used}

As mentioned in \cref{subsec:dataset_evaluation}, to evaluate the performance of the text generation MLLMs such as Face-LLaVA on traditional face analysis benchmarks, we need to convert the generated text into a prediction label. To that extent we follow synonym matching similar to \cite{li2024facialfaba} for facial expression recognition, facial attribute detection and deepfake detection, to categorize the given reason or description to one of the classes. We analyzed the top words from the annotated descriptions of different classes across all tasks and compiled a list of mutually exclusive synonyms for each class, as shown in \cref{tab:supp_synonyms}. For each of the classes within the before-mentioned tasks, we come up with a list of at least 10 synonyms to match to based on the top words occuring in descriptions of our dataset. 

Similar to FABAInstruct \cite{li2024facialfaba}, to map a given description to a class, we first remove all the negative sentences from the description. Then, we first match for synonyms on the first sentence of the description. If there are matches in the first sentence, then we output the majority voted class to be the dominant class in the given description. If the first sentence did not result in any synonym matches, then we perform majority voting on the entire description. The intuition behind matching the first sentence first is that the response from MLLMs starts with a conclusion or summary sentence and later sentences contain detailed description and reasoning over the first sentence.

While synonyms matching works in an expected way for facial expression recognition and facial attribute detection, manual verification of the descriptions revealed that for deepfake detection the model response might sometimes contain synonyms related to the wrong class more than the predicted class. Since we use majority rating, this would result in the response getting classified to the wrong class even if the response predicts the correct predicted class. To overcome this challenge, we explicitly prompt to start the description of the current sample with the ground truth label of the image (see \cref{fig:gemini_annotation_prompts}). While this restricts the description quality in terms of variety in responses, it makes automatic string parsing to extract model prediction from the description easier. Moreover, the goal of this work is not to show that the model can generate varied responses, but rather to exhibit the reasoning capabilities of MLLMs for face-related tasks. Hence, training with such annotations makes Face-LLaVA generate responses that are easier to parse automatically.

\subsection{String parsing}
\label{subsec:string_parsing}

For the age estimation and deepfake detection tasks, we simply use regex parsing to convert the given description to categorical or numerical labels (after removing the negative sentences from the description). This works quite well for the AU detection task in fetching the FACS codes from the description, however, for age estimation, sometimes the description contains numerical values other than what the model wants to predict (see \cref{fig:utkface_samples_comparison_baselines}). Similar to the previous paragraph for deepfake detection, during data annotation through Gemini, we explicitly prompt to start the description of the current sample with the ground truth age of the person (see \cref{fig:gemini_annotation_prompts}). 

\subsection{Traditional metrics and evaluation protocol}
\label{subsec:traditional_metrics}
\noindent\textbf{Facial Expression Recognition.} For DFEW \cite{jiang2020dfew} and Crema-D \cite{cremad_dataset} datasets, similar to previous works \cite{dfew_m3dfel,dfew_s2d,xiang2024mtcaedfer,cremad_pthnet}, we report weighted average recall (WAR) and unweighted average recall (UAR), as the class distribution in these datasets are quite imbalanced. WAR captures the model's ability to perform well on the majority classes of the dataset. It is calculated as the weighted sum of the recall scores for each class and the weights are determined by the number of samples belonging to a class. Unweighted average recall captures model's ability to perform well on all the classes including the under-represented classes. It is computed as the average of the recall scores for each class. For DFEW \cite{jiang2020dfew}, we used the official five-fold cross validation splits to report all the numbers and for Crema-D \cite{cremad_dataset} we perform subject-exclusive 5-fold cross validation to report the results consistent with the baselines \cite{xiang2024mtcaedfer,cremad_pthnet}. For RAF-DB \cite{rafdb} dataset, similar to the previous baselines \cite{li2024facialfaba,rafdb_apvit,transfer_rafdb} we report just the overall accuracy on the official test set consisting of about 3k examples. 

\noindent\textbf{Action Unit Detection.} In line with the previous works we perform subject exclusive cross validation for reporting our results on BP4D \cite{ZHANG2014692_bp4d_dataset} and DISFA \cite{disfa_dataset}. For both the datasets, we report the average F1 score over the set of possible AUs. For DISFA, the available AUs are AU1, AU2, AU4, AU6, AU9, AU12, AU25 and AU26, while for BP4D the AU list contains AU1, AU2, AU4, AU6, AU7, AU10, AU12, AU14, AU15, AU23 and AU24. 

\noindent\textbf{Facial Attribute Detection.} We report the mean accuracy over each of the 40 facial attributes in the test set of CelebA \cite{attr_liu_celeba} dataset containing around 20k images.

\noindent\textbf{Age Estimation.} For both MORPH II \cite{morph_ii_dataset} and UTKFace \cite{zhifei2017cvpr_utkface} datasets, we report the mean absolute error between the ground truth and the predicted integer ages on the official test splits for both the datasets. For MORPH II, the test set consists of around 5k images and for UTKFACE the test set consists around 4k images.

\noindent\textbf{Deepfake Detection.} We use the official test split of FaceForensics++ \cite{roessler2019faceforensicspp} dataset to report our numbers. We only use the low quality samples from the dataset for both constructing the \emph{FaceInstruct-1M} dataset (and training) and testing. For all the zero-shot MLLM baselines, we chunk the test videos into a constant duration of three seconds and compute the prediction of a video as the majority voted prediction over its constituent chunks. Since for the MLLM baselines and Face-LLaVA, we can only get a text output for this task, so we only report the accuracy on the test set. Note that, the test set of FF++ is imbalanced and has 80\% fake videos and only 20\% real videos, hence making 80\% accuracy as a baseline.

\begin{figure*}
    \centering
    \includegraphics[width=0.7\linewidth]{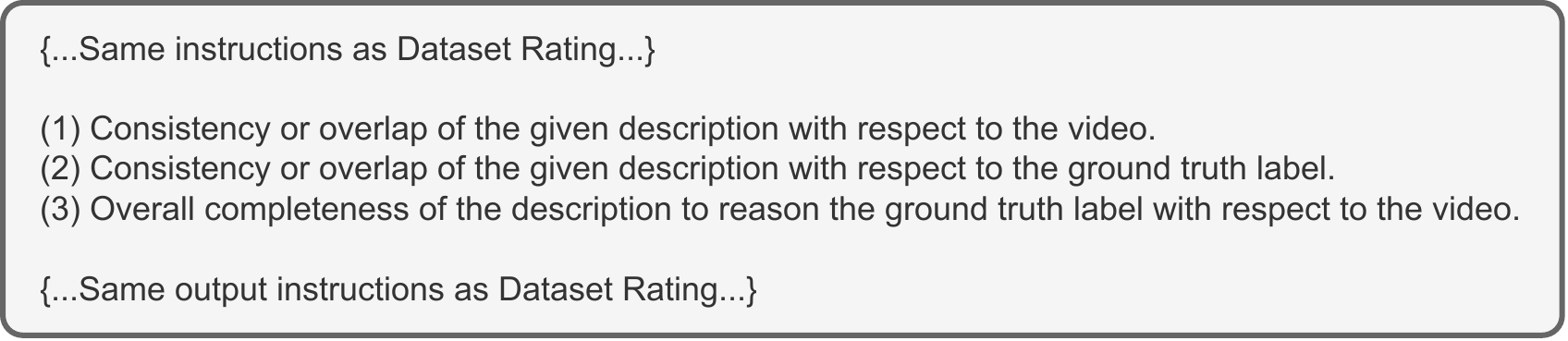}
    \caption{Instructions for GPT4o-mini automatic evaluation of the reasoning capabilites of different models. Note that we only change the rating criterias from \cref{fig:gpt_filtering_prompts} to the ones mentioned above to obtain GPT-ratings for the given reasoning outputs for different models.}
    \label{fig:gpt_evaluation}
\end{figure*}

\subsection{GPT-Evaluation}

As mentioned in \cref{subsec:evaluation_protocol}, we employ GPT4o-mini \cite{gpt4omini} for automatically evaluating the reasoning capabilities of baselines in comparison to Face-LLaVA. The instructions (prompts) used for this evaluation are similar to those described in \cref{fig:gpt_filtering_prompts} for dataset rating and filtering. We only change the criterias for evaluating the descriptions to those described in \cref{fig:gpt_evaluation} for this evaluation. 

\section{Implementation details}
\label{sec:supp_implementation_details}

\subsection{Face-LLaVA}

To implement our model, we start with the baseline architecture of Video-LLaVA \cite{lin-etal-2024-videollava} which has a Vicuna-7B backbone and add our novel FRLP and FRGCA modules (as described in \cref{sec:face_llava_method}) on top of it. To keep the extra computation introduced by the projection layers minimal, we use single layer MLPs within FRLP, as opposed to a 2-layer MLP with GeLU used in the vision projector \cite{lin-etal-2024-videollava}. Moreover, we only use a single block of cross-attention inside FRGCA with 8 attention heads. We do not apply layer normalization to the output of FRGCA. Maximum context window of the LLM is 2048 tokens and maximum context window of the tokenizer is 3072. 

For training, we use deepspeed\footnote{\href{https://www.deepspeed.ai/}{https://www.deepspeed.ai/}} to parallelize training on a single NVIDIA DGX node with 8*H100 GPUs. We train the model using an AdamW optimizer only for one epoch for both the stages. Face-Region Pretraining takes about 5 hours and Finetuning takes around 12 hours on the entire dataset for the abovementioned environment.  Learning rates for the pretraining and finetuning stages are kept as 1e-4 and 2e-5 respectively with cosine learning rate schedule.

\subsection{Baselines}
\label{subsec:implementation_baselines}

We use the official inference code of the baselines. For models which can handle both video and images, we use the corresponding inference code for video and image related tasks. To ensure a fair comparison, for models which have an option to set the fps \cite{damonlpsg2025videollama3,qwen25vl,zhang2024llava_video} for extracting frames for video inputs, we set the fps in such a way that we extract either 8 frames or frames at 1 fps, whichever leads to higher number of frames. 

Since we do not finetune the baseline MLLMs and simply use their pretrained weights for inference, we engineer our prompts for each task and model pair to ensure optimum results. For fixed-set prediction tasks such as facial expression recognition, facial attribute detection and deepfake detection, we provide the list of possible classes in the prompt to supervise and restrict the model responses. Moreover, for all the tasks we use a separate prompt for reasoning evaluation and evaluation using traditional metrics. This is done because the responses generated by MLLMs are sometimes hard to parse with the techniques mentioned in \cref{subsec:string_parsing,subsec:synonyms_used}. So to report the performance of baselines for \cref{subsec:traditional_evaluation_results} we explicitly prompt the baselines to generate its response as a single word or a list of words.

\section{Reasoning comparison with baselines}
\label{subsec:reasoning_comparison_samples_supplementary}
\begin{figure}[t!]
    \centering
    \begin{subfigure}[t]{0.49\textwidth}
        \centering
        \includegraphics[width=\textwidth]{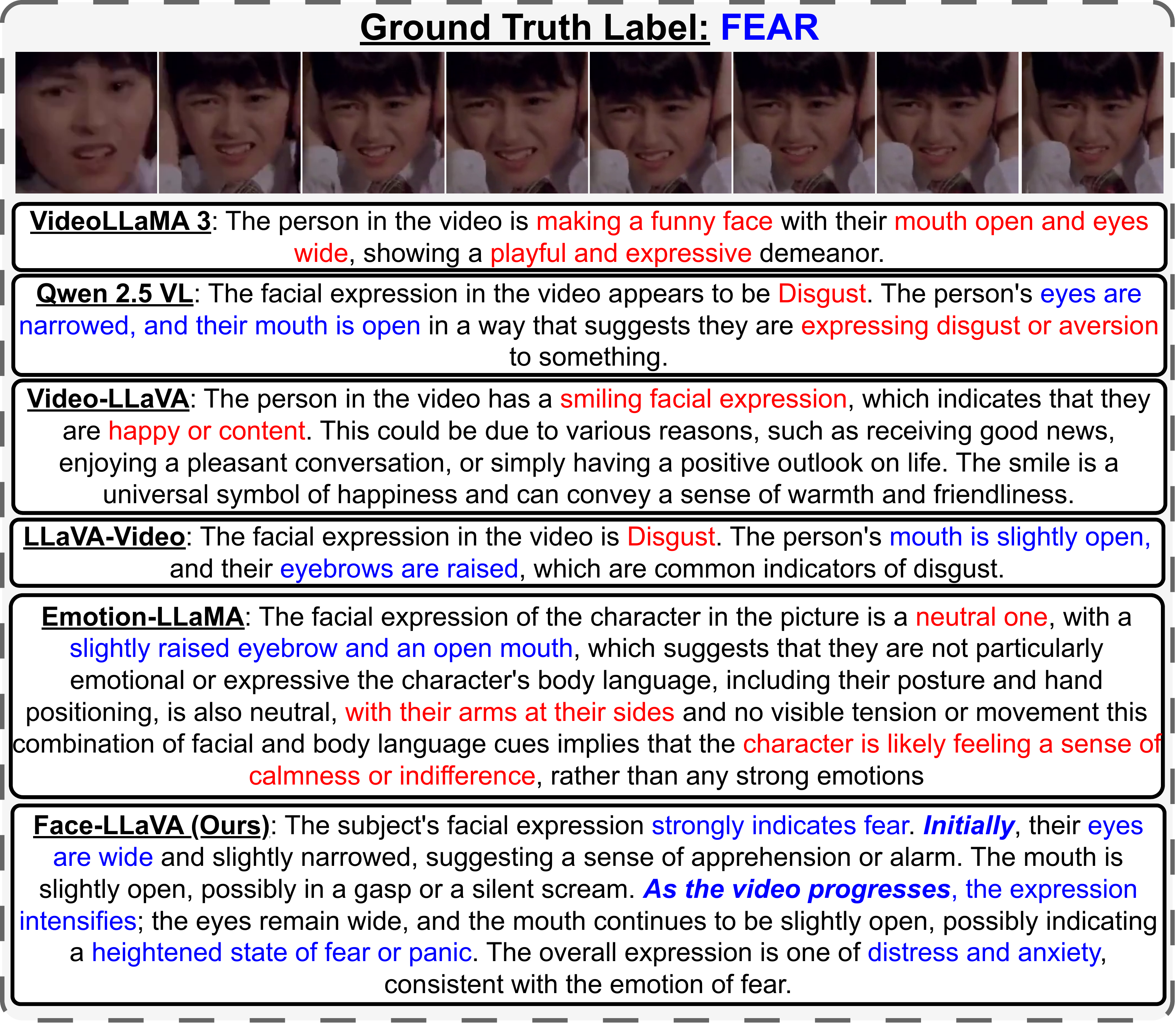}
    \end{subfigure} 
    \hfill
    \begin{subfigure}[t]{0.49\textwidth}
        \centering
        \includegraphics[width=\textwidth]{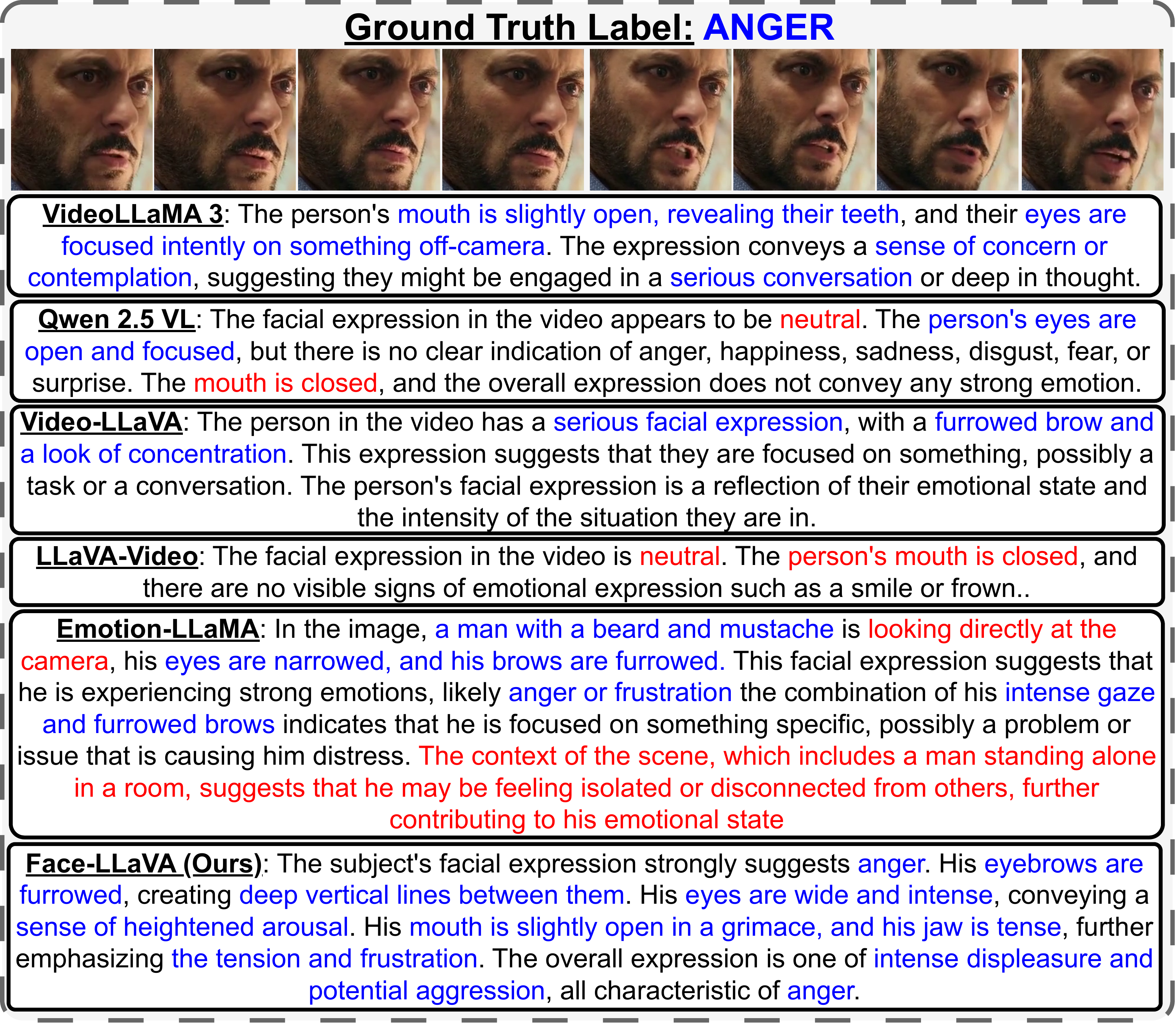}
    \end{subfigure} \\
    \begin{subfigure}[t]{0.49\textwidth}
        \centering
        \includegraphics[width=\textwidth]{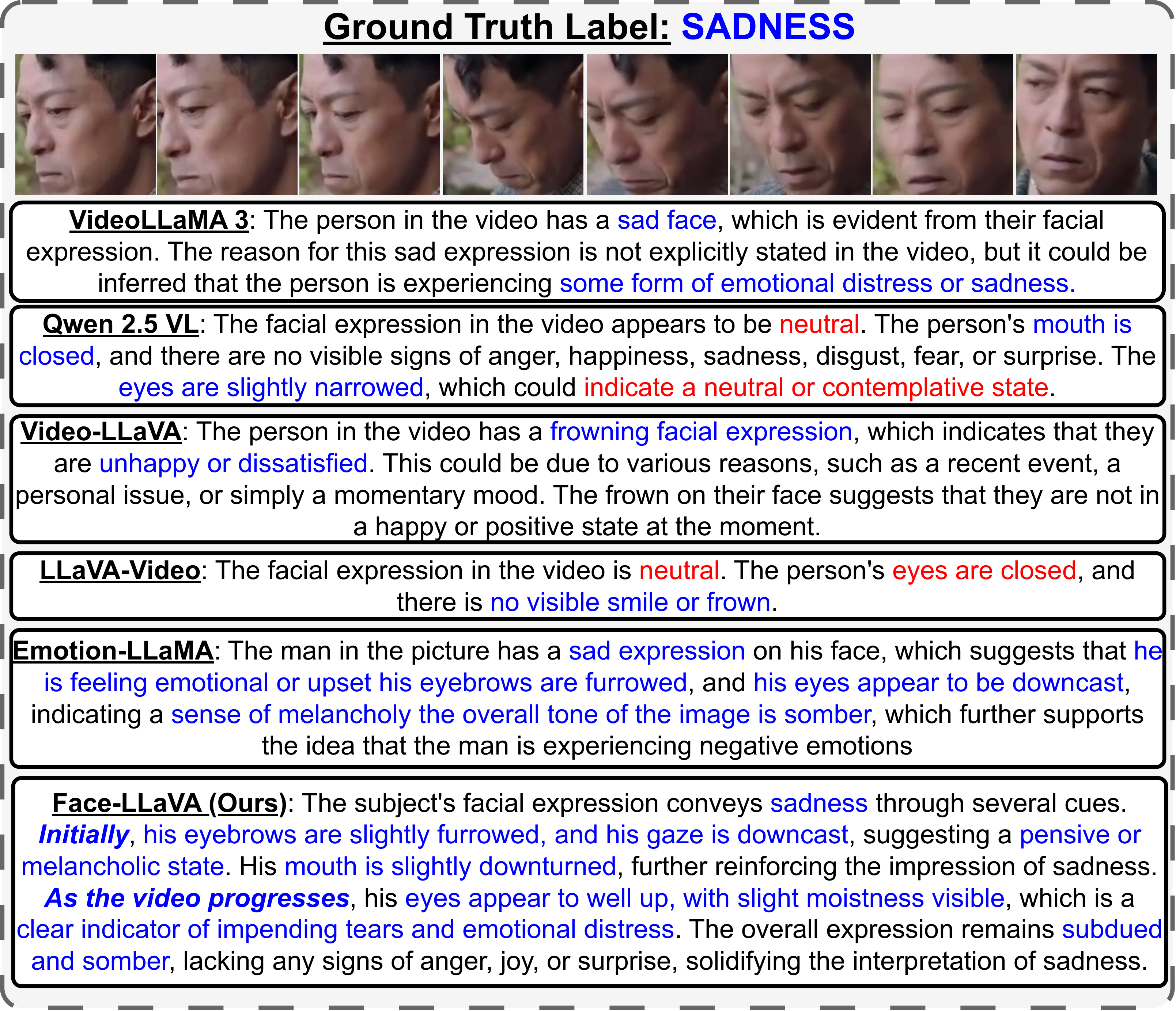}
    \end{subfigure}%
    \hfill
    \begin{subfigure}[t]{0.49\textwidth}
        \centering
        \includegraphics[width=\textwidth]{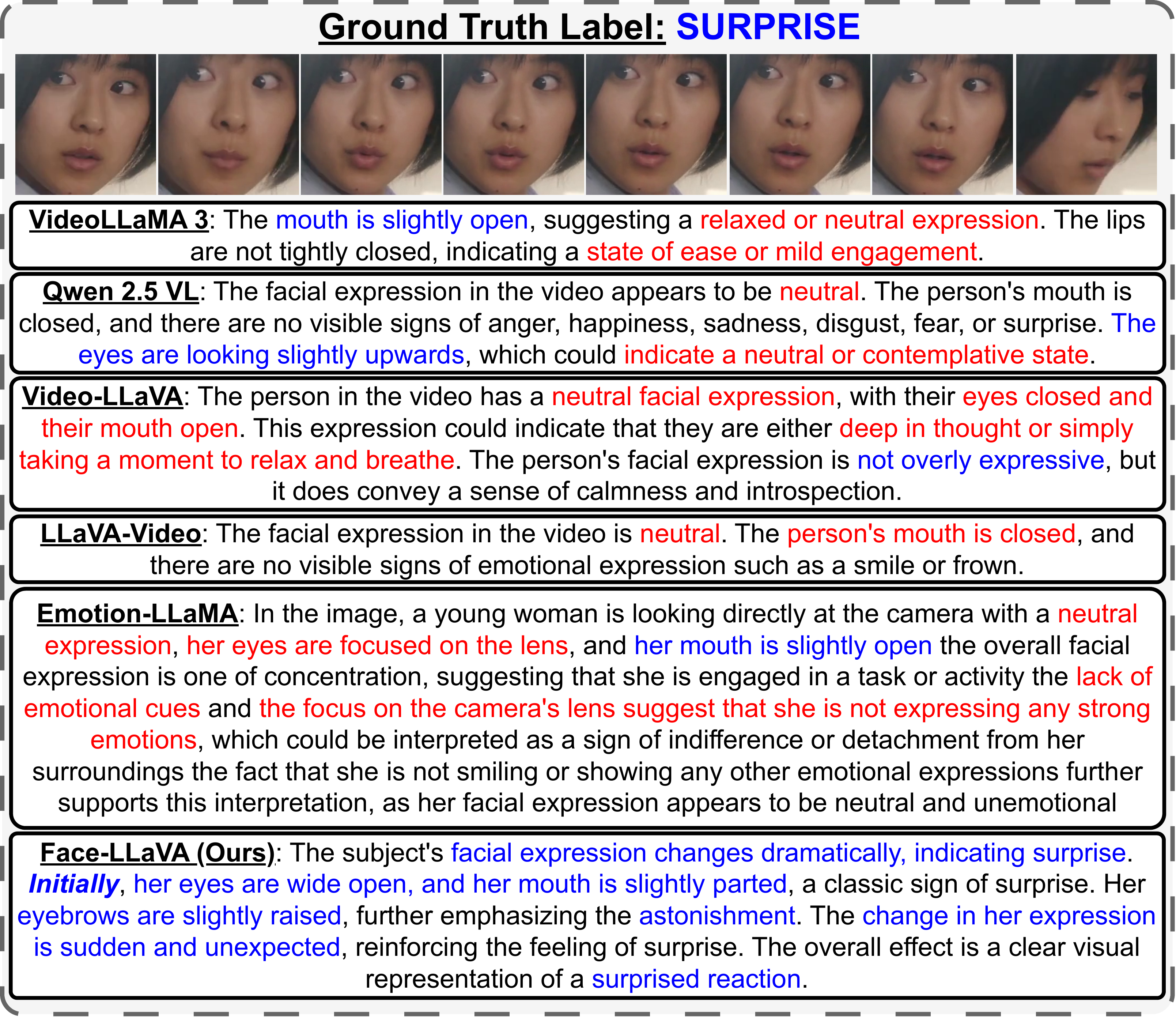}
    \end{subfigure}%
    \caption{Comparison of descriptions or reasoning obtained by Face-LLaVA with other baselines for facial expression recognition. \textcolor{blue}{Blue text} indicates alignment with the ground truth and \textcolor{red}{red text} indicates wrong reasoning or hallucinations.}
    \label{fig:dfew_samples_comparison_baselines}
\end{figure}

\begin{figure}[t!]
    \centering
    \begin{subfigure}[t]{0.32\textwidth}
        \centering
        \includegraphics[width=\textwidth]{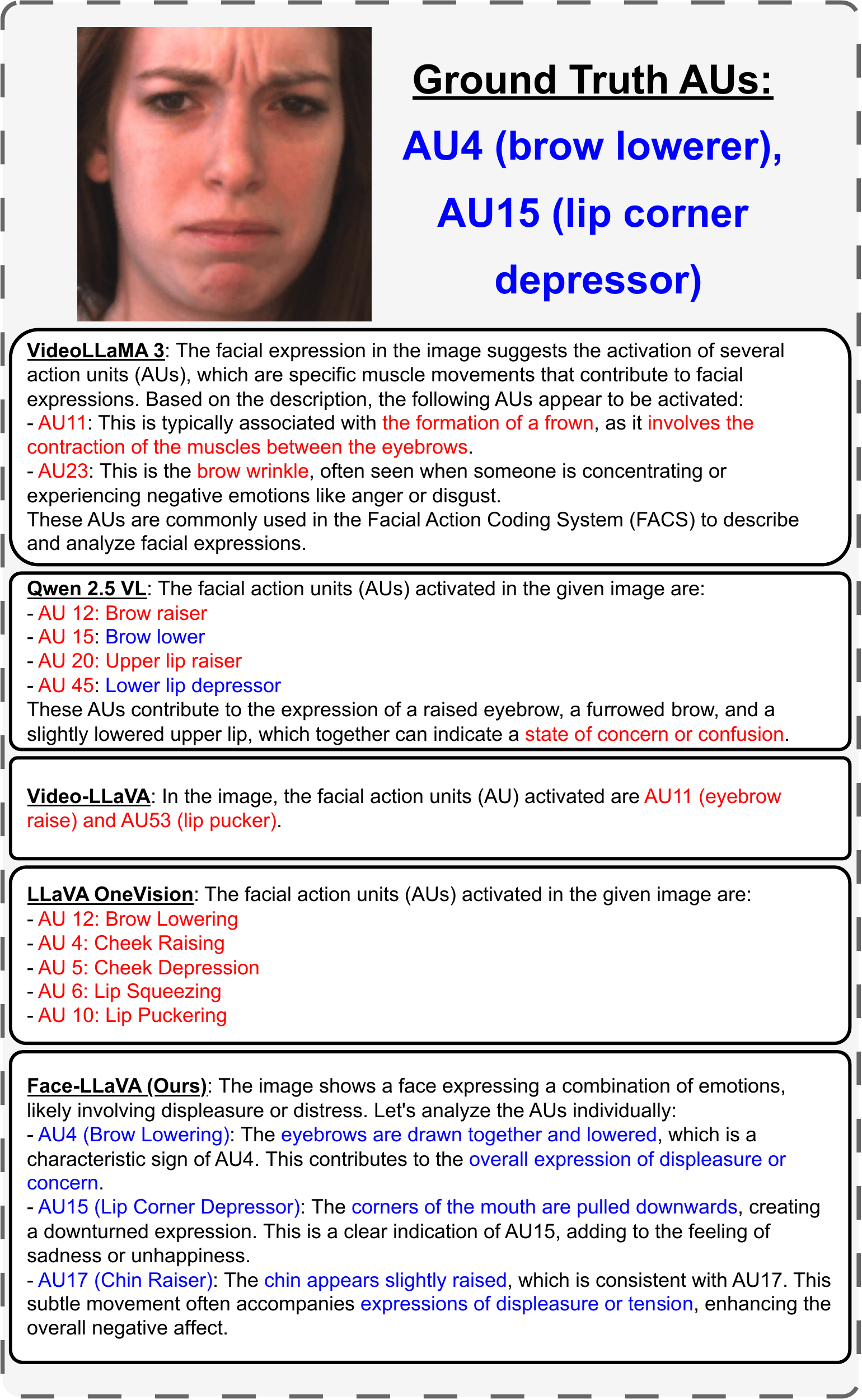}
    \end{subfigure} 
    \hfill
    \begin{subfigure}[t]{0.32\textwidth}
        \centering
        \includegraphics[width=\textwidth]{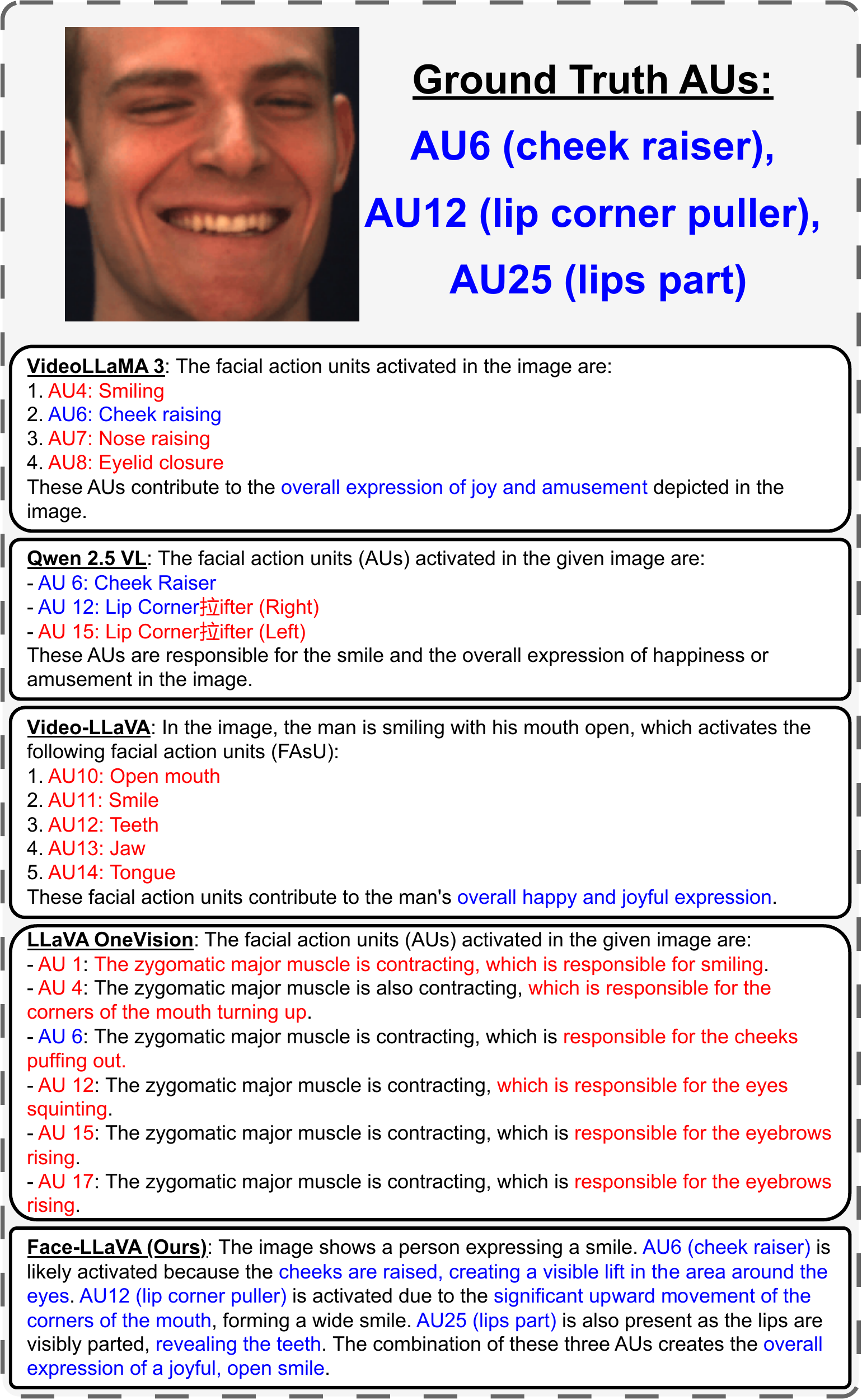}
    \end{subfigure}
    \hfill
    \begin{subfigure}[t]{0.32\textwidth}
        \centering
        \includegraphics[width=\textwidth]{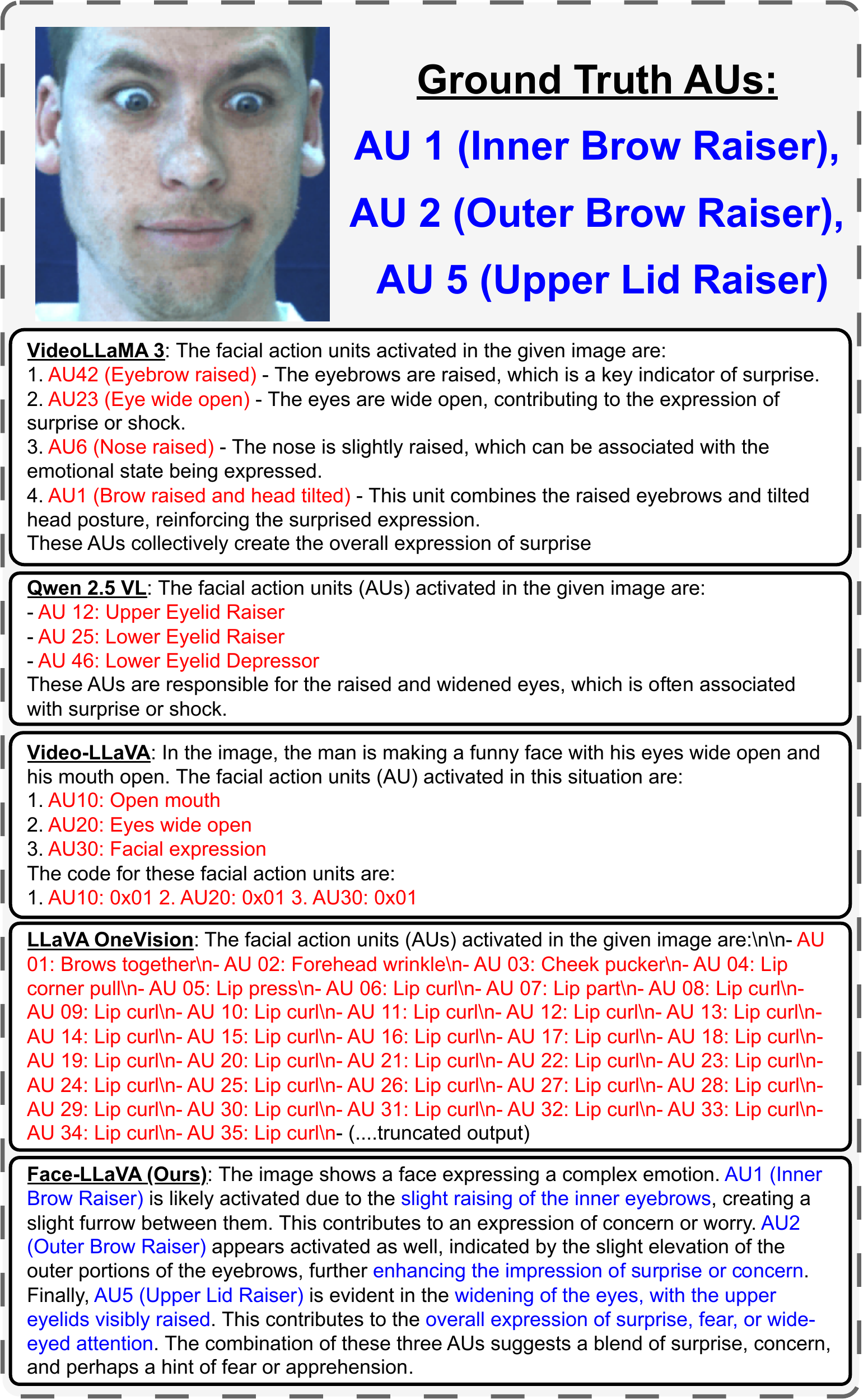}
    \end{subfigure} 
    \caption{Comparison of descriptions or reasoning obtained by Face-LLaVA with other baselines for action unit detection. \textcolor{blue}{Blue text} indicates alignment with the ground truth and \textcolor{red}{red text} indicates wrong reasoning or hallucinations. Notice that for AU detection, all the existing baselines hallucinate badly in getting the FACS action unit codes correct. However, our model is able to correctly capture both the FACS encoding and the correct activated AUs. Moreover, the left-most sample had a missing annotation for AU 17 (chin raiser) which is clearly visible in the image and Face-LLaVA correctly identifies the same.}
    \label{fig:disfa_samples_comparison_baselines}
\end{figure}

\begin{figure}[t!]
    \centering
    \begin{subfigure}[t]{0.32\textwidth}
        \centering
        \includegraphics[width=\textwidth]{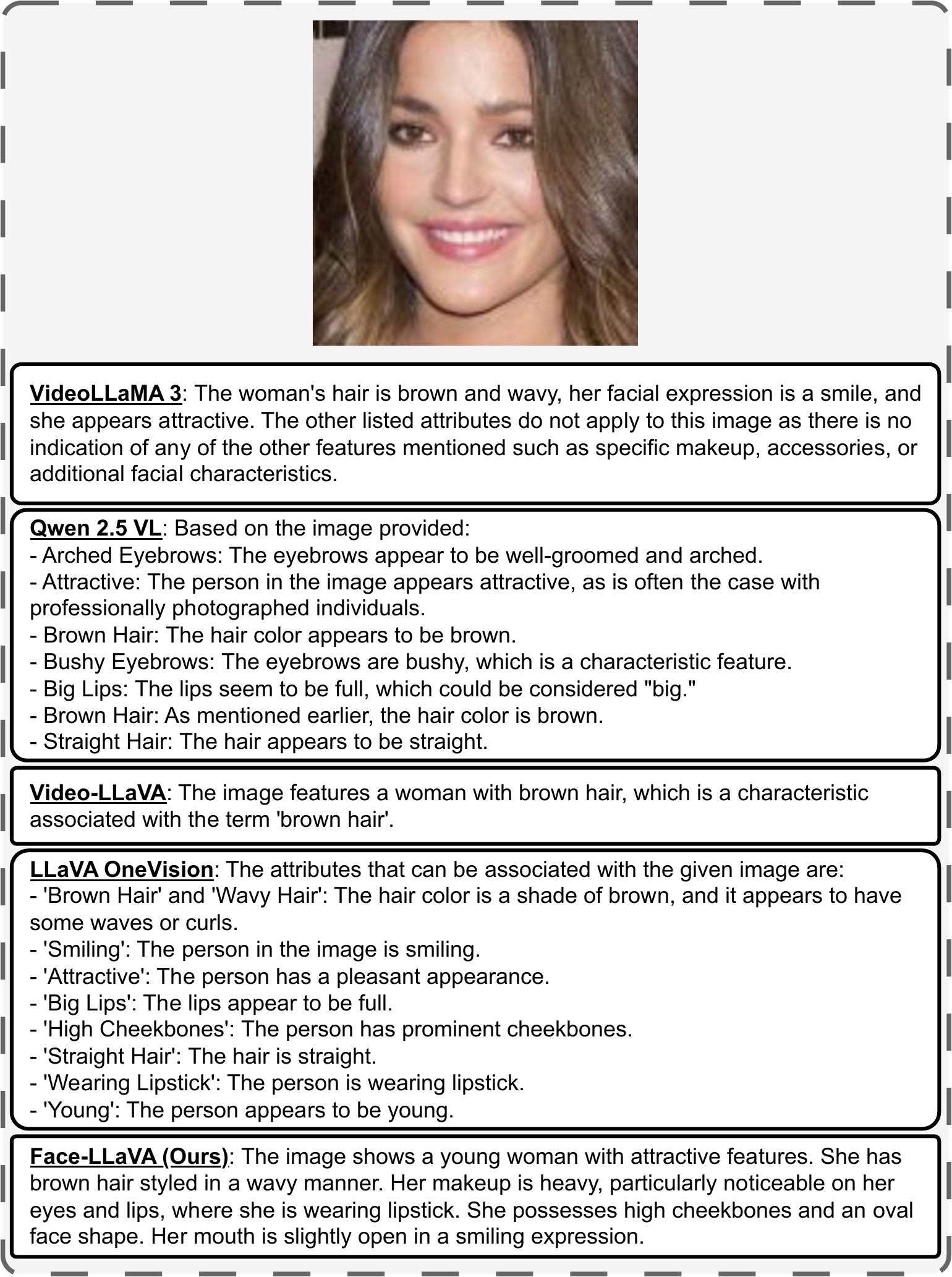}
    \end{subfigure} 
    \hfill
    \begin{subfigure}[t]{0.32\textwidth}
        \centering
        \includegraphics[width=\textwidth]{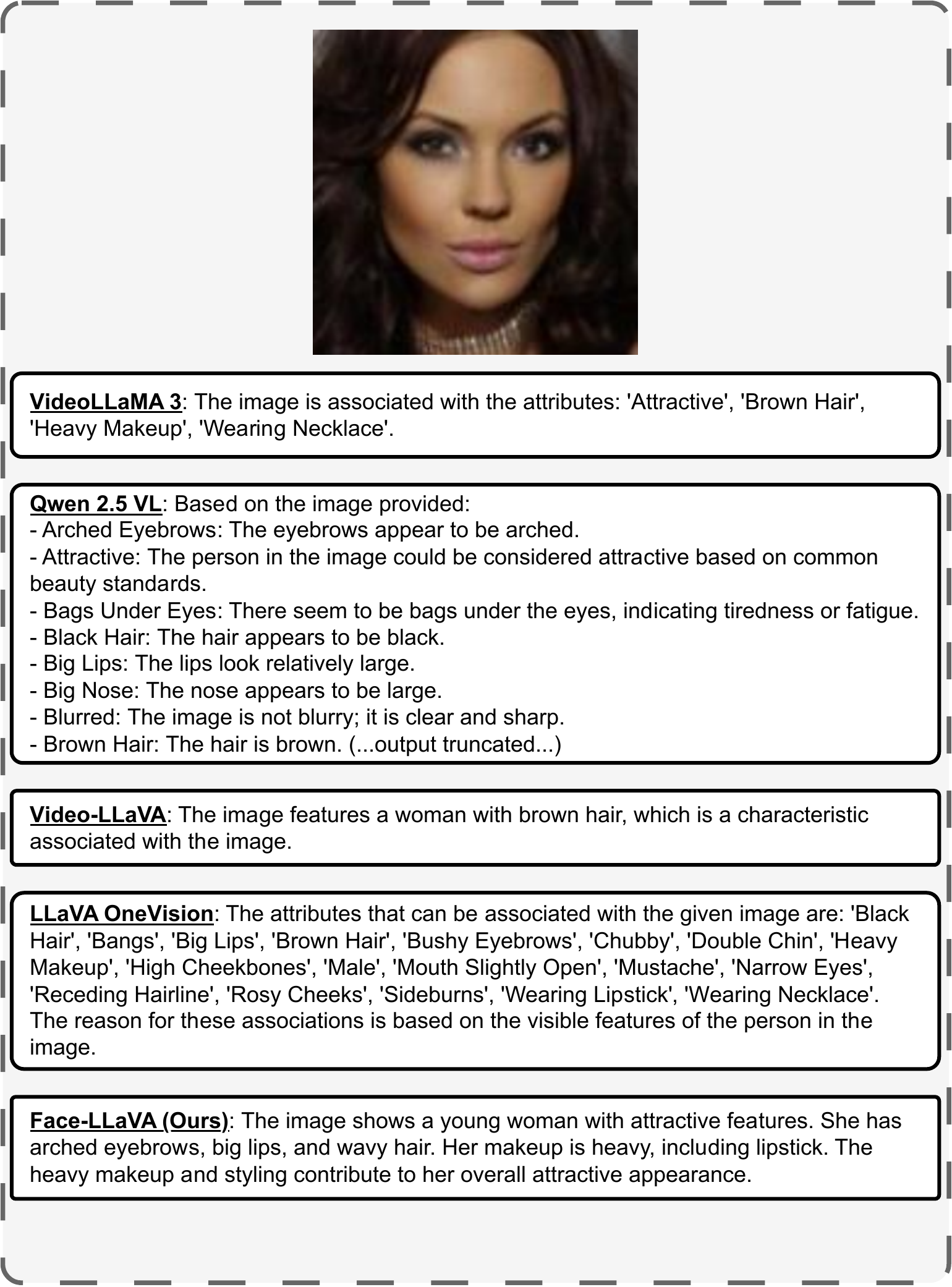}
    \end{subfigure}
    \hfill
    \begin{subfigure}[t]{0.32\textwidth}
        \centering
        \includegraphics[width=\textwidth]{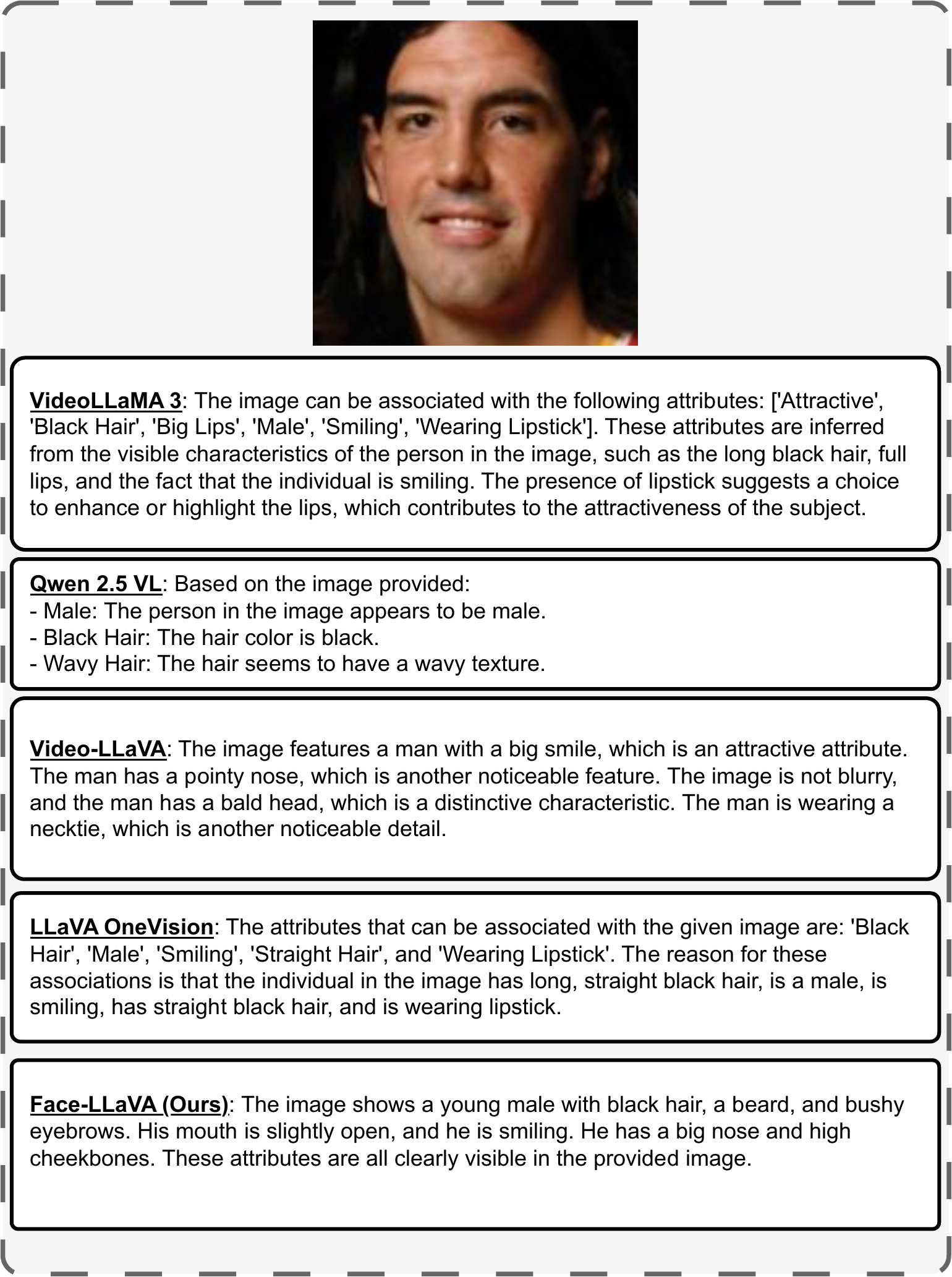}
    \end{subfigure} 
    \caption{Comparison of descriptions or reasoning obtained by Face-LLaVA with other baselines for facial attribute detection. Note that for this task, we do not provide the ground truth label in the figure as the ground truth is clearly identifiable from the image. Also, note that for getting the baseline predictions for this task, we provide the list of possible attributes to the input prompt for the baselines to restrict their responses. Moreover, for computing the mean accuracy on CelebA \cite{attr_liu_celeba} test set for this task for the baselines, we use a separate prompt where we ask the model to just detect the visible attributes and not describe them. This is done because our synonym matching does not work so accurately on formatted text (as shown above) generated by the baselines.}
    \label{fig:celeba_samples_comparison_baselines}
\end{figure}

\begin{figure}[t!]
    \centering
    \begin{subfigure}[t]{0.32\textwidth}
        \centering
        \includegraphics[width=\textwidth]{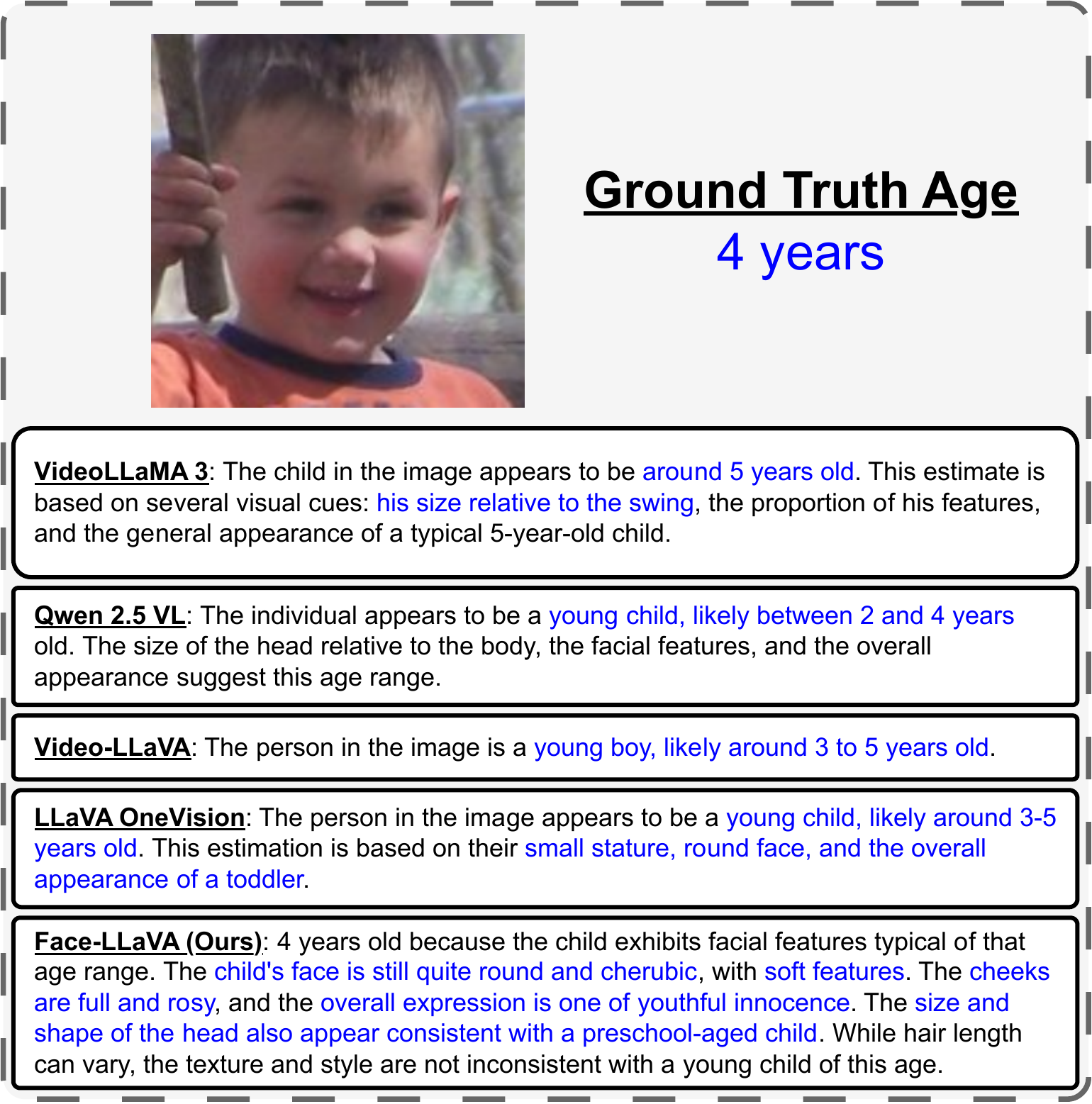}
    \end{subfigure} 
    \hfill
    \begin{subfigure}[t]{0.32\textwidth}
        \centering
        \includegraphics[width=\textwidth]{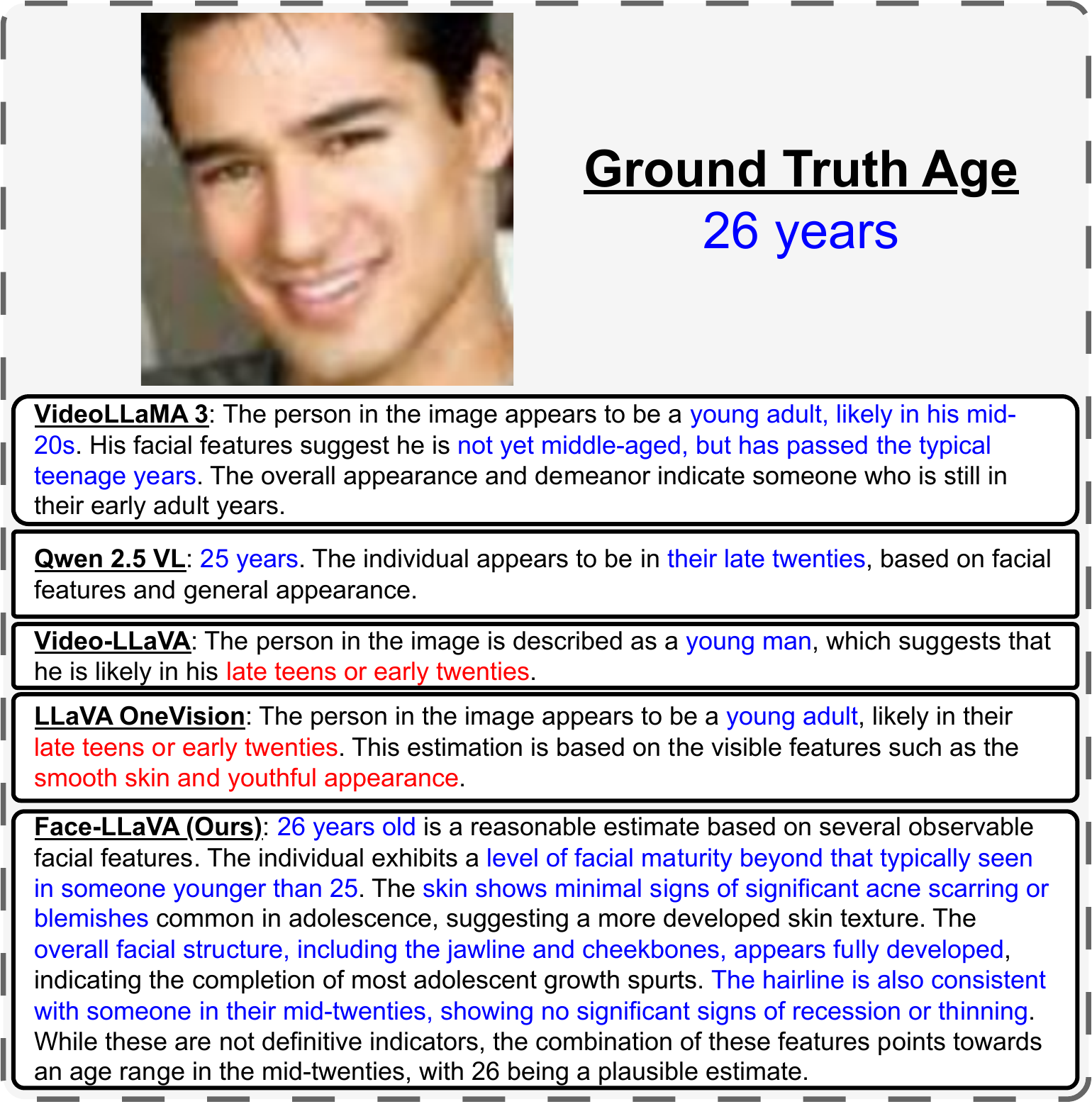}
    \end{subfigure}
    \hfill
    \begin{subfigure}[t]{0.32\textwidth}
        \centering
        \includegraphics[width=\textwidth]{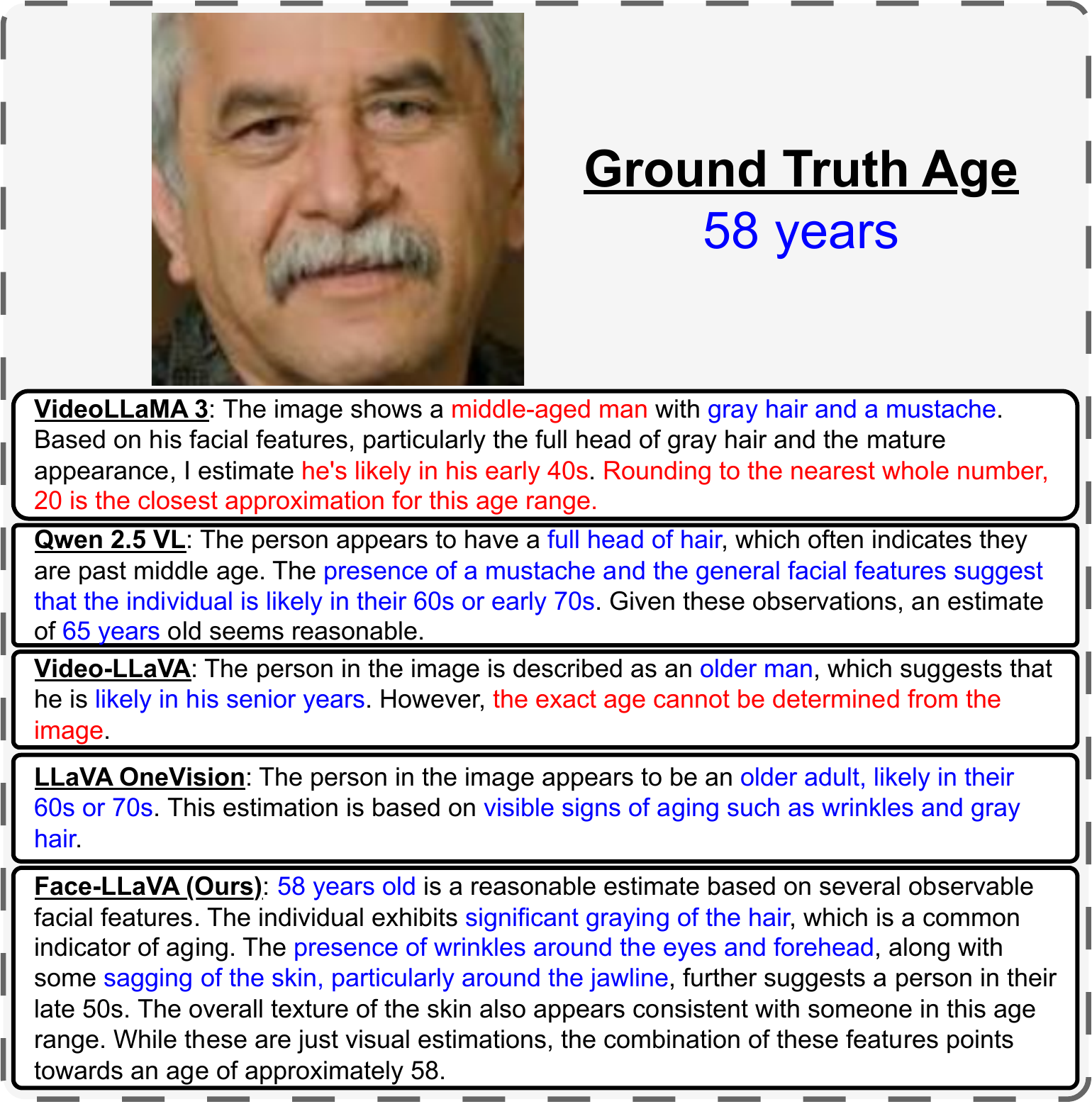}
    \end{subfigure} 
    \caption{Comparison of descriptions or reasoning obtained by Face-LLaVA with other baselines for age estimation. \textcolor{blue}{Blue text} indicates alignment with the ground truth and \textcolor{red}{red text} indicates wrong reasoning or hallucinations. We can notice that the baselines provide minimal description for reasoning their predictions. Moreover, the reason is not based only on the facial features and has some inconsistencies (e.g. VideoLLaMA 3 \cite{damonlpsg2025videollama3} makes wrong age prediciton even with correct reasoning for the right-most example.) In contrast to the baselines, Face-LLaVA is able to predict the age of the person in the image correctly with a detailed description specifically related to the facial features of the person in the image. Finally, it is important to note that we use a different prompt when we compute the performance of the baselines on traditional metrics (mean absolute error - refer \cref{subsec:traditional_metrics}) as the baselines usually provide an age range in their descriptions and usual string parsing will have some inconsistencies due to that.}
    \label{fig:utkface_samples_comparison_baselines}
\end{figure}

\begin{figure}[t!]
    \centering
    \begin{subfigure}[t]{0.32\textwidth}
        \centering
        \includegraphics[width=\textwidth]{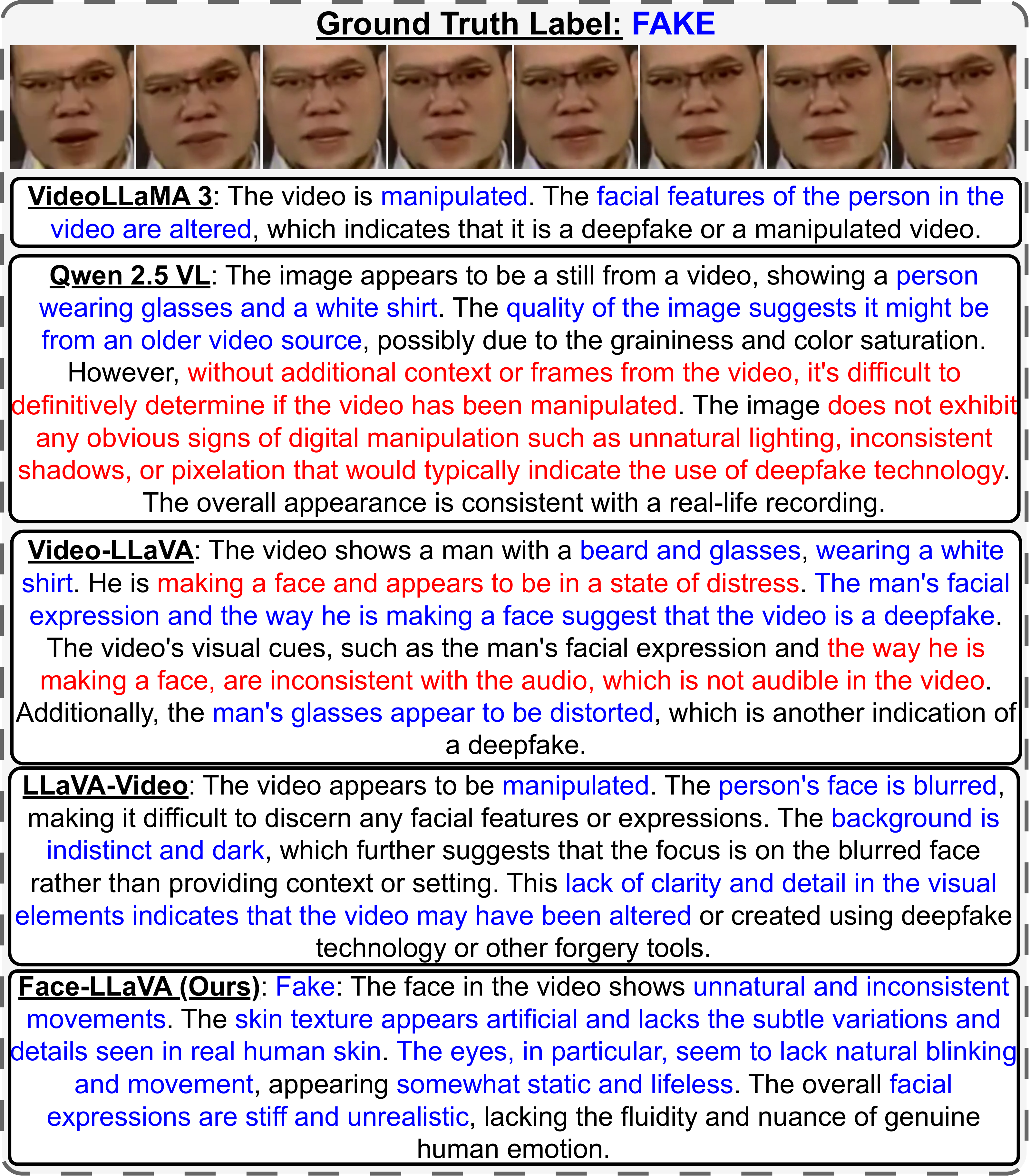}
    \end{subfigure} 
    \hfill
    \begin{subfigure}[t]{0.32\textwidth}
        \centering
        \includegraphics[width=\textwidth]{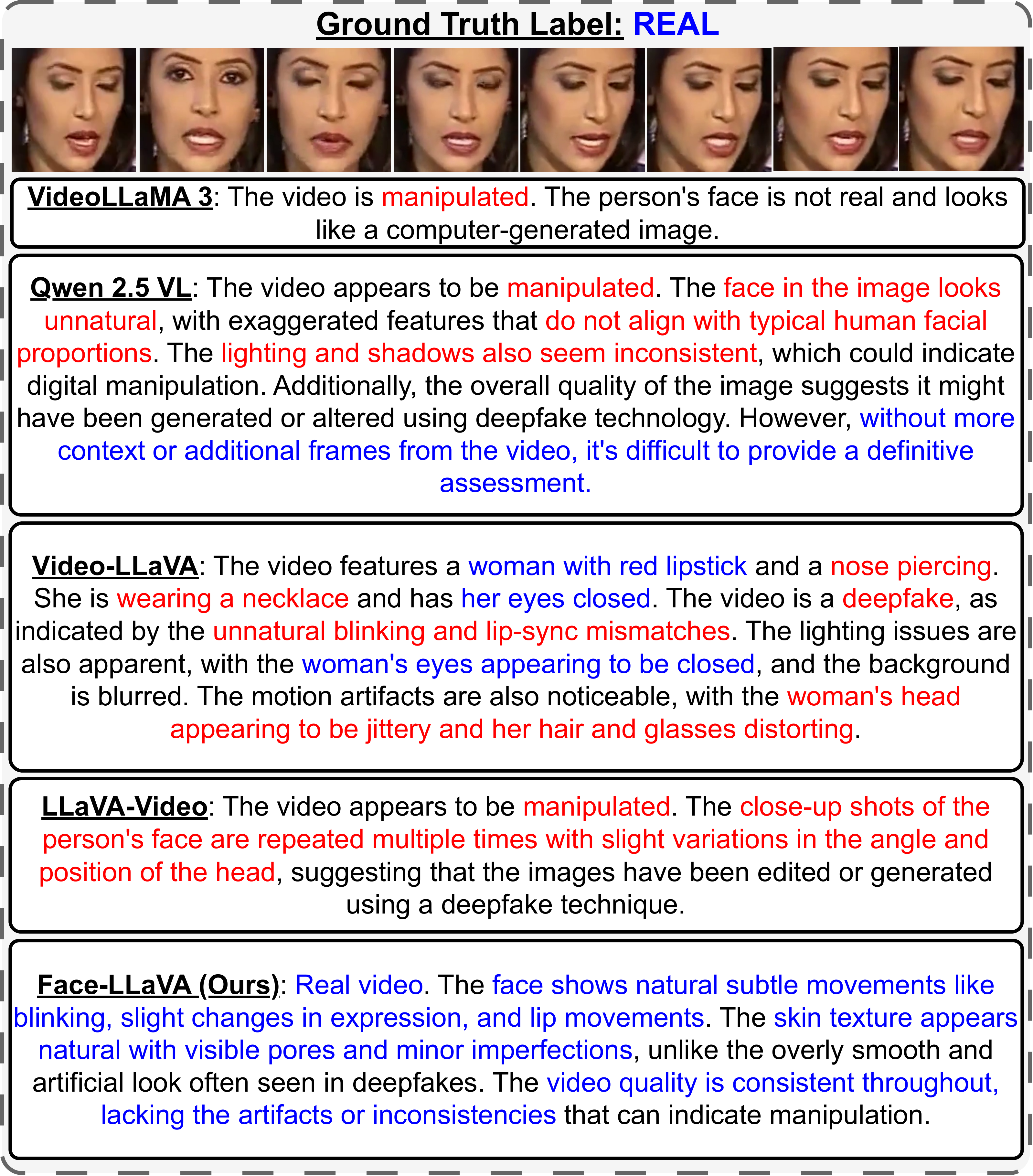}
    \end{subfigure}
    \hfill
    \begin{subfigure}[t]{0.32\textwidth}
        \centering
        \includegraphics[width=\textwidth]{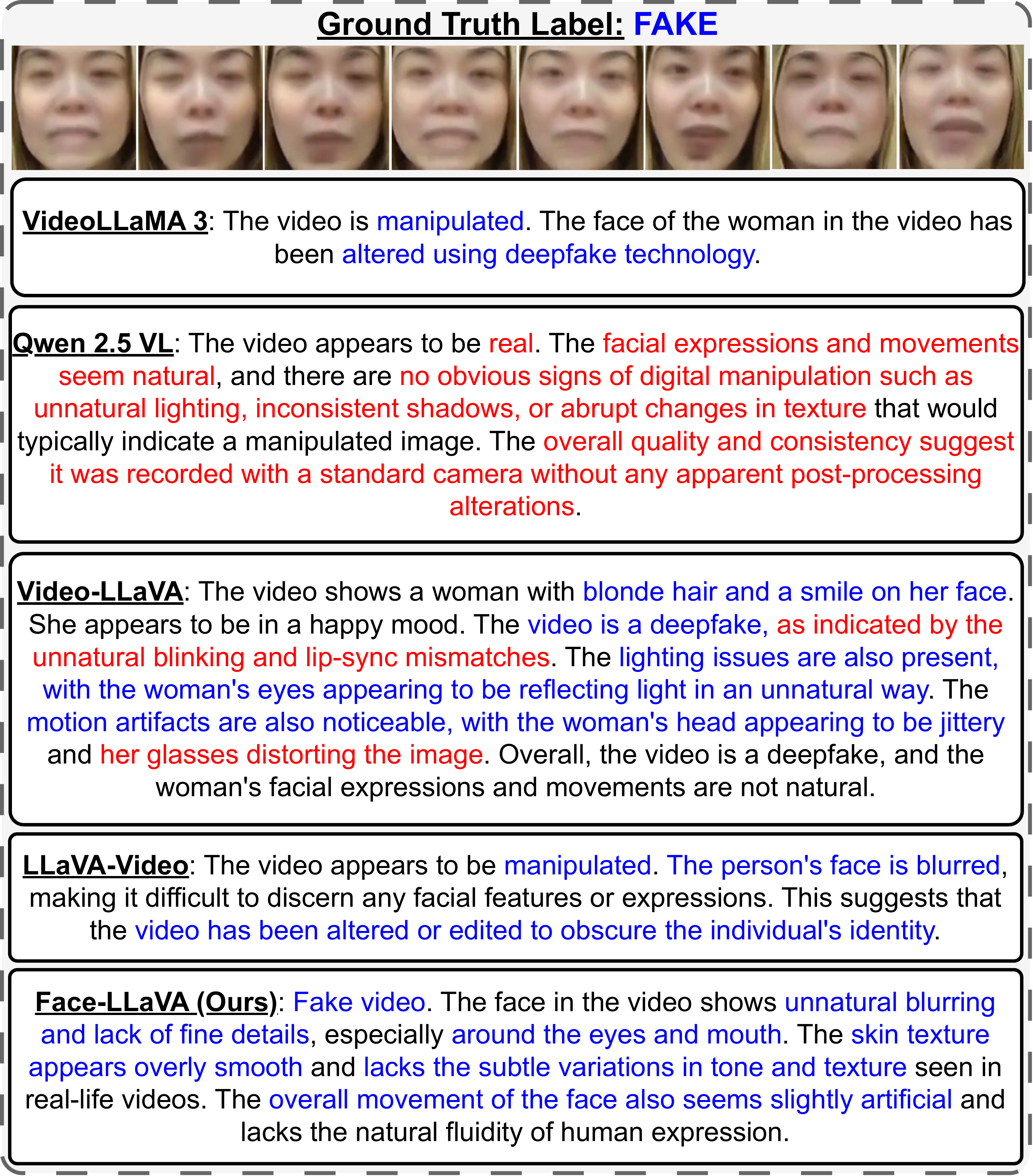}
    \end{subfigure} 
    \caption{Comparison of descriptions or reasoning obtained by Face-LLaVA with other baselines for deepfake detection. \textcolor{blue}{Blue text} indicates alignment with the ground truth and \textcolor{red}{red text} indicates wrong reasoning or hallucinations. Other MLLMs hallucinate with their reasoning and provide descriptions that are not face-specific. Face-LLaVA on the other hand provides consistent descriptions that are face specific. Similar to attribute detection and age estimation, we use a different prompt for computing accuracy for the baselines on the FaceForensics++ \cite{roessler2019faceforensicspp} dataset for \cref{tab:age_celeba_dfd}.}
    \label{fig:deepfake_samples_comparison_baselines}
\end{figure}

In this section, we illustrate the reasoning capabilities of Face-LLaVA in comparison to the baseline MLLMs. For each task, we pick diverse samples from the \emph{FaceInstruct-1M} test set to show the comparison. \cref{fig:dfew_samples_comparison_baselines,fig:disfa_samples_comparison_baselines,fig:utkface_samples_comparison_baselines,fig:celeba_samples_comparison_baselines,fig:deepfake_samples_comparison_baselines} show this comparison in detail. Readers are encouraged to refer to the caption of these figures for a detailed description about the samples.

\section{Failure cases}
\label{sec:failure_cases}

\begin{figure*}
    \centering
    \includegraphics[width=0.85\linewidth]{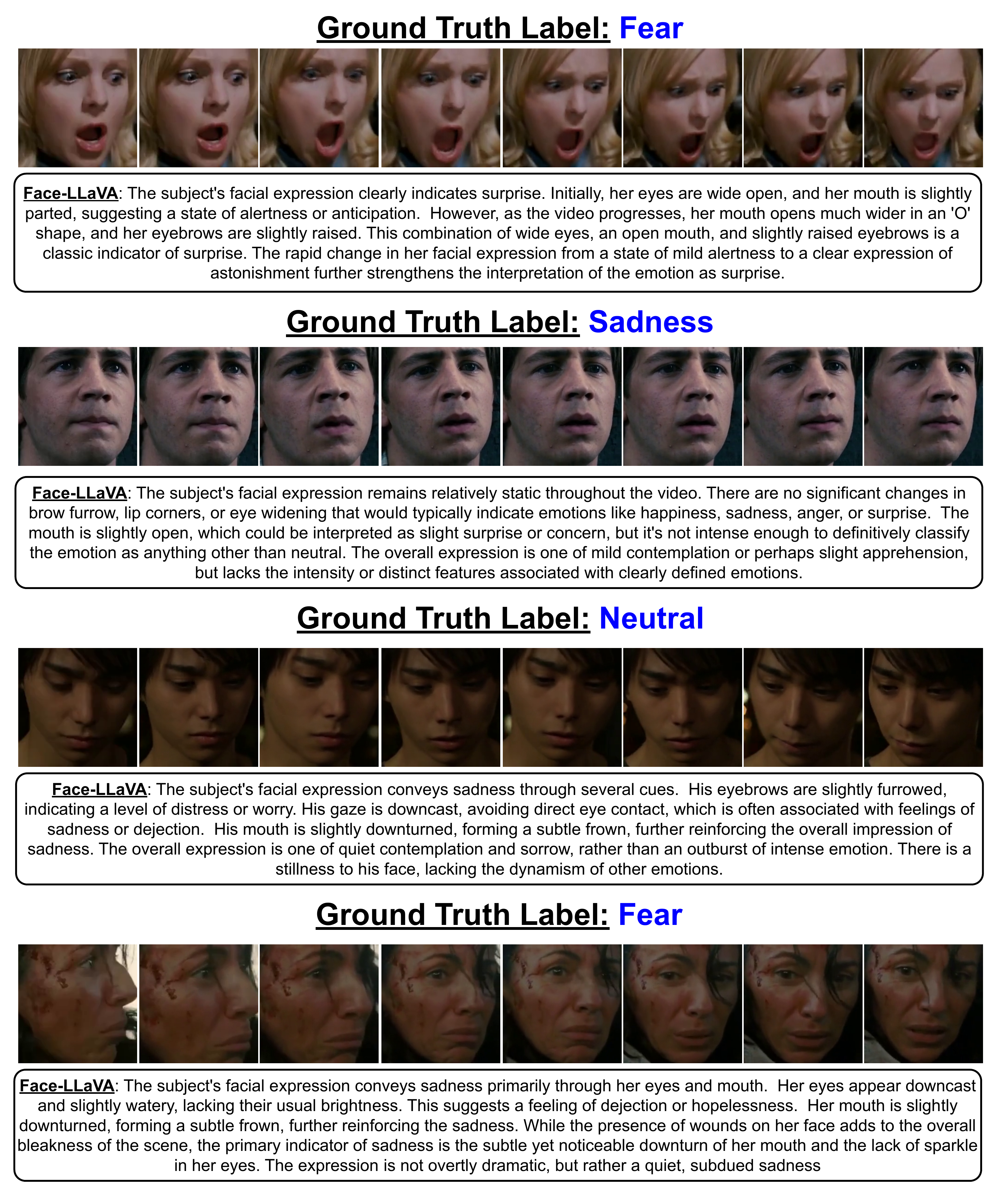}
    \caption{Failure cases for facial expression recognition task on some samples of DFEW \cite{jiang2020dfew}. In the samples above, expressions labeled as ground truth \emph{fear} are often misclassified as \emph{sadness} or \emph{surprise}. Closer inspection of the videos reveals that these instances indeed exhibit mixed emotions. Additionally, ground truth labels are assigned based on the full scene, incorporating background context and audio, whereas Face-LLaVA makes predictions solely from the face-cropped video.}
    \label{fig:failure_fer}
\end{figure*}

\begin{figure*}
    \centering
    \includegraphics[width=0.85\linewidth]{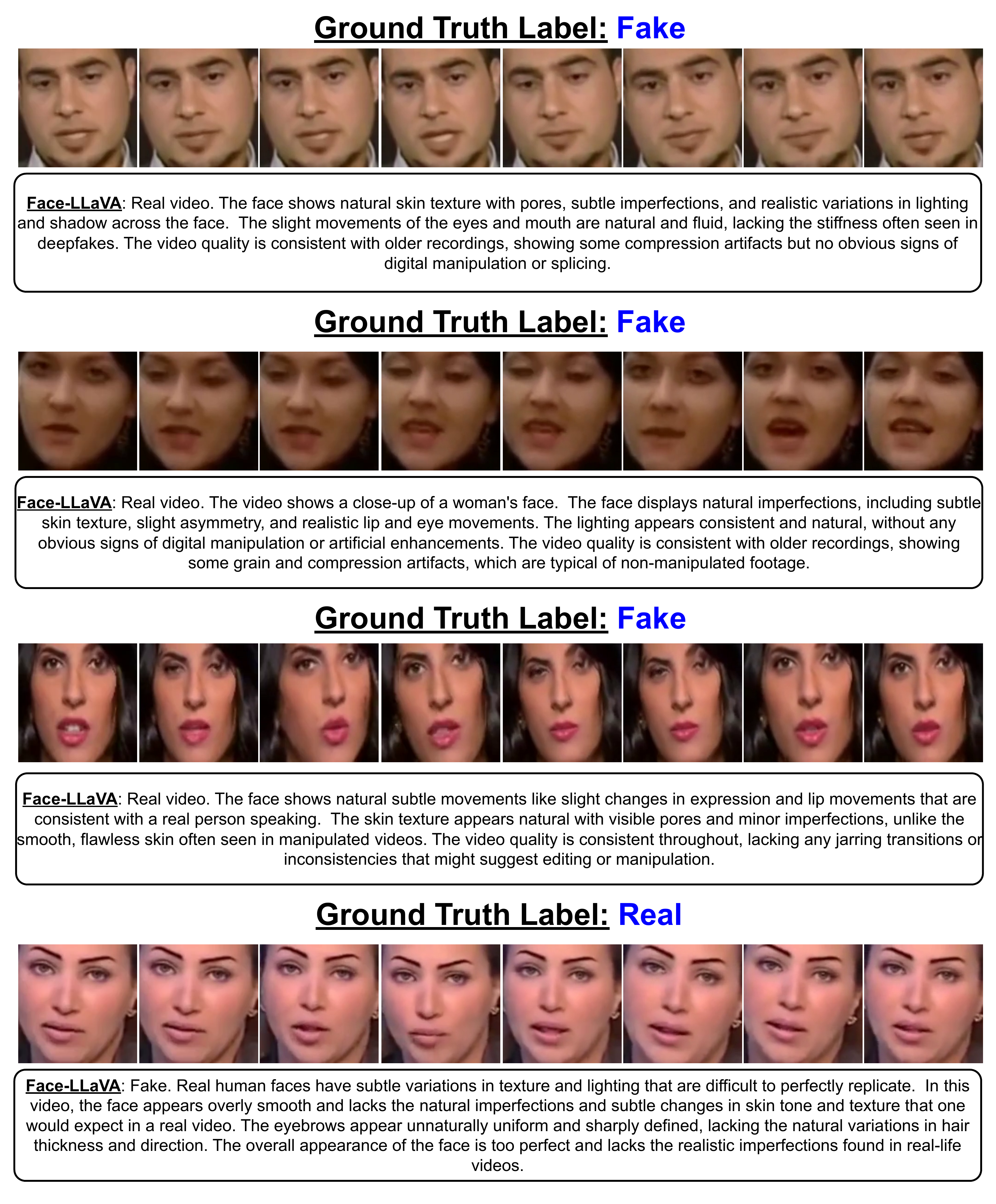}
    \caption{Failure cases for deepfake detection task on some samples of FaceForensics++ \cite{roessler2019faceforensicspp}. In the first and third samples above, determining whether the videos are manipulated is challenging even for human beings, as they were altered using a face re-enactment technique (NeuralTextures). In the last example, the model mistakenly classifies a real video as fake, attributing its decision to the smoothness of the skin and makeup around the eyes, interpreting the face as \emph{“too perfect”} to be real.}
    \label{fig:failure_df}
\end{figure*}

\begin{figure*}
    \centering
    \includegraphics[width=0.8\linewidth]{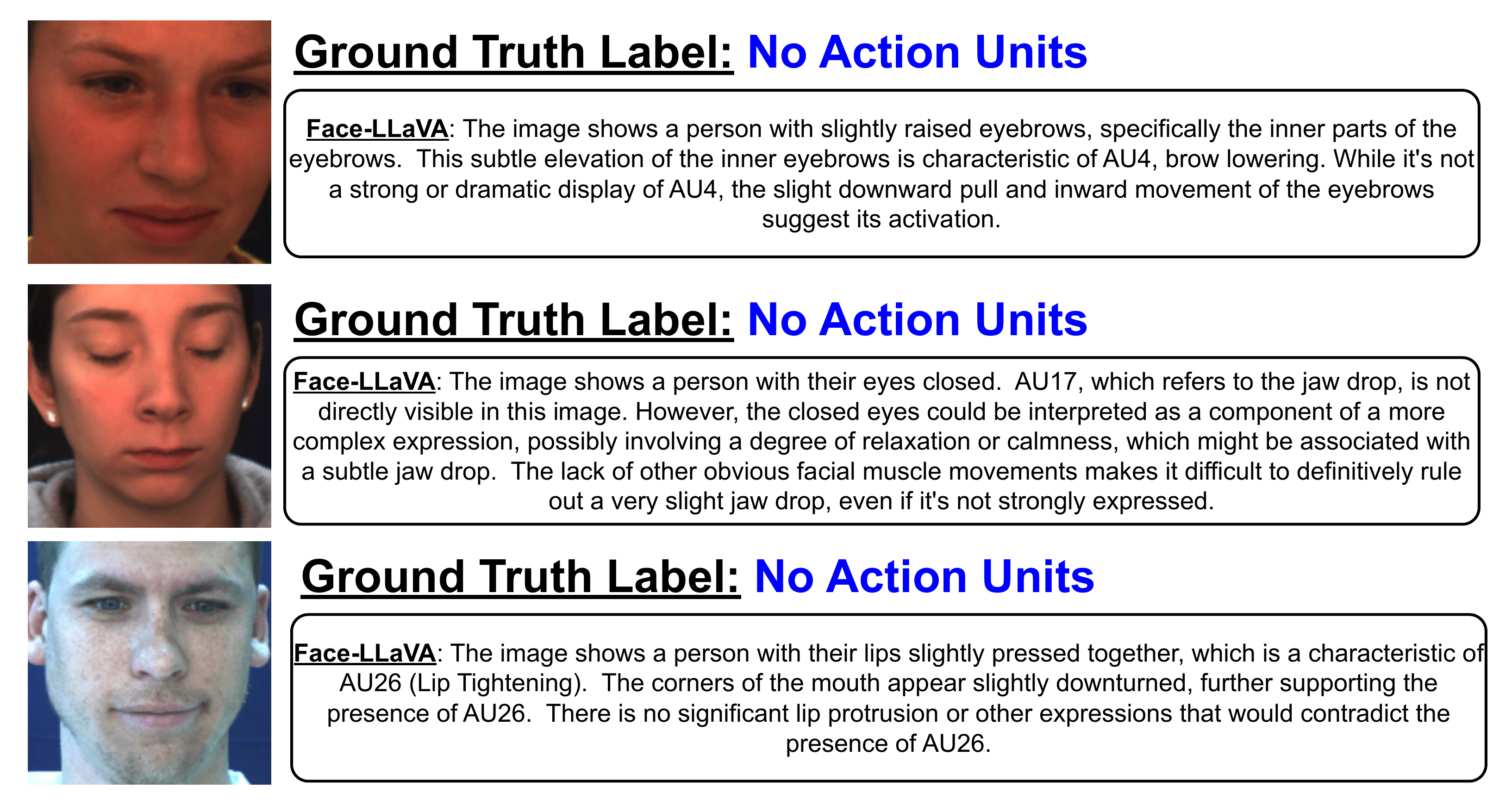}
    \caption{Failure cases for action unit detection task on some samples of DISFA \cite{disfa_dataset}. Our model fails mostly on the cases when the ground truth sample does not have any action units activated.}
    \label{fig:failure_au}
\end{figure*}

\begin{figure*}
    \centering
    \includegraphics[width=0.8\linewidth]{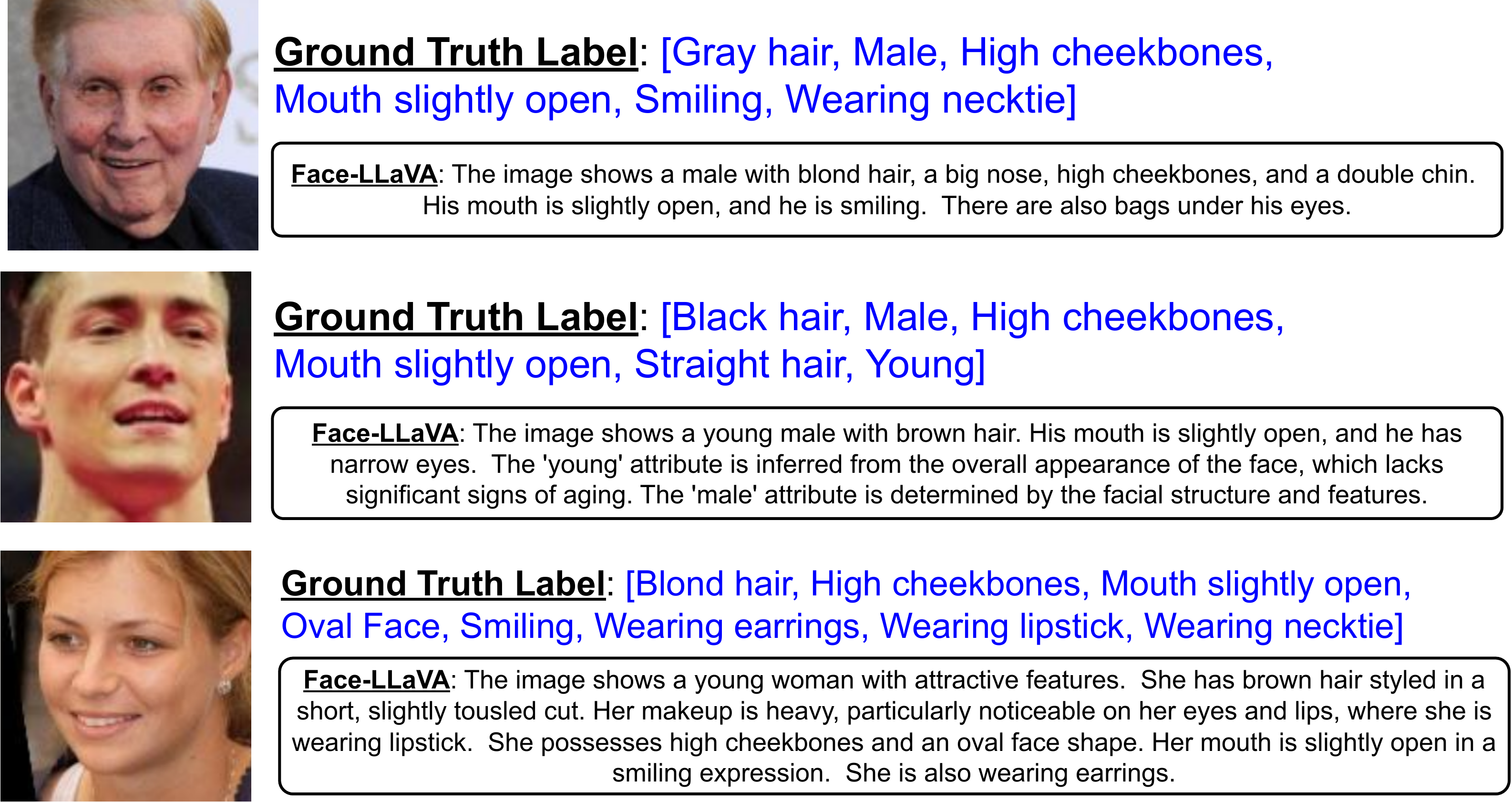}
    \caption{Failure cases for facial attribute detection task on some samples of CelebA \cite{attr_liu_celeba}. In the above examples, while the model's predictions slightly differ from the ground truth, they remain consistent with the visual content of the image, suggesting potential imperfections in the dataset annotations of CelebA.}
    \label{fig:failure_attr}
\end{figure*}

\begin{figure*}
    \centering
    \includegraphics[width=0.8\linewidth]{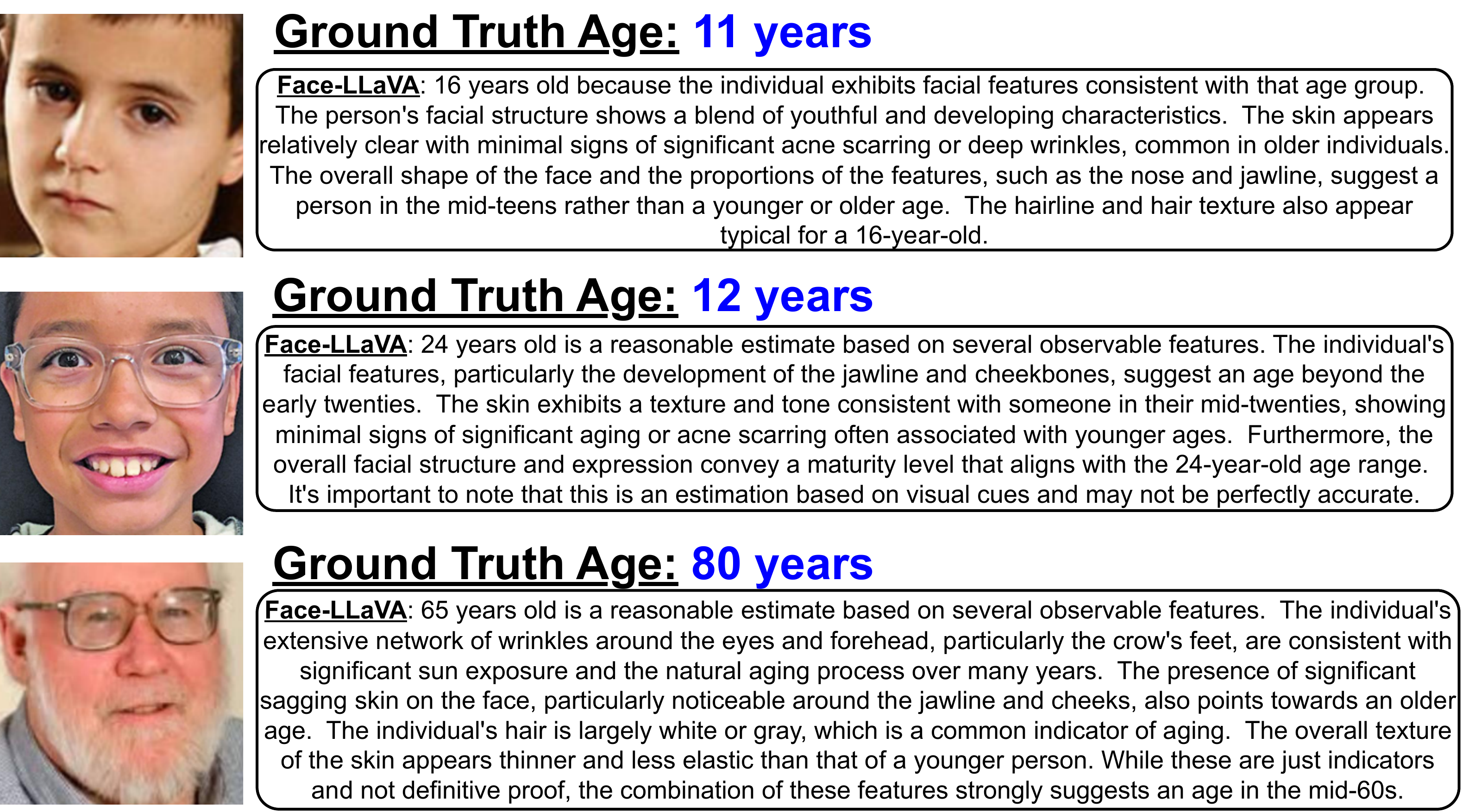}
    \caption{Failure cases for age estimation task on some samples of UTKFace \cite{zhifei2017cvpr_utkface}. Except for the middle sample, the error in prediction for the other two samples are acceptable and even for humans it is hard to judge the age.}
    \label{fig:failure_age}
\end{figure*}

This section contains the failure cases for Face-LLaVA for different tasks. \cref{fig:failure_age,fig:failure_au,fig:failure_df,fig:failure_fer,fig:failure_attr} show examples for each of the tasks where Face-LLaVA gives slightly inconsistent responses. In most of the cases, the inaccuracy of prediction with the ground truth label is due to a corner case (multiple expressions present in the face, no action unit activated, etc.) or when the prediction is difficult to make even for humans (deepfake detection and age estimation). In almost all the cases the model output and descriptions align with the video or the image thereby showing excellent vision language grounding for facial tasks.


\end{document}